\documentclass[10pt,twocolumn,letterpaper]{article}

\usepackage{cvpr}              %

\usepackage[table]{xcolor}
\usepackage{graphicx}
\usepackage{amsmath}
\usepackage{amssymb}
\usepackage{booktabs}
\usepackage{subcaption}

\usepackage{array}
\usepackage{xfrac}
\newcolumntype{P}[1]{>{\centering\arraybackslash}p{#1}}

\usepackage[pagebackref,breaklinks,colorlinks]{hyperref}

\usepackage[capitalize]{cleveref}
\crefname{section}{Sec.}{Secs.}
\Crefname{section}{Section}{Sections}
\Crefname{table}{Table}{Tables}
\crefname{table}{Tab.}{Tabs.}

\newcommand{\Mat}{\boldsymbol}

\newcommand{\real}{\mathbb{R}}

\DeclareMathOperator{\mean}{\mathbb{E}}
\DeclareMathOperator{\KL}{\mathcal{D}_{KL}}

\def\confName{CVPR}
\def\confYear{2023}

\begin{document}

\title{NeuralLift-360: Lifting An In-the-wild 2D Photo to A 3D Object with 360\textdegree~Views}

\author{%
Dejia Xu \\
\texttt{\small dejia@utexas.edu} \\
\And
Yifan Jiang \\
\texttt{\small yifanjiang97@utexas.edu} \\
Peihao Wang\\
\texttt{\small peihaowang@utexas.edu} \\
\AND
\And
Zhiwen Fan \\
\texttt{\small zhiwenfan@utexas.edu} \\
\And
Zhangyang Wang \\
\texttt{\small atlaswang@utexas.edu} \\
}

\author{
  Dejia Xu, Yifan Jiang, Peihao Wang, Zhiwen Fan, Yi Wang, Zhangyang Wang\\
  {VITA Group, University of Texas at Austin}\\
  \small{\texttt{\{dejia,atlaswang\}@utexas.edu}} \\
}

\maketitle

\begin{abstract}
Virtual reality and augmented reality (XR) bring increasing demand for 3D content generation.
However, creating high-quality 3D content requires tedious work from a human expert.
In this work, we study the challenging task of lifting a single image to a 3D object and, for the first time, demonstrate the ability to generate a plausible 3D object with $360^\circ$ views that corresponds well with the given reference image. By conditioning on the reference image, our model can fulfill the everlasting curiosity for synthesizing novel views of objects from images. Our technique sheds light on a promising direction of easing the workflows for 3D artists and XR designers. 
We propose a novel framework, dubbed \textbf{NeuralLift-360}, that utilizes a depth-aware neural radiance representation (NeRF) and learns to craft the scene guided by denoising diffusion models. 
By introducing a ranking loss, our NeuralLift-360 can be guided with rough depth estimation in the wild.
We also adopt a CLIP-guided sampling strategy for the diffusion prior to provide coherent guidance.
Extensive experiments demonstrate that our NeuralLift-360 significantly outperforms existing state-of-the-art baselines.
Project page: \url{https://vita-group.github.io/NeuralLift-360/}
\end{abstract}

\section{Introduction}
\label{sec:intro}

Creating 3D content has been a long-standing problem in computer vision.
This problem enables various applications in game studios, 
home decoration, virtual reality, and augmented reality. 
Over the past few decades, the manual task has dominated real scenarios, which requires tedious professional expert modeling. Modern artists rely on special software tools (e.g., Blender, Maya3D, 3DS Max, etc.) and time-demanding manual adjustments to realize imaginations and transform them into virtual objects. Meanwhile, automatic 3D content creation pipelines serve as effective tools to facilitate human efforts.
These pipelines
typically capture hundreds of images and leverage multi-view stereo~\cite{yao2018mvsnet} to quickly model fine-grained virtual landscapes.

More recently, researchers have started aiming at a more ambitious goal, to create a 3D object from a single image~\cite{durou2008numerical,li2018megadepth,favaro2005geometric,johnston2017scaling,fan2017point,wu2017marrnet,saito2019pifu,wang2018pixel2mesh}. This enables broad applications since it greatly reduces the prerequisite to a minimal request. Existing approaches can be mainly categorized into two major directions.
One line of work utilizes learning priors from large-scale datasets with multi-view images~\cite{wang2018pixel2mesh,saito2019pifu,wu2017marrnet,fan2017point,johnston2017scaling}. These approaches usually learn a conditional neural network to predict 3D information based on input images. However, due to their poor generalization ability, drastic performance degradation is observed when the testing image is out-of-distribution.

Another direction constructs the pipeline on top of the depth estimation techniques~\cite{wang2020normalgan}. Under the guidance of monocular depth estimation networks~\cite{ranftl2021vision,ranftl2020towards}, the 2D image is firstly back-projected to a 3D data format (e.g., point cloud or multi-plane image) and then re-projected into novel views. After that, advanced image inpainting techniques are then adopted to fill missing holes~\cite{shih20203d} produced during the projection. However, most of them can be highly affected by the quality of the estimated depth maps. Though LeRes~\cite{leres} attempted to rectify the predicted depth by refining the projected 3D point cloud, their results do not generalize well to an arbitrary image in the wild.
Overall, the aforementioned approaches are either adopted in limited scenarios (e.g. face or toy examples~\cite{shapenet2015,yu2021pixelnerf}), or only produce limited viewing directions~\cite{xu2022sinnerf,deng2022depth} when being applied to scenes in the wild. Different from all these approaches, we focus on a more challenging task and for the first time, show promising results by \textbf{lifting a single in-the-wild image into a 3D object with $360^\circ$ novel views}.

Attracted by the dramatic progress of neural volumetric rendering on 3D reconstruction tasks, we consider building our framework based on Neural Radiance Fields (NeRFs)~\cite{mildenhall2021nerf}. The original NeRF takes hundreds of training views and their camera poses as inputs to learn an implicit representation. Subsequent models dedicate tremendous efforts~\cite{jain2021putting,yu2021pixelnerf,deng2022depth,xu2022sinnerf}  to apply NeRF to sparse training views. The most similar work~\cite{xu2022sinnerf} to our method proposes to optimize a NeRF using only a single image and its corresponding depth map. However, it renders limited views from a small range of angles, and the prerequisite of a high-quality depth map largely constrains its practical usage.

To address the above issues, we propose a novel framework, coined as \textbf{NeuralLift-360}, which aims to ease the creation of 3D assets by building a bridge to convert diverse in-the-wild 2D photos to sophisticated 3D contents in $360^\circ$ views and enable its automation. The major challenge in our work is that the content on the back side is hidden and hard to hallucinate. To tackle these hurdles, we consider the diffusion priors together with the monocular depth estimator as the cues for hallucination. Modern diffusion models~\cite{imagen, ldm, ramesh2022hierarchical} are trained on a massive dataset (e.g., 5B text-to-image pairs~\cite{ko2022large}). During inference time, they can generate impressive photorealistic photos based on simple text inputs. By adopting these learning priors with CLIP~\cite{radford2021learning} guidance, NeuralLift-360 can generate plausible 3D consistent instances that correspond well to the given reference image while only requiring minimal additional input, the correct user prompts. Moreover, rather than simply taking the depth map from the pre-trained depth estimator as geometry supervision, we propose to use the relative ranking information from the rough depth during the training process. This simple strategy is observed to robustly mitigate geometry errors for depth estimations in the wild.

Our contributions can be summarized as follows,

\begin{itemize}
    \item Given a single image in the wild, we demonstrate promising results of them being lifted to 3D. We use NeRF as an effective scene representation and integrate prior knowledge from the diffusion model.
    \item We propose a CLIP-guided sampling strategy that effectively marries the prior knowledge from the diffusion model with the reference image.
    \item When the reference image is hard to describe exactly, we finetune the diffusion model on the single image while maintaining its ability to generate diverse contents to guide the NeRF training.
    \item We introduce scale-invariant depth supervision that uses the ranking information. This design alleviates the need for accurate multi-view consistent depth estimation and broadens the application of our algorithms.
\end{itemize}

\section{Related Works}

\looseness=-1
\subsection{3D from Single Image}
Inferring 3D worlds from a single image has been a popular direction over the past years. Previous works try to solve this problem by firstly adopting a monocular depth estimator (e.g., MiDaS~\cite{ranftl2020towards}) to predict the 3D geometry and then using multi-plane images (MPI)~\cite{shih20203d,jampani2021slide} or point cloud~\cite{mu20223d} to render artistic effects. Meanwhile, the missing holes from novel views are inpainted by a pre-trained modern deep network~\cite{yu2019free}. While these approaches are powerful and efficient, the depth estimator could be unstable, and the inpainting techniques may produce artifacts that make the rendered images look fake. While our method also applies a monocular depth estimator as the geometry cues, we propose a depth ranking loss that can mitigate the instability of depth estimation in the wild. Later approaches~\cite{yu2021pixelnerf,lin2022vision} train their model on a large-scale 3D assets dataset. During inference time, they directly predict a 3D object using a single image. However, their performance faces drastic degradation when being applied to an in-the-wild image, due to the domain gap.
The most similar work to ours is SinNeRF~\cite{xu2022sinnerf}, which trains a neural radiance field only on a single RGB image and its corresponding depth map. However, their approach requires a high-quality depth map for geometry guidance, and it can only render views from a small range of angles. Thanks to the proposed depth ranking loss, our method can reconstruct impressive geometry information only using guidance from a pre-trained monocular depth estimator.

\subsection{Neural Radiance Fields}
Neural Radiance Fields (NeRFs)~\cite{nerf} have shown encouraging performance for view synthesis.
After its origin, tremendous efforts have been devoted to improving it~\cite{verbin2022ref,barron2021mip,barron2022mip,guo2022nerfren,suhail2022light,chen2022aug,muller2022instant,reiser2021kilonerf,sun2022direct,fridovich2022plenoxels,wang2022clip,fan2022unified,jain2022zero,mvsnerf,wang2021ibrnet,liu2022neural,yu2021pixelnerf}. 
Unlike most NeRF-like models, which at least require several different views to reconstruct scenes, we focus on reconstructing 3D content with single views. 
 Analogously, recent works~\cite{jain2022zero,dreamfusion} try to generate 3D content from text inputs directly. Although they also rely on the text prompt as guidance, their rendered objects are only conditioned on text sentences without image constraints, which can not reach our goal. Different from them, we aim at lifting a single image (either real or synthetic) to photorealistic 3D objects.

\begin{figure*}
    \centering
    \includegraphics[width=0.8\textwidth]{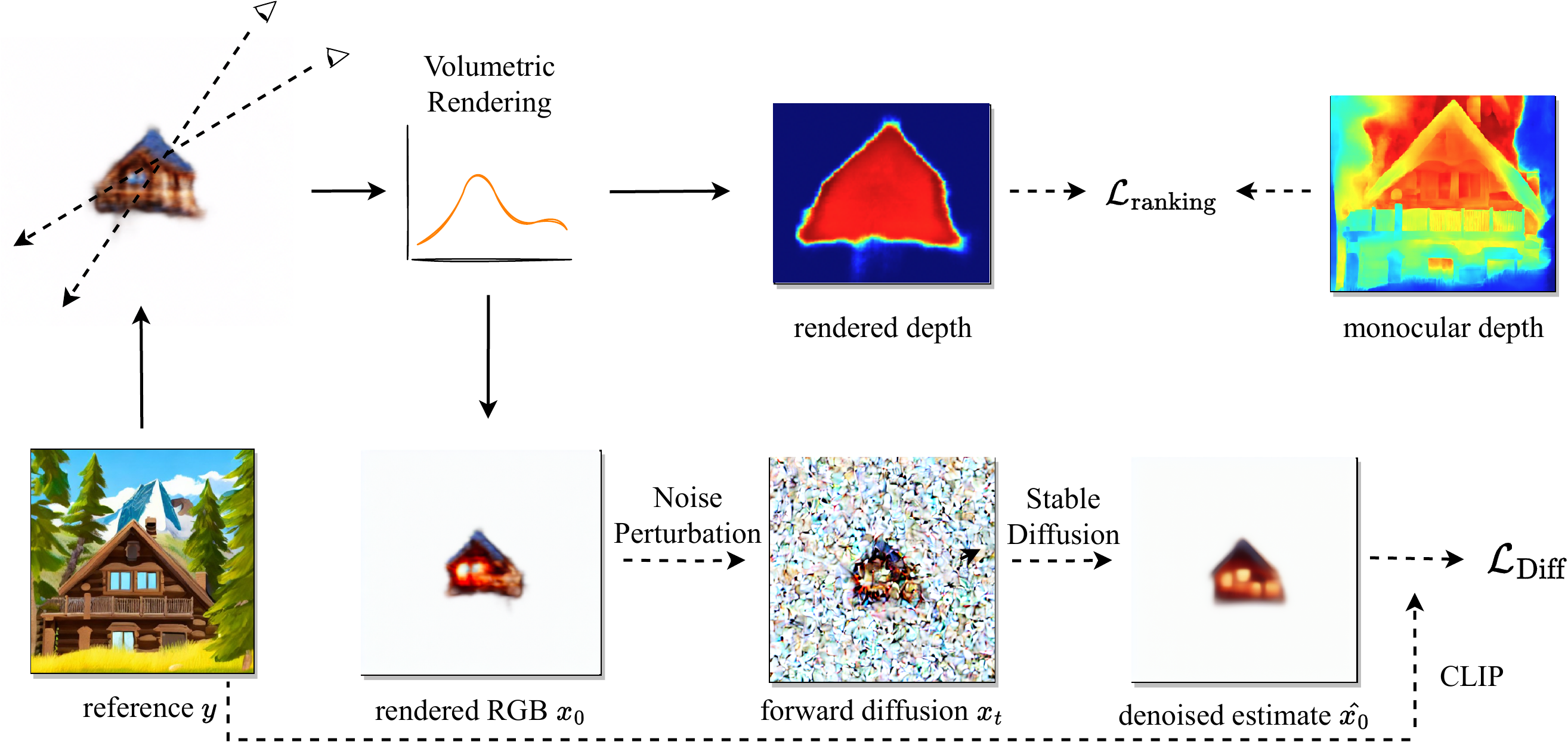}
    \caption{Overview of the proposed pipeline. NeuralLift-360 learns to recover a 3D object from a single reference image $y$. We derive a prior distillation loss $\mathcal{L}_\text{diff}$ for CLIP-guided diffusion prior and utilize rough depth estimation with the help of a ranking loss $\mathcal{L}_\text{ranking}$.}
    \label{fig:framework}
    \vspace{-4mm}
\end{figure*}

\section{Preliminaries}

\paragraph{Neural Radiance Field} NeRFs~\cite{nerf} render images by sampling 5D coordinates
(location $(x, y, z)$ and viewing direction $(\theta, \phi)$) along camera rays. After mapping them to color $(r, g, b)$ and volume density $\sigma$, additional volumetric rendering techniques are adopted to fuse the information along the ray.
Given $\Mat{o}$ as the camera origin and $\Mat{d}$ as the ray direction, 3D point locations along the ray can be expressed as  $\Mat{r}(t) = \Mat{o}+t\Mat{d}$. The predicted color accumulated is defined as follows:
\begin{equation}
\Mat{C}(\Mat{r})=\int_{t_{n}}^{t_{f}} T(t) \sigma(\Mat{r}(t)) \Mat{c}(\Mat{r}(t), \Mat{d}) d t,
\label{eq:nerf}
\end{equation}
where $T(t)=\exp \left(-\int_{t_{n}}^{t} \sigma(\Mat{r}(s)) d s\right)$, $\sigma(\cdot)$ and $\Mat{c}(\cdot, \cdot)$ are densities and color predictions. To numerically evaluate Eq. \ref{eq:nerf}, the continuous integral is estimated using quadrature~\cite{nerf}.
Our model is built upon Instant-NGP~\cite{muller2022instant}, which reduces computation cost by inferring color and density from a multiresolution hash table implemented by CUDA.

\paragraph{Denoising Diffusion Model}
Denoising diffusion models generate images by gradually denoising from a gaussian noise $p\left(\mathbf{x}_T\right)=\mathcal{N}(\mathbf{0}, \mathbf{I})$ and transforming into a certain data distribution. The forward diffusion process $q(\Mat{x}_{t} | \Mat{x}_{t-1})$ adds Gaussian noise to the image $\Mat{x}_t$. The marginal distribution can be written as: $q\left(\Mat{x}_t \mid \Mat{x_0}\right)=\mathcal{N}\left(\alpha_t \Mat{x_0}, \sigma_t^2 \Mat{I}\right)$, where $\alpha_t$ and $\sigma_t$ are designed to converge to $\mathcal{N}(\Mat{0}, \Mat{I})$ when $t$ is at the end of the forward process~\cite{kingma2021variational,song2020score}. The reverse diffusion process $p(\Mat{x}_{t-1} | \Mat{x}_t)$ learns to denoise. Given infinitesimal timesteps, the reverse diffusion process can be approximated with Gaussian~\cite{song2020score} related with an optimal MSE denoiser~\cite{sohl2015deep}. The diffusion models are designed as noise estimators $\Mat{\epsilon}_\theta(\Mat{x}_t, t)$ taking noisy images as input and estimating the noise. They are trained via optimizing the weighted evidence lower bound (ELBO)~\cite{ho2020denoising,kingma2021variational}:
\begin{equation}
\mathcal{L}_{\text{ELBO}}(\theta)= \mean\left[w(t)\left\|\Mat{\epsilon}_\theta\left(\alpha_t \Mat{x}_0+\sigma_t \Mat{\epsilon} ; t\right)-\Mat{\epsilon}\right\|_2^2\right],
\label{eq:ddpm}
\end{equation}
where $\Mat{\epsilon} \sim \mathcal{N}(\mathbf{0}, \mathbf{I})$, $w(t)$ is a weighting function. In practice, setting $w(t)=1$ delivers good performance~\cite{ho2020denoising}.
Sampling from a diffusion model can be either stochastic \cite{ho2020denoising} or deterministic \cite{song2020denoising}. After sampling $\Mat{x}_T \sim \mathcal{N}(\mathbf{0}, \mathbf{I})$,  we can gradually reduce the noise level and reach a clean image with high quality at the end of the iterative process.

\section{Method}

\paragraph{Overview.} In this section, we propose our method, dubbed NeuralLift-360, which reconstructs a 3D object from a single image.
An overview of NeuralLift-360 is illustrated in Fig. \ref{fig:framework}.
NeuralLift-360 combines the best of the two worlds: NeRF and diffusion model, where the former is leveraged to universally model a 3D scene and the latter is utilized to hallucinate the unseen views.
To inject knowledge of a diffusion model into a radiance field, we derive a training framework in Sec. \ref{sec:prob_model}, whose supervision consists of two parts: (1) a prior distillation loss from a generative model (Sec. \ref{sec:diff_prior}), (2) a straight loss on the given image (Sec. \ref{sec:rough_depth}).
Moreover, we introduce a domain adaption technique to adapt the diffusion prior to in-the-wild images (Sec. \ref{sec:ft}).

\subsection{Probabilistic-Driven 3D Lifting} \label{sec:prob_model}

We begin by formulating our setting along the way, introducing our notations.
Given an image $\Mat{y} \in \real^N$ and its text description $\Mat{z} \in \real^D$, NeuralLift-360 intends to reconstruct a 3D scene $\Mat{V} \in \real^M$, where $\Mat{V}$ can be either a radiance volume or the network parameters of a NeRF.
We regard $\Mat{y}$, $\Mat{z}$ and $\Mat{V}$ as random variables.
To reconstruct $\Mat{V}$ from $\Mat{y}$ and $\Mat{z}$, we maximize a log-posterior and apply Bayesian rule:
\begin{align} \label{eqn:max_posterior}
\log p(\Mat{V} | \Mat{y}, \Mat{z}) = \log p(\Mat{y} | \Mat{V}, \Mat{z}) + \log p(\Mat{V} | \Mat{z}) + {const.},
\end{align}
One can see the prior term $p(\Mat{V} | \Mat{z})$ provides complementary information to lift an image into 3D space.
However, specifying this prior can be intractable, and directly optimizing objective Eq. \ref{eqn:max_posterior} is also sampling inefficient.
Instead, we introduce camera pose $\Mat{\Phi} \in \mathbb{SO}(3) \times \real^3$ as a latent variable, then the likelihood term can be rewritten as:
\begin{align}
p(\Mat{y} | \Mat{V}, \Mat{z}) &= \int p(\Mat{y} | \Mat{V}, \Mat{\Phi}, \Mat{z}) p(\Mat{\Phi} | \Mat{V}, \Mat{z}) d\Mat{\Phi} \\
\label{eqn:likelihood} & = \int p(\Mat{y} | \Mat{V}, \Mat{\Phi}, \Mat{z}) \frac{p(\Mat{V} | \Mat{\Phi}, \Mat{z}) p(\Mat{\Phi})}{p(\Mat{V} | \Mat{z})} d\Mat{\Phi},
\end{align}
where we assume $\Mat{\Phi}$ is independent of $\Mat{z}$. Merging Eq. \ref{eqn:max_posterior} and Eq. \ref{eqn:likelihood}, we can get rid of the implicit prior term and obtain a sampling efficient ELBO of the original objective as below:
\begin{align}
&\log \mean_{\Mat{\Phi} \sim p(\Mat{\Phi})} p(\Mat{y} | \Mat{V}, \Mat{\Phi}, \Mat{z}) p(\Mat{V} | \Mat{\Phi}, \Mat{z}) \\
&\ge \mean_{\Mat{\Phi} \sim p(\Mat{\Phi})} \left[ \log p(\Mat{y} | \Mat{V}, \Mat{\Phi}, \Mat{z}) + \log p(\Mat{V} | \Mat{\Phi}, \Mat{z}) \right].
\end{align}
Furthermore, we define $p(\Mat{y} | \Mat{V}, \Mat{\Phi}, \Mat{z}) = p(\Mat{y} | h(\Mat{V}, \Mat{\Phi}), \Mat{z})$ and $p(\Mat{V} | \Mat{\Phi}, \Mat{z}) = p(h(\Mat{V}, \Mat{\Phi}) | \Mat{z})$, where $h(\Mat{V}, \Mat{\Phi})$ renders the 3D scene with respect to the camera pose $\Mat{\Phi}$. Then we derive our final training objective for NeuralLift-360:
\begin{align} \label{eqn:prob_objective}
\mathcal{L} = -\mean_{\Mat{\Phi}} \left[ \underbrace{\log p(\Mat{y} | h(\Mat{V}, \Mat{\Phi}), \Mat{z})}_{\text{referenced loss}} + \underbrace{\log p(h(\Mat{V}, \Mat{\Phi}) | \Mat{z})}_{\text{non-referenced loss}} \right].
\end{align}
At each training step, we sample a camera pose with respect to a distribution (more on Sec. \ref{sec:regularizer}), and then render the view associated with this camera parameter (via volume rendering Eq. \ref{eq:nerf}).
The loss computed on the rendered image consists of two parts: image-referenced guidance and reference-free regularization.
The referenced loss will enforce $\Mat{V}$ to have a close appearance with given $\Mat{y}$, while the non-referenced term hallucinates the image information with prior knowledge.
In the next section, we introduce a diffusion model to implement our loss function Eq. \ref{eqn:prob_objective}.

\subsection{CLIP-guided Diffusion Priors}
\label{sec:diff_prior}
Diffusion models~\cite{ho2020denoising, song2019generative, song2020denoising, song2020score}, which synthesize images by gradually denoising from a gaussian noise, have undeniably been successful in the realm of image generation, editing, and processing tasks.
When trained on large-scale datasets for text-to-image tasks, these models can accurately represent the distribution of images based on textual descriptions~\cite{ramesh2021zero,ramesh2022hierarchical,ldm}.
Consequently, we are inspired to utilize a diffusion model's generative prior to complement the 3D information.
To accomplish this, we re-parameterize our Eq. \ref{eqn:prob_objective} with a diffusion model.
Nevertheless, the derived loss function contains both discriminative and generative terms.
Taking inspiration from classifier guidance techniques~\cite{lecun2006tutorial, song2020score,liu2019more}, we show that we can leverage the training loss of a diffusion model to surrogate the generative part and utilize CLIP space similarity to model the discriminative part.

Following the derivation in \cite{ho2020denoising} and \cite{luo2022understanding}, we lower bound Eq. \ref{eqn:prob_objective} by:
\begin{align}
\nonumber &\log p_{\theta}(\Mat{y} | \Mat{x}, \Mat{z}) + \log p_{\theta}(\Mat{x} | \Mat{z}) \\
\nonumber &\ge -\mean_{t, \Mat{\epsilon}\sim\mathcal{N}(\Mat{0}, \Mat{I})} \left[ \KL(q(\Mat{x}_{t-1} | \Mat{x}_t, \Mat{x}_0) \Vert p_{\theta}(\Mat{x}_{t-1} | \Mat{x}_t, \Mat{y}, \Mat{z}) \right] \\
\label{eqn:diff_ELBO} &= -\mean_{t, \Mat{\epsilon}\sim\mathcal{N}(\Mat{0}, \Mat{I})} \left[ w(t) \left\lVert \nabla \log p_{\theta}(\Mat{x}_{t} | \Mat{y}, \Mat{z}) - \Mat{\epsilon} \right\rVert_2^2 \right],
\end{align}
where $\Mat{x}_t = \alpha_t \Mat{x}_0 + \sigma_t \Mat{\epsilon}$ and $w(t)$ is a constant dependent on noise scale $\alpha_t$ and $\sigma_t$.
The expectation of $t \sim \mathcal{U}\{1, \cdots, T\}$ is usually taken over a uniform distribution.
We provide detailed derivation in supplementary.
At first glance, we may directly plug in the pre-trained denoiser $\Mat{\epsilon}_{\theta}(\Mat{x}_t; \Mat{z})$ of a diffusion model to surrogate $\nabla p_{\theta}(\Mat{x}_{t} | \Mat{y}, \Mat{z})$ similar to \cite{dreamfusion}. However, note that $p_{\theta}(\Mat{x}_{t} | \Mat{y}, \Mat{z})$ is also conditioned on the reference image $\Mat{y}$, which requires further efforts to inject the reference image information into the diffusion process.

To this end, NeuralLift-360 leverages a Mixed Reference Guidance that guides the diffusion process with a mixture of references.
Similar to classifier guidance techniques \cite{song2020score, lecun2006tutorial, luo2022understanding,liu2019more}, we rewrite $p_{\theta}(\Mat{x}_{t} | \Mat{y}, \Mat{z}) = p_{\theta}(\Mat{y} | \Mat{x}_t, \Mat{z}) p_{\theta}(\Mat{x}_t | \Mat{z})/p_{\theta}(\Mat{y} | \Mat{z})$, and we adopt classifier-free guidance to model $p_{\theta}(\Mat{x}_t | \Mat{z})$, then we can obtain a modified score function \cite{song2019generative} as below:
\begin{align} \label{eqn:score}
\nonumber \Mat{\epsilon}_{\theta}(\Mat{x}_{t} ; \Mat{y}, \Mat{z}) &= (1 + w) \Mat{\epsilon}_{\theta}(\Mat{x}_{t}; \Mat{z}) - w \Mat{\epsilon}_{\theta}(\Mat{x}_{t}) \\
& + \sqrt{1-{\alpha}_t} \nabla \log p_{\theta}(\Mat{y} | \Mat{x}_t, \Mat{z}),
\end{align}
where $w$ is a hyper-parameter that controls the strength of the classifier guidance. We further measure the discrimination with a distance metric on the feature space $p_{\theta}(\Mat{y} | \Mat{x}_t, \Mat{z}) \propto \exp\left[-\phi \left(\frac{\Mat{x}_t - \sigma_t \Mat{\epsilon}_{\theta}(\Mat{x}_t; \Mat{z})}{\alpha_t}, \Mat{y}\right)\right]$.
Specifically, we choose the inner product on the CLIP embedding space \cite{radford2021learning} as the distance metric, i.e., $\log p_{\theta}(\Mat{y} | \Mat{x}_t, \Mat{z}) \propto -\left\langle F\left(\frac{\Mat{x}_t - \sigma_t \Mat{\epsilon}_{\theta}(\Mat{x}_t)}{\alpha_t}\right), F(\Mat{y}) \right\rangle$, which spurs on semantic similarity between denoised rendered image and reference image $\Mat{y}$, where $F$ is the CLIP image encoder.

In each training iteration, we alternate two camera pose sampling strategies: (1) we fix a camera pose $\Mat{\Phi}_0$ (i.e., a Delta distribution) as the reference view, and when rendering on that view, we only penalize the RGB differences by modeling $p(\Mat{y} | h(\Mat{V}, \Mat{\Phi}), \Mat{z}) = \mathcal{N}(\Mat{y} | h(\Mat{V}, \Mat{\Phi}), \sigma^2 \Mat{I})$, (2) we stochastically sample camera poses, and adopt the diffusion ELBO (Eq. \ref{eqn:diff_ELBO}) to estimate the evidence of rendered image together with prior $p_{\theta}(\Mat{x} | \Mat{z})$.
Combining all together, we can summarize our training loss as below:
\begin{align} \label{eqn:final_loss}
&\mathcal{L} = \frac{1}{\sigma^2} \underbrace{\left\lVert h(\Mat{V}, \Mat{\Phi}_0) - \Mat{y} \right\rVert_2^2}_{\mathcal{L}_{\text{photometric}}} \\
&+ \underbrace{\mean_{\Mat{\Phi}, t, \Mat{\epsilon}\sim\mathcal{N}(\Mat{0}, \Mat{I})} \left[ w(t) \left\lVert \Mat{\epsilon}_{\theta}(\alpha_t h(\Mat{V}, \Mat{\Phi}) + \sigma_t \Mat{\epsilon} | \Mat{y}, \Mat{z}) - \Mat{\epsilon} \right\rVert_2^2 \right]}_{\mathcal{L}_{\text{diff}}}. \nonumber
\end{align}
Compared with DreamFusion \cite{dreamfusion}, NeuralLift-360's training objective grounds the 3D scene with the given image which enables the generated scene to have consistent appearance with the specified object.

\begin{figure}
\centering
\begin{subfigure}{0.80\columnwidth}
\includegraphics[width=0.24\textwidth,height=1.5cm]{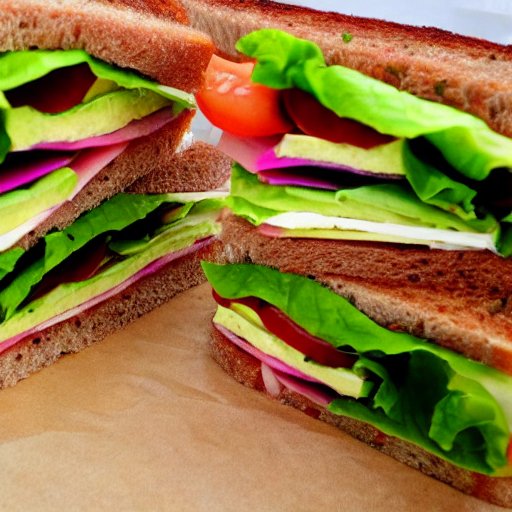}
\includegraphics[width=0.24\textwidth,height=1.5cm]{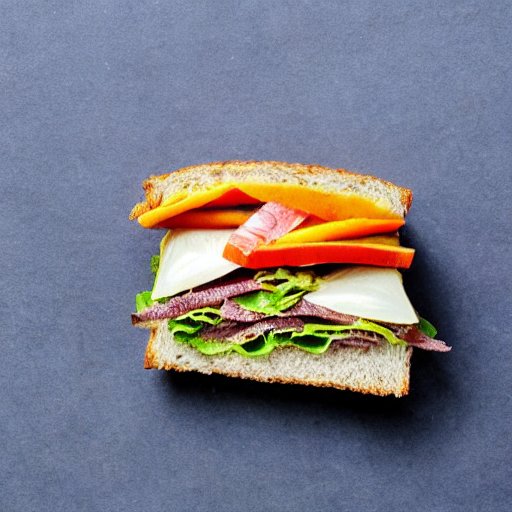}
\includegraphics[width=0.24\textwidth,height=1.5cm]{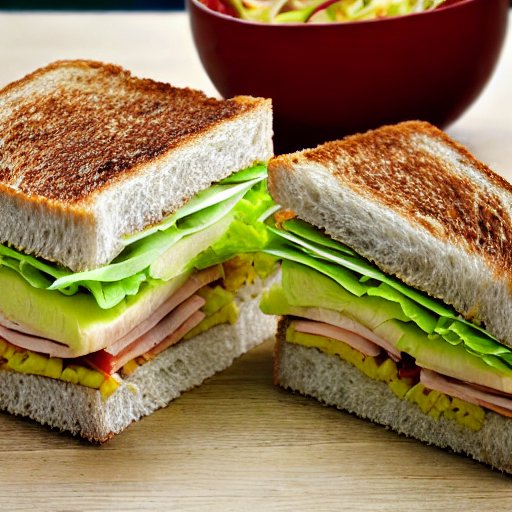}
\includegraphics[width=0.24\textwidth,height=1.5cm]{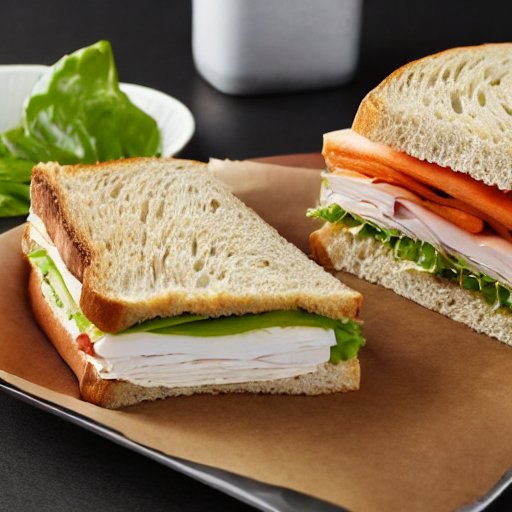}
\caption{Pre-trained}
\end{subfigure}
\begin{subfigure}{0.19\columnwidth}
\includegraphics[width=\textwidth,height=1.5cm]{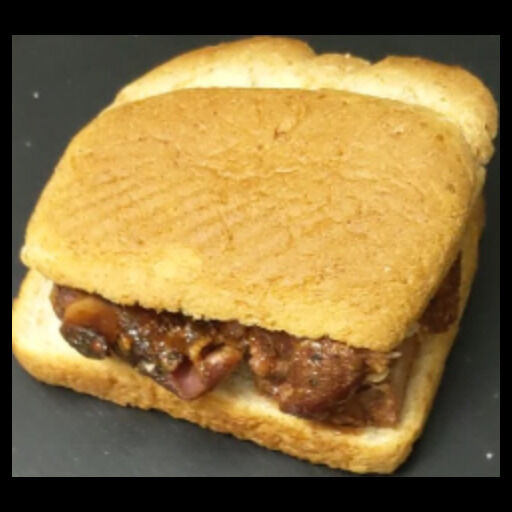}
\caption{Ref}
\end{subfigure}
\\
\begin{subfigure}{0.99\columnwidth}
\includegraphics[width=0.193\textwidth]{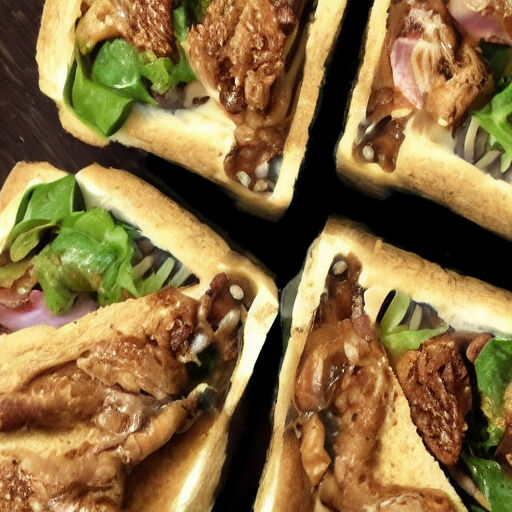}
\includegraphics[width=0.193\textwidth]{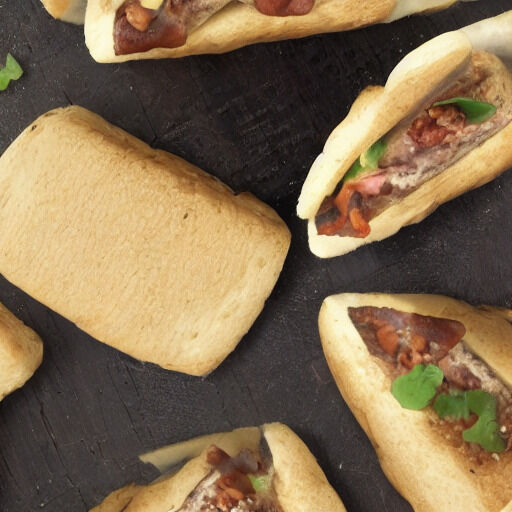}
\includegraphics[width=0.193\textwidth]{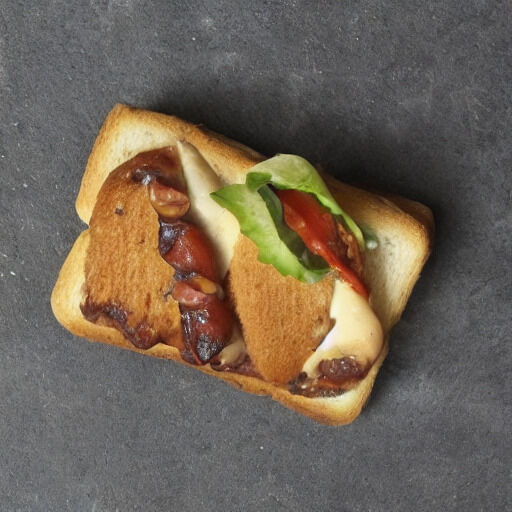}
\includegraphics[width=0.193\textwidth]{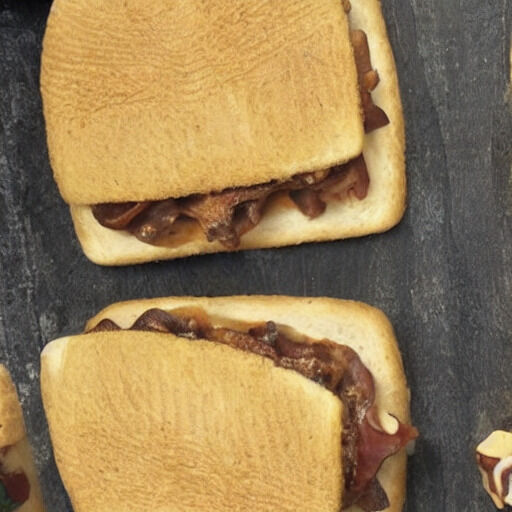}
\includegraphics[width=0.193\textwidth]{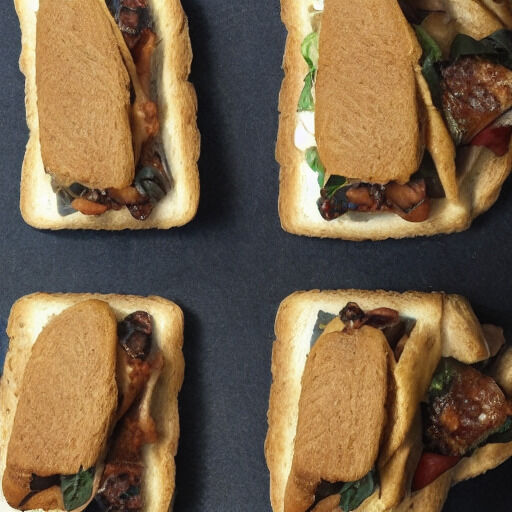}
\caption{Fine-tuned}
\end{subfigure}
\vspace{-2mm}
\caption{After fine-tuning, 
our model can generate diverse images similar to the reference image (b) while not identical.}
\vspace{-3mm}
\label{fig:ftsd}
\end{figure}

\subsection{Domain Adaption to In-the-wild Images}
\label{sec:ft}

A primary challenge, however, lies in assigning an appropriate text prompt for our reference image. Given that large diffusion models are predominantly trained for text-to-image tasks, they typically require a text embedding as conditional input.
For images in the wild, it is generally hard to exactly describe an image since images contain more fine-grained information than texts.

Adapting to in-the-wild images necessitates finetuning the Stable Diffusion model on the reference image
Textual inversion~\cite{gal2022image} and Dreambooth~\cite{ruiz2022dreambooth} enable finetuning the Stable Diffusion model on a limited number of images for a subject-driven generation. However, in our situation, we only have access to a single image, which significantly hampers the effectiveness of these methods. Imagic~\cite{kawar2022imagic} and UniTune~\cite{valevski2022unitune} support finetuning on a single image, but their primary objective is to perform real image editing, which can be quite successful even if the diffusion model overfits the given image. Nevertheless, in our context, we need the diffusion model to generate images resembling the provided reference but not identical to it. If the supervision for each view is the same, NeRF will otherwise produce an isotropic surface.

We draw inspiration from Pivotal Tuning~\cite{roich2022pivotal} and Imagic~\cite{kawar2022imagic} when fine-tuning Stable Diffusion~\cite{ldm} on our own. To start with, we optimize the CLIP text embedding $\Mat{z}$ so that it fits our reference image $\Mat{y}$ better.
During finetuning, we use Eq.~\ref{eq:ddpm} to supervise the noise estimate. As we expect that our denoised result will resemble the reference image, we also apply a CLIP image loss to the estimated denoised image. $\widehat{\Mat{x}}_0$,
\begin{equation}
\begin{aligned}
\mathcal{L}_\text{finetune} &=\mathbb{E}_{t, \boldsymbol{\epsilon}}\left[\left\|\boldsymbol{\epsilon}-\Mat{\epsilon}_{\theta}\left(\mathbf{x}_t; \Mat{z}\right)\right\|_2^2\right]\\
&-\mathbb{E}_{t, \boldsymbol{\epsilon}}\left\langle F\left(\frac{\Mat{x}_t - \sigma_t \Mat{\epsilon}_{\theta}(\Mat{x}_t; \Mat{z})}{\alpha_t}\right), F(\Mat{y}) \right\rangle,
\end{aligned}
\end{equation}
where $F(\cdot)$ is the pre-trained CLIP image encoder.

\looseness=-1
Then, we fix the optimized text embedding $\Mat{z}^{*}$ and fine-tune the diffusion model $\boldsymbol{\epsilon}_\theta(\cdot)$ using the same loss function as above.
To preserve the generalization ability of the diffusion model, we introduce several augmentations during the finetuning. Specifically, in each iteration, we perform random image-level augmentations such as flip, transpose, shift, scale, and rotate. As shown in Fig.~\ref{fig:ftsd}, we are able to overcome the overfitting issue and preserve the diffusion model's ability to generate diverse similar but not identical contents.
Similar to Imagic~\cite{kawar2022imagic} and UniTune~\cite{roich2022pivotal}, we observe that after finetuning, the optimized text embedding will have an overfitting issue and it is critical to perform interpolation between the original text embedding and the optimized text embedding to preserve the diffusion model's ability to generate diverse contents. We obtain $\Mat{z}^\prime = \eta \Mat{z}^{*} + (1 - \eta) \Mat{z}$, where $\eta$ is a weighting hyper-parameter.

During sampling, we aim to amplify the conditional likelihood for our optimized embedding $\Mat{z}^{*}$, while decreasing both the unconditional likelihood and the conditional likelihood for the original embedding $\Mat{z}$. Hence for 50\% of all iterations, we use the original Eq.~\ref{eqn:score}, while for the others, we replace the unconditional sampling $\Mat{\epsilon}_{\theta}(\Mat{x}_{t})$ in Eq.~\ref{eqn:score} with conditional sampling using the original text embedding $\Mat{z}$:
\begin{equation}
\begin{aligned}
\widetilde{\Mat{\epsilon}}_{\theta}(\Mat{x}_{t} ; \Mat{y}, \Mat{z}^\prime, \Mat{z}) &= (1+w)\Mat{\epsilon}_{\theta}(\Mat{x}_{t}; \Mat{z}^\prime) \\
- w\Mat{\epsilon}_{\theta}(\Mat{x}_{t}; \Mat{z}) &+  \sqrt{1-{\alpha}_t} \nabla \log p_{\theta}(\Mat{y} | \Mat{x}_t, \Mat{z}^\prime),
\label{eq:promptcfg}
    \end{aligned}
\end{equation}

\subsection{Supervision from Rough Depth}
\label{sec:rough_depth}
While the aforementioned pipeline 
laid the foundation for the success of our method, we observe that obtaining a high-quality 3D object without geometry regularization is inherently challenging, if relying solely on learning priors.  As shown in SinNeRF~\cite{xu2022sinnerf} and DS-NeRF~\cite{deng2022depth}, accurate depth information can largely help improve the rendering quality. 
However, their impressive outcomes are contingent upon the 3D consistency of the accessible high-quality depth, a requirement that is not easily fulfilled in real-life scenarios.

In spite of the significant progress in monocular depth estimation for in-the-wild scenarios, current state-of-the-art depth estimators are insufficient for recovering 3D shapes. Due to unknown camera baselines and stereoscopic post-processing in data, these methods~\cite{lang2010nonlinear,wang2019web,chen2016single,yin2020diversedepth,ranftl2020towards,xian2018monocular} utilize loss functions invariant to shift and scale or adopt ranking loss that utilizes relative information. Therefore, their results cannot be used to reconstruct plausible 3D scene shapes directly due to unknown depth shifts~\cite{yin2021learning}. Though there are also attempts~\cite{yin2021learning} that improve off-the-shelf depth estimators, they generally can not generalize well to unseen objects in the wild.

Given that the acquired depth is imperfect, the absolute depth scale lacks the reliability necessary for recovering a high-quality 3D object. In contrast to previous depth-supervised NeRFs~\cite{deng2022depth,xu2022sinnerf}, which rely solely on precise values from the depth map, our approach represents the first attempt to maximize the utilization of depth information by adopting the relative ranking information.
Drawing inspiration from~\cite{chen2016single,pavlakos2018ordinal}, we adopt a pairwise depth ranking loss to supervise the rendered depth from our NeRF backbone.
In this way, we enforce the depth we rendered to be consistent with the depth from the reference view in terms of internal ranking instead of actual value.
The loss is formulated as follows,
\begin{equation}
\mathcal{L}_\text{ranking} = \begin{cases} \max \left(z_{j_k}-z_{i_k}, 0\right), & r_k \in\{>\} \\ \max\left(z_{i_k}-z_{j_k}, 0\right), & r_k \in\{<\}
\end{cases}
\end{equation}
where $r_k$ refers to the pairwise ranking relationship between pixel $z_{i_k}$ and $z_{j_k}$ in the monocular depth estimation. This loss function is used alongside the photo-metric loss and diffusion loss  mentioned in Eq.~\ref{eqn:final_loss} and serves as additional supervision on geometry.

\begin{figure*}
    \centering
   \begin{tabular}{P{0.12\textwidth}P{0.14\textwidth}P{0.2\textwidth}P{0.2\textwidth}P{0.18\textwidth}}
    \scriptsize Reference & \scriptsize DSNeRF~\cite{deng2022depth} & \scriptsize DietNeRF~\cite{jain2021putting} & \scriptsize SinNeRF~\cite{xu2022sinnerf} & \scriptsize \textbf{NeuralLift-360}\\
    \end{tabular}
\begin{subfigure}{0.1 \textwidth}
\includegraphics[width=\textwidth]{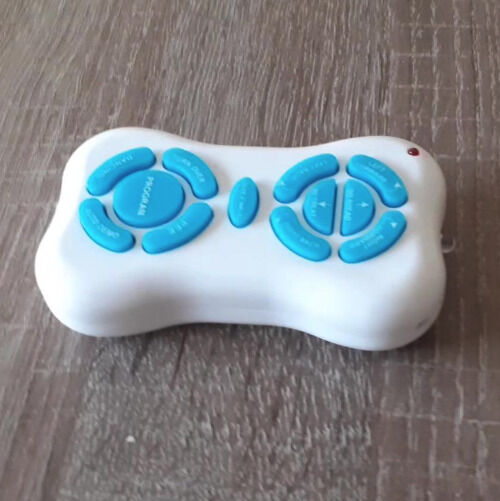}
\end{subfigure}
\begin{subfigure}{0.1 \textwidth}
\includegraphics[width=\textwidth]{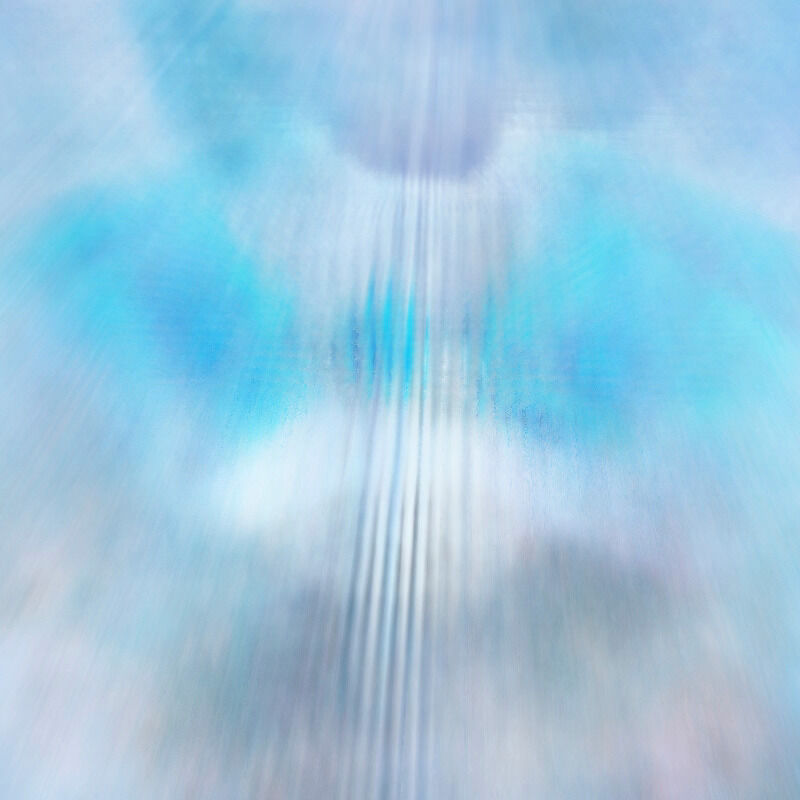}
\end{subfigure}
\begin{subfigure}{0.1 \textwidth}
\includegraphics[width=\textwidth]{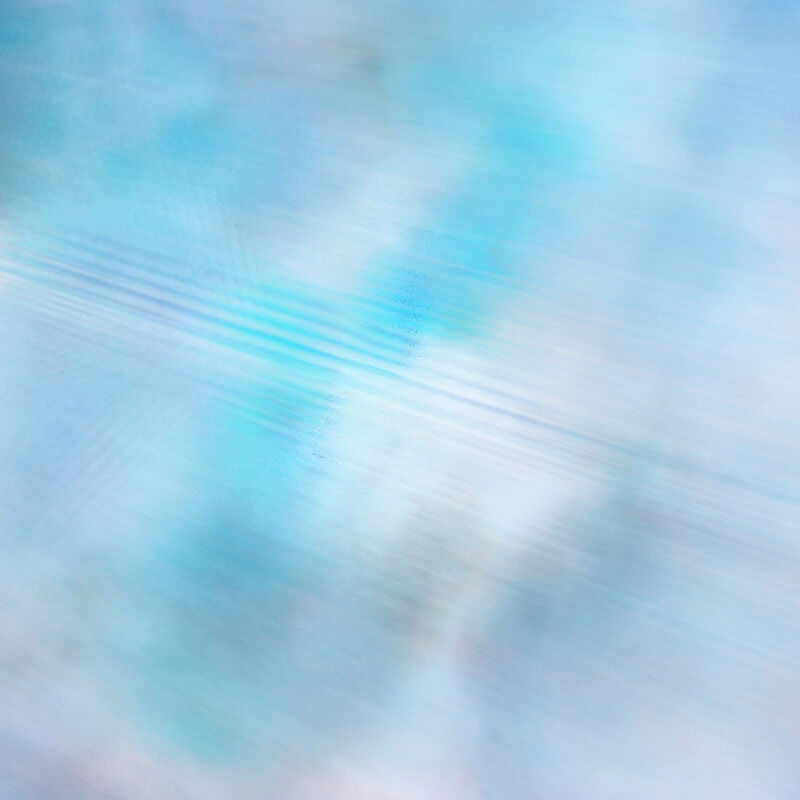}
\end{subfigure}
\begin{subfigure}{0.1 \textwidth}
\includegraphics[width=\textwidth]{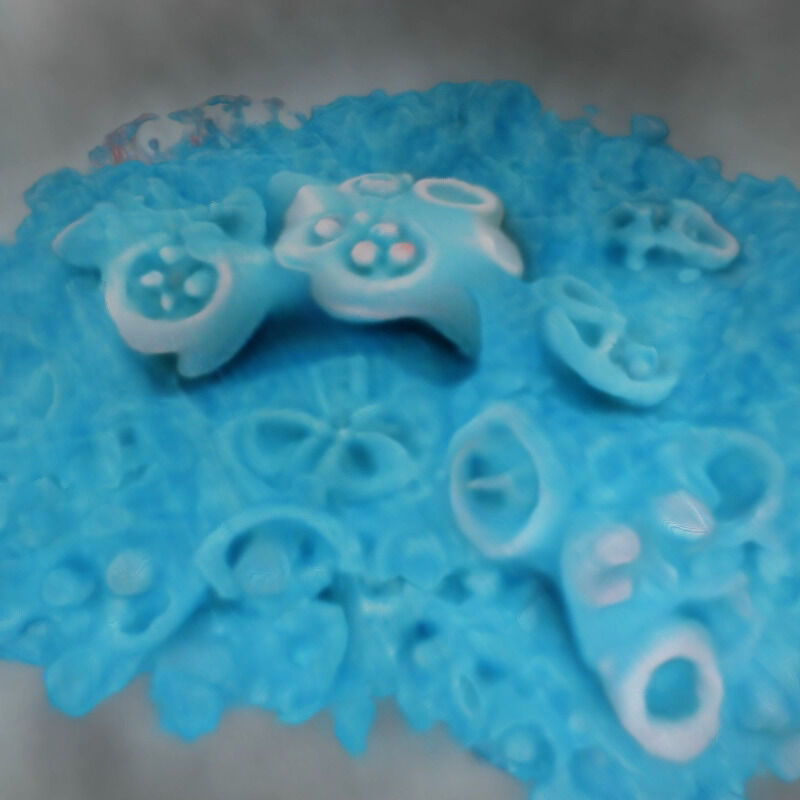}
\end{subfigure}
\begin{subfigure}{0.1 \textwidth}
\includegraphics[width=\textwidth]{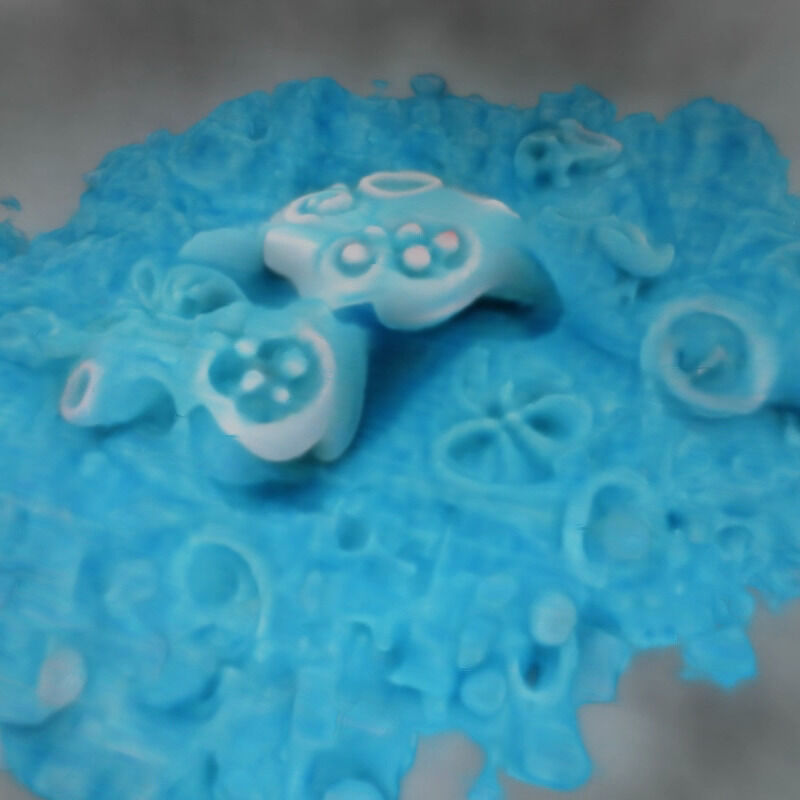}
\end{subfigure}
\begin{subfigure}{0.1 \textwidth}
\includegraphics[width=\textwidth]{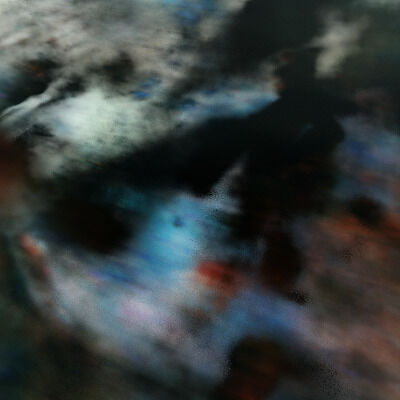}
\end{subfigure}
\begin{subfigure}{0.1 \textwidth}
\includegraphics[width=\textwidth]{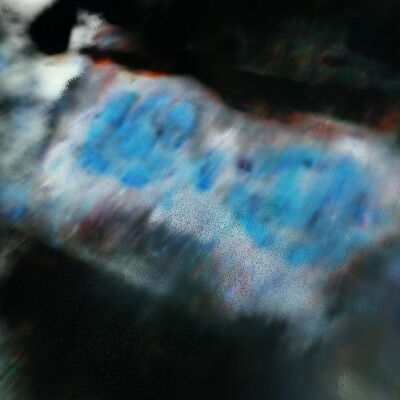}
\end{subfigure}
\begin{subfigure}{0.1 \textwidth}
\includegraphics[width=\textwidth]{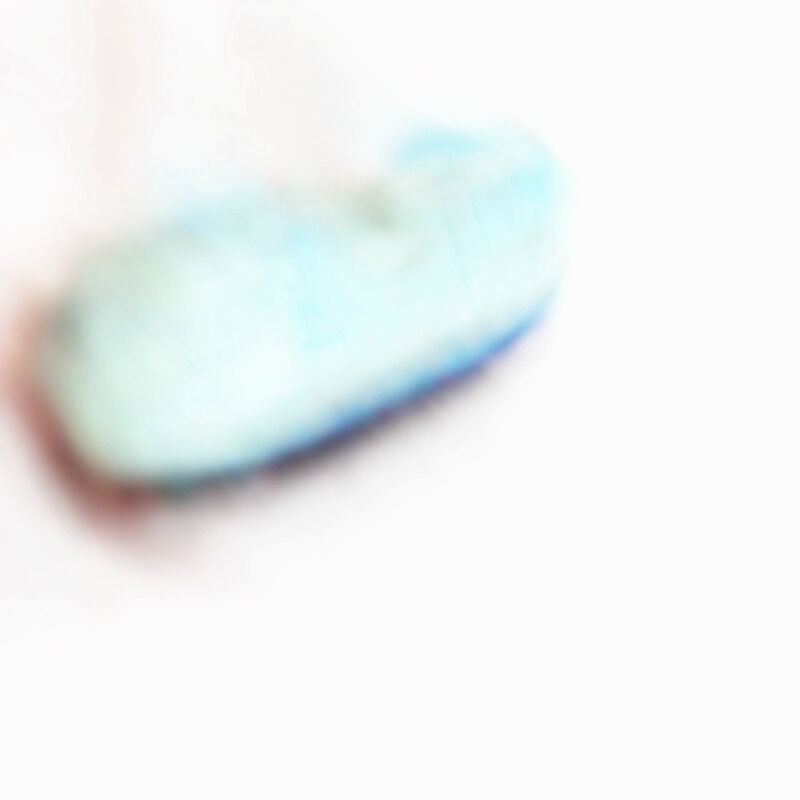}
\end{subfigure}
\begin{subfigure}{0.1 \textwidth}
\includegraphics[width=\textwidth]{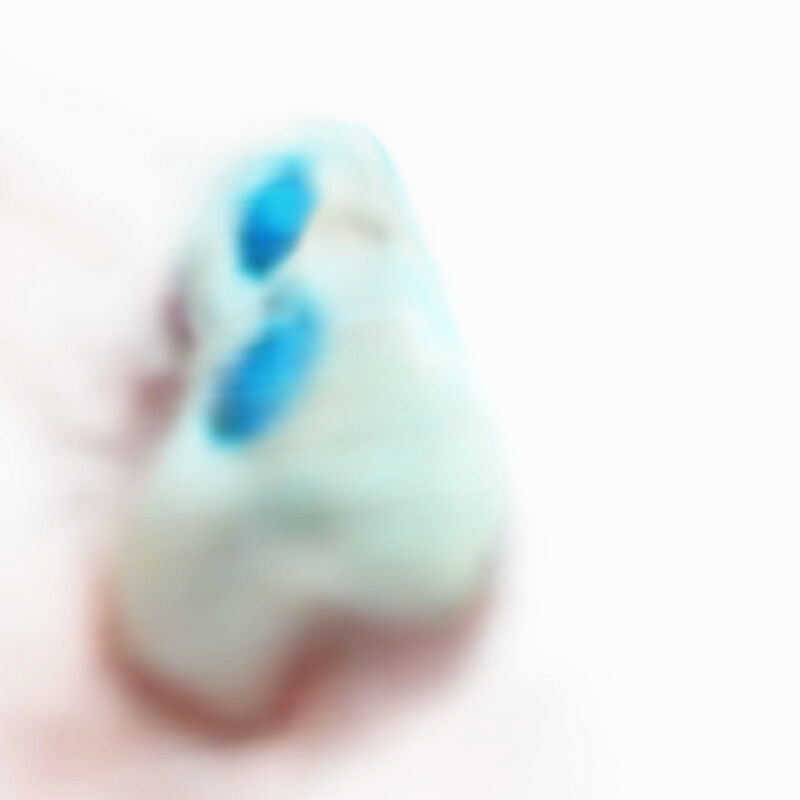}
\end{subfigure}
\\
\hspace*{.1\textwidth}
\begin{subfigure}{0.1 \textwidth}
\includegraphics[width=\textwidth]{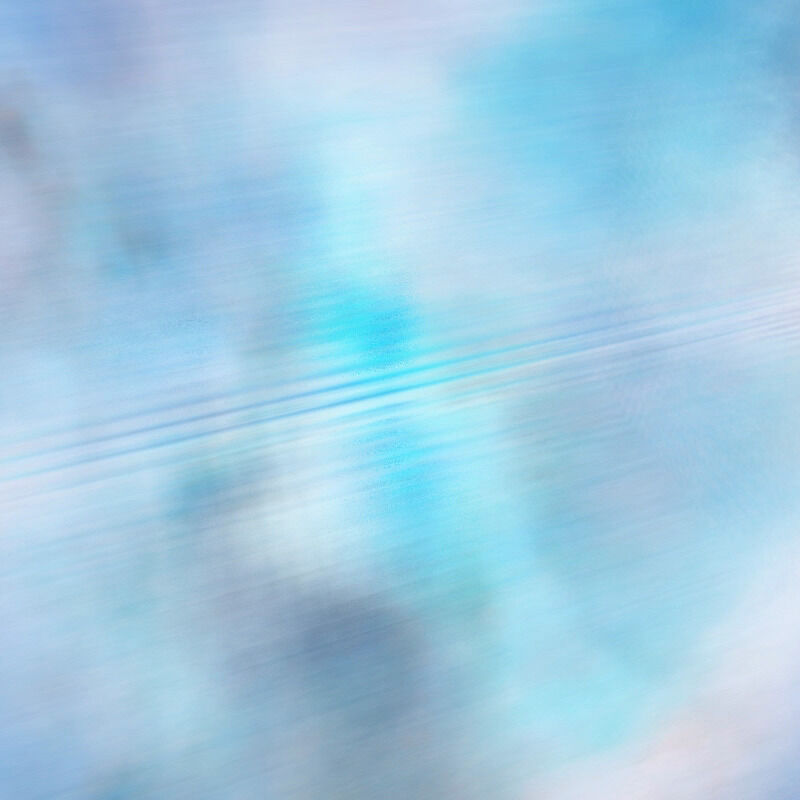}
\end{subfigure}
\begin{subfigure}{0.1 \textwidth}
\includegraphics[width=\textwidth]{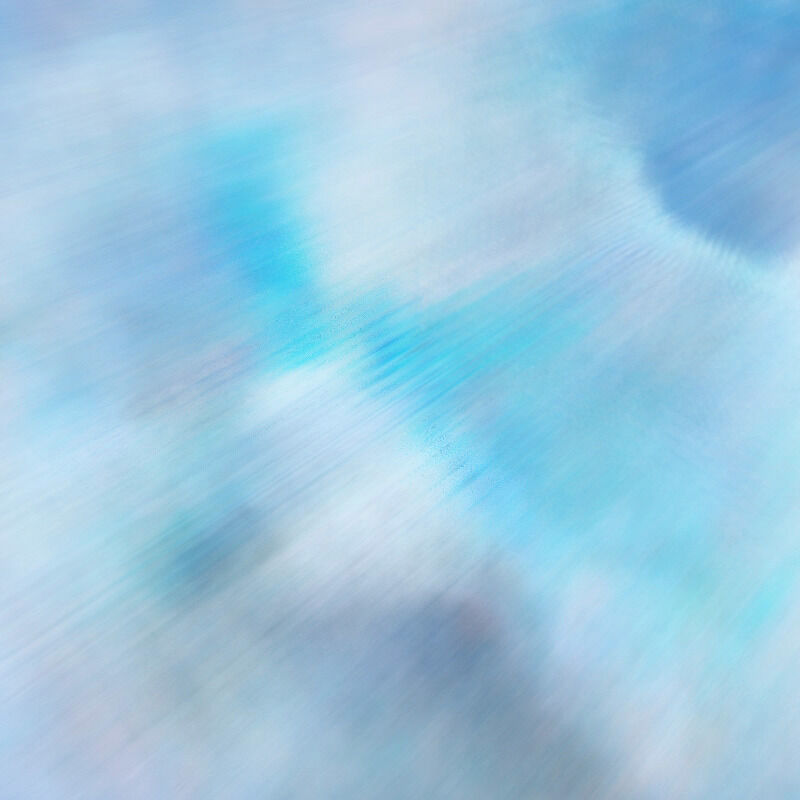}
\end{subfigure}
\begin{subfigure}{0.1 \textwidth}
\includegraphics[width=\textwidth]{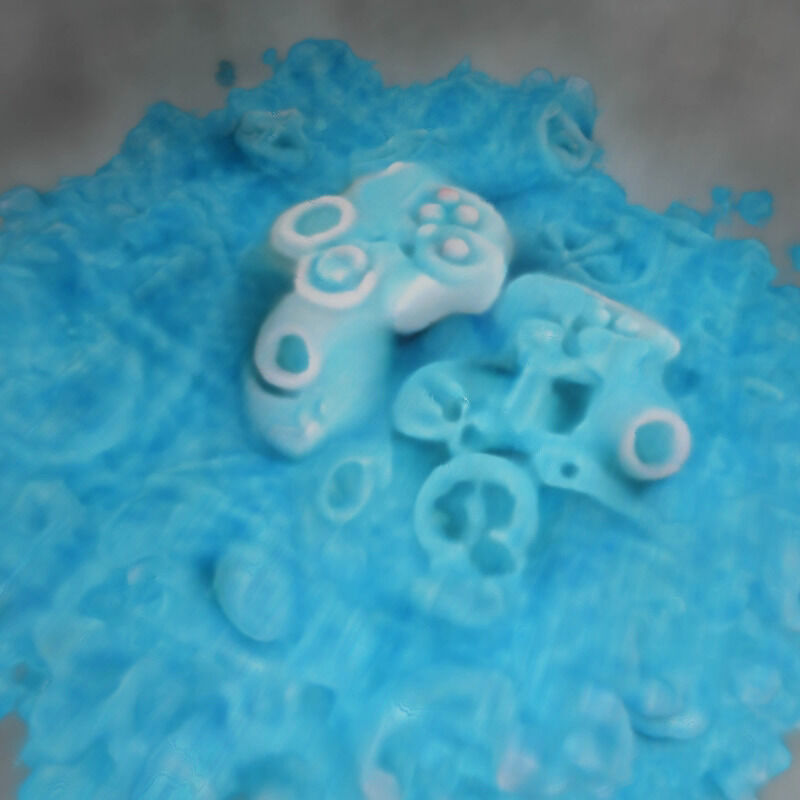}
\end{subfigure}
\begin{subfigure}{0.1 \textwidth}
\includegraphics[width=\textwidth]{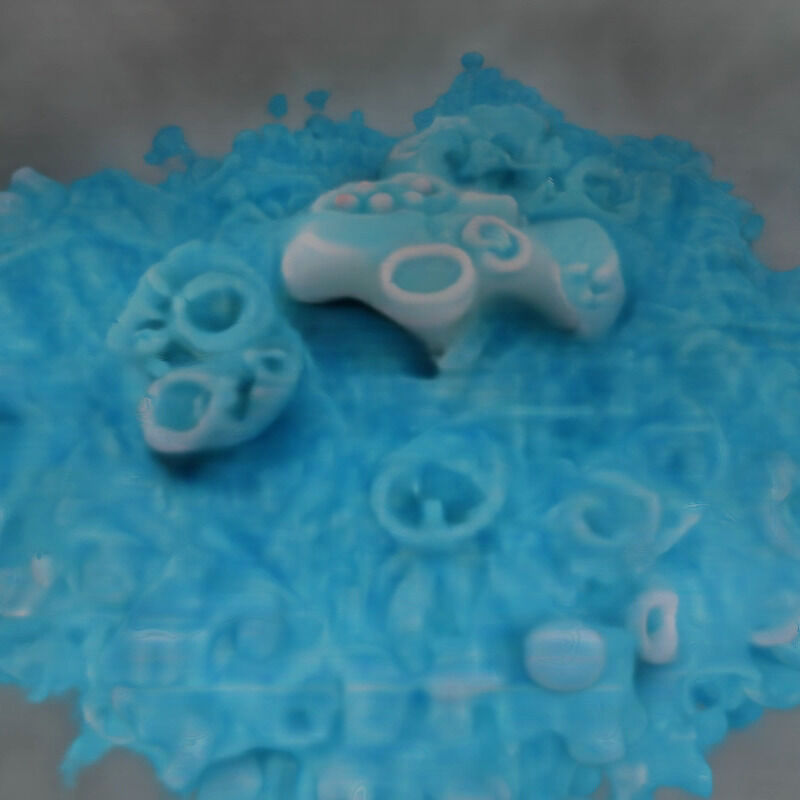}
\end{subfigure}
\begin{subfigure}{0.1 \textwidth}
\includegraphics[width=\textwidth]{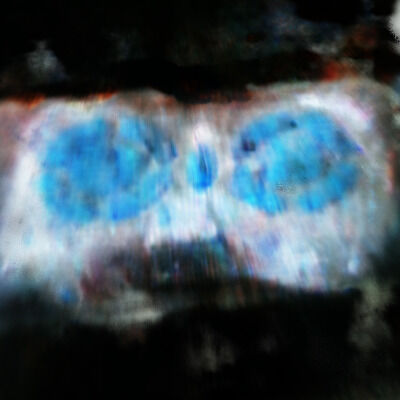}
\end{subfigure}
\begin{subfigure}{0.1 \textwidth}
\includegraphics[width=\textwidth]{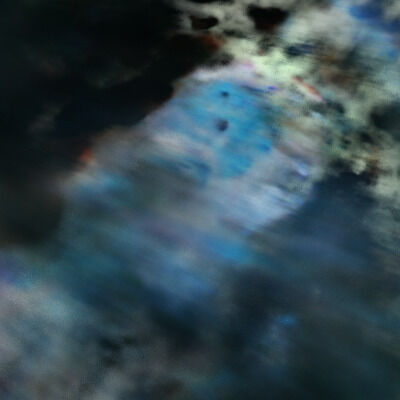}
\end{subfigure}
\begin{subfigure}{0.1 \textwidth}
\includegraphics[width=\textwidth]{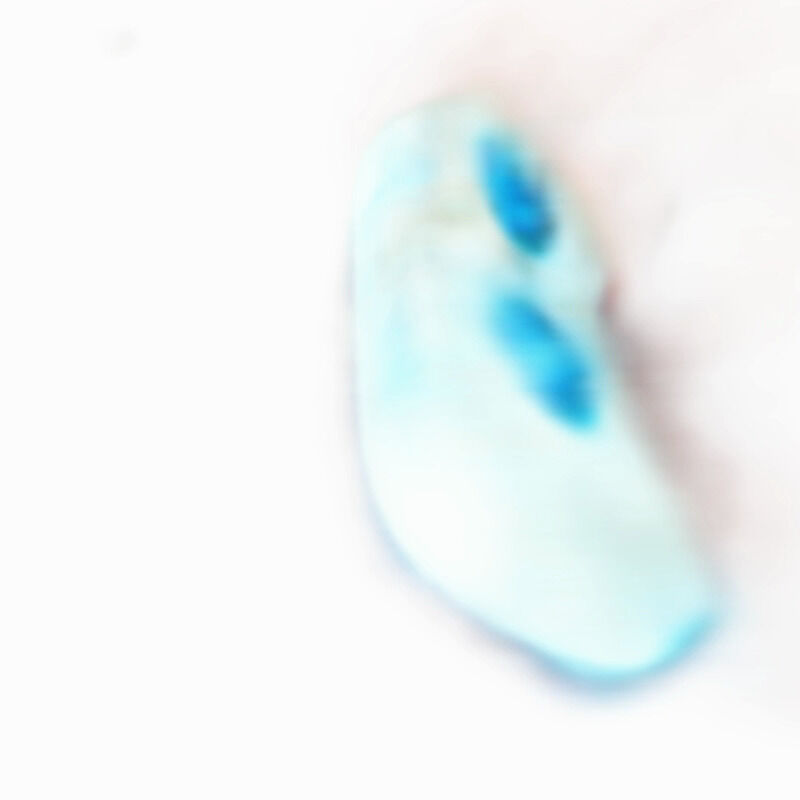}
\end{subfigure}
\begin{subfigure}{0.1 \textwidth}
\includegraphics[width=\textwidth]{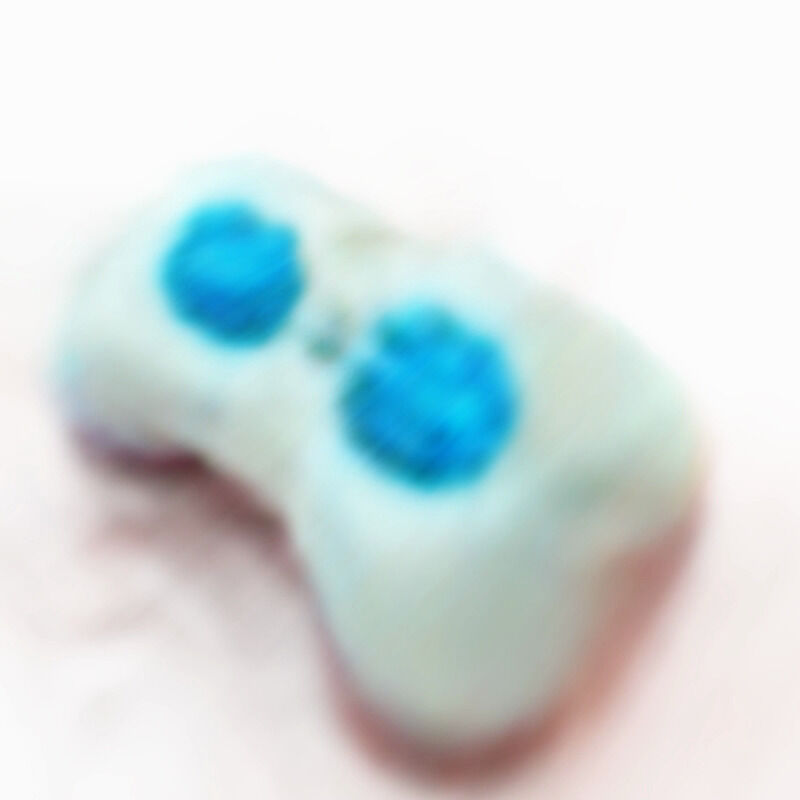}
\end{subfigure}
\begin{subfigure}{0.1 \textwidth}
\includegraphics[width=\textwidth]{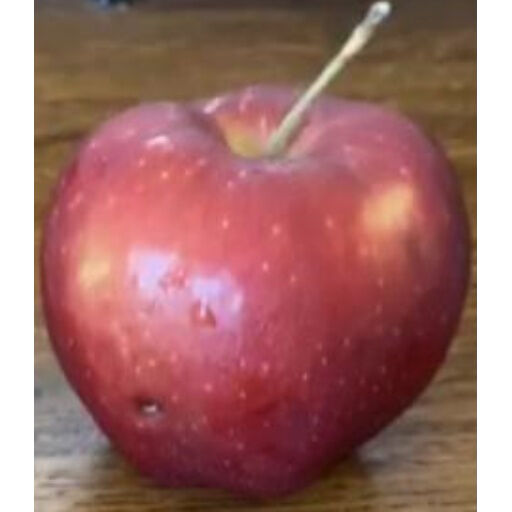}
\end{subfigure}
\begin{subfigure}{0.1 \textwidth}
\includegraphics[width=\textwidth]{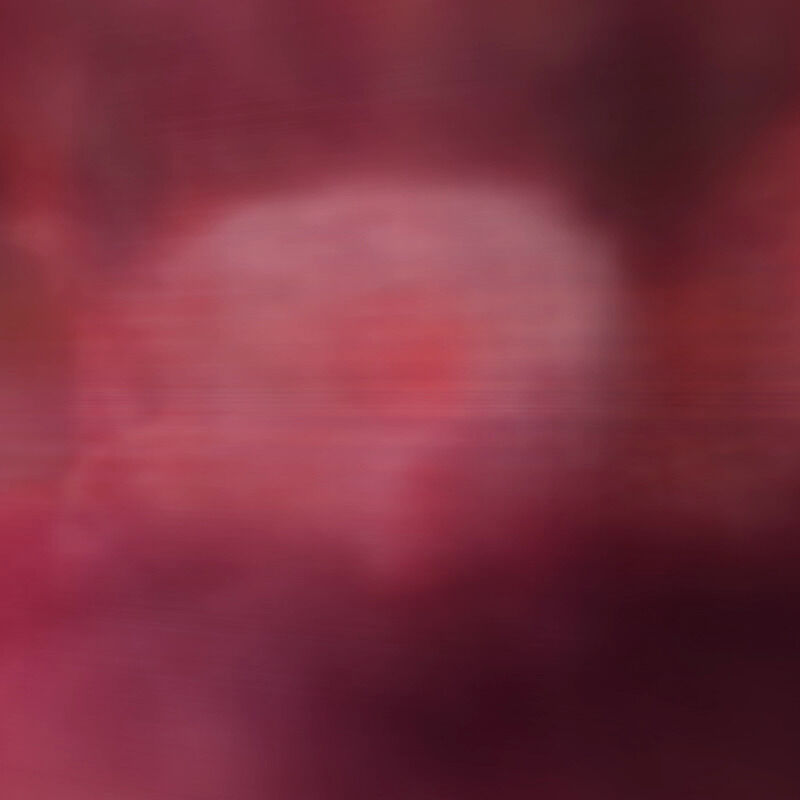}
\end{subfigure}
\begin{subfigure}{0.1 \textwidth}
\includegraphics[width=\textwidth]{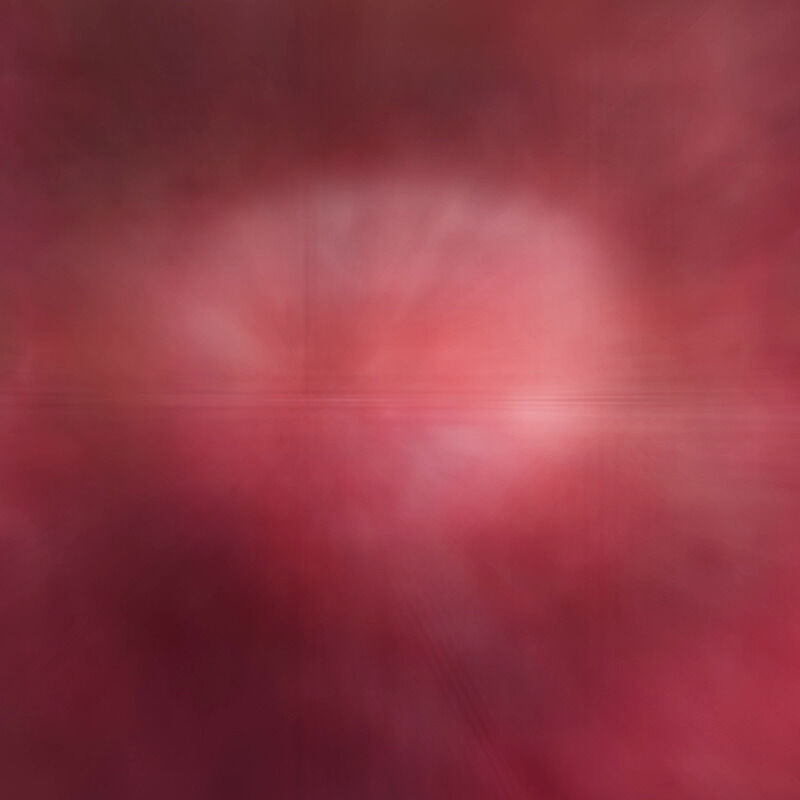}
\end{subfigure}
\begin{subfigure}{0.1 \textwidth}
\includegraphics[width=\textwidth]{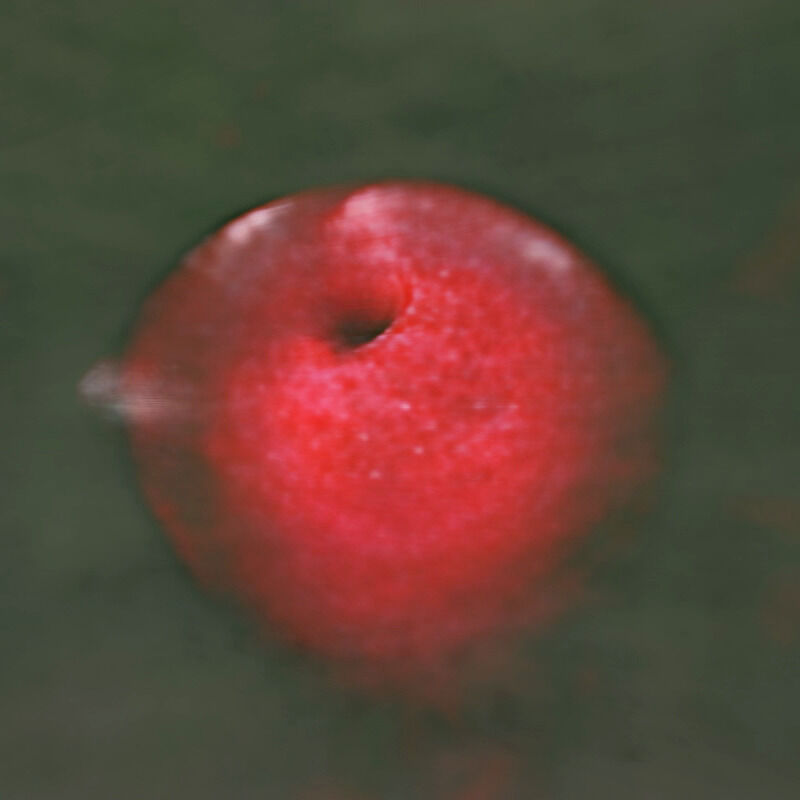}
\end{subfigure}
\begin{subfigure}{0.1 \textwidth}
\includegraphics[width=\textwidth]{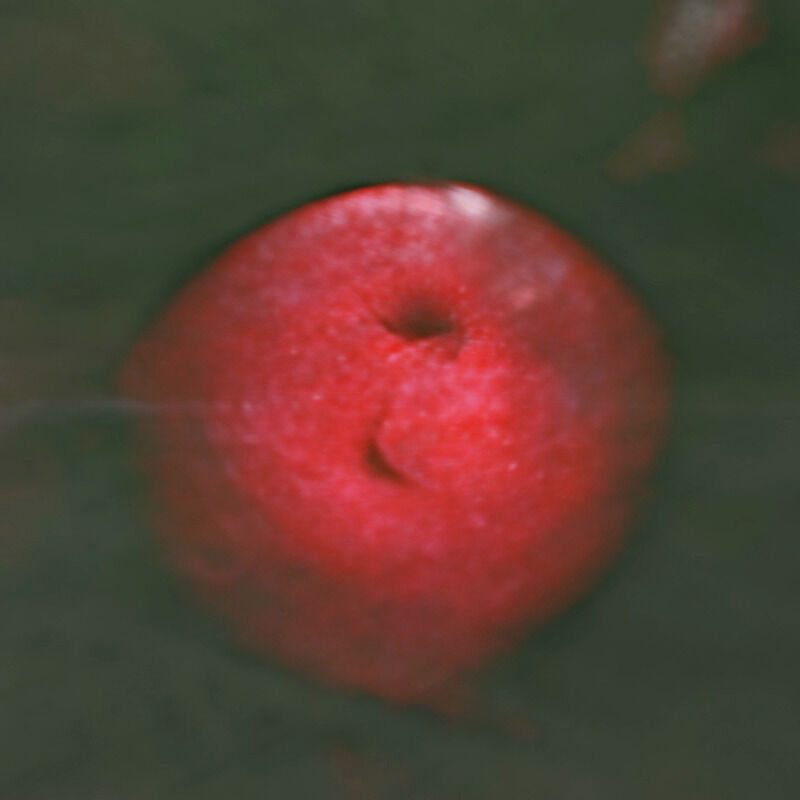}
\end{subfigure}
\begin{subfigure}{0.1 \textwidth}
\includegraphics[width=\textwidth]{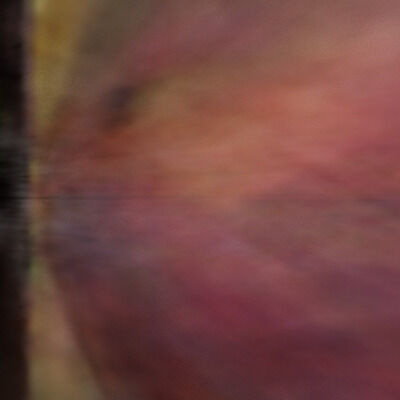}
\end{subfigure}
\begin{subfigure}{0.1 \textwidth}
\includegraphics[width=\textwidth]{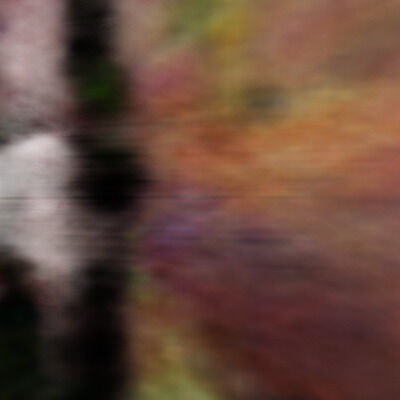}
\end{subfigure}
\begin{subfigure}{0.1 \textwidth}
\includegraphics[width=\textwidth]{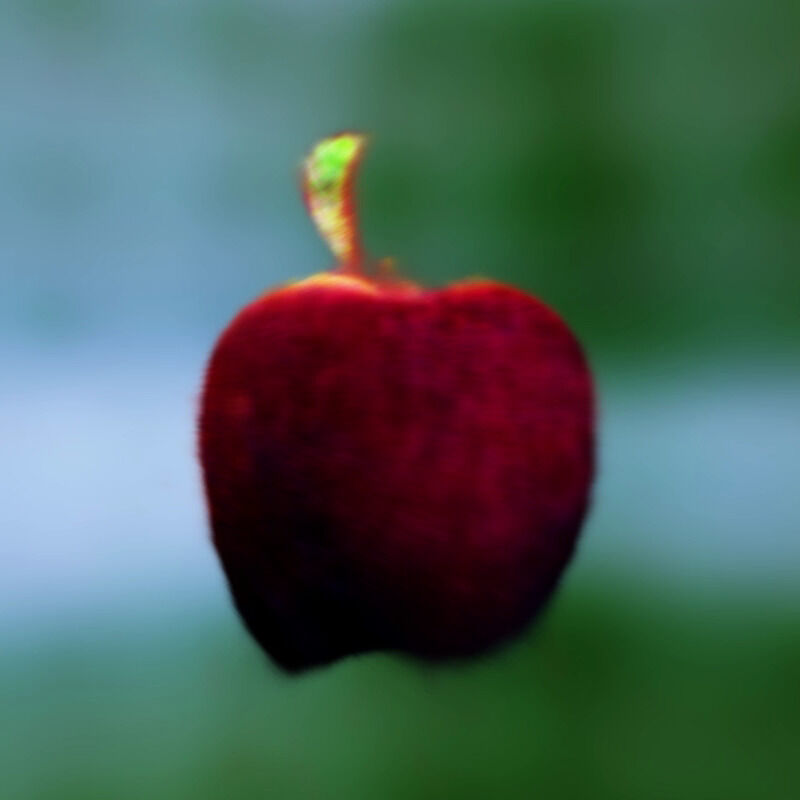}
\end{subfigure}
\begin{subfigure}{0.1 \textwidth}
\includegraphics[width=\textwidth]{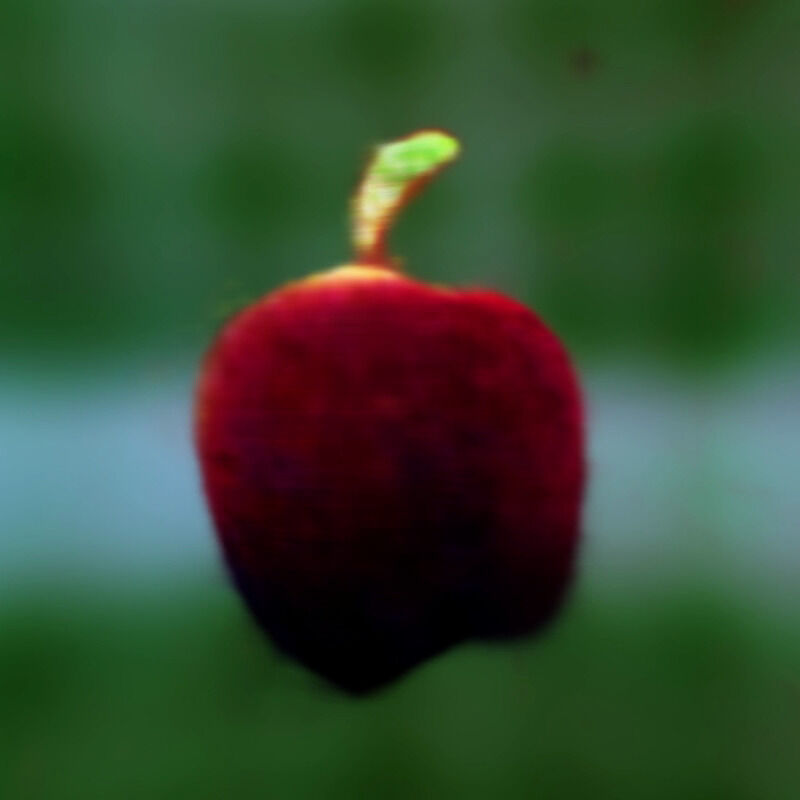}
\end{subfigure}
\\
\hspace*{.1\textwidth}
\begin{subfigure}{0.1 \textwidth}
\includegraphics[width=\textwidth]{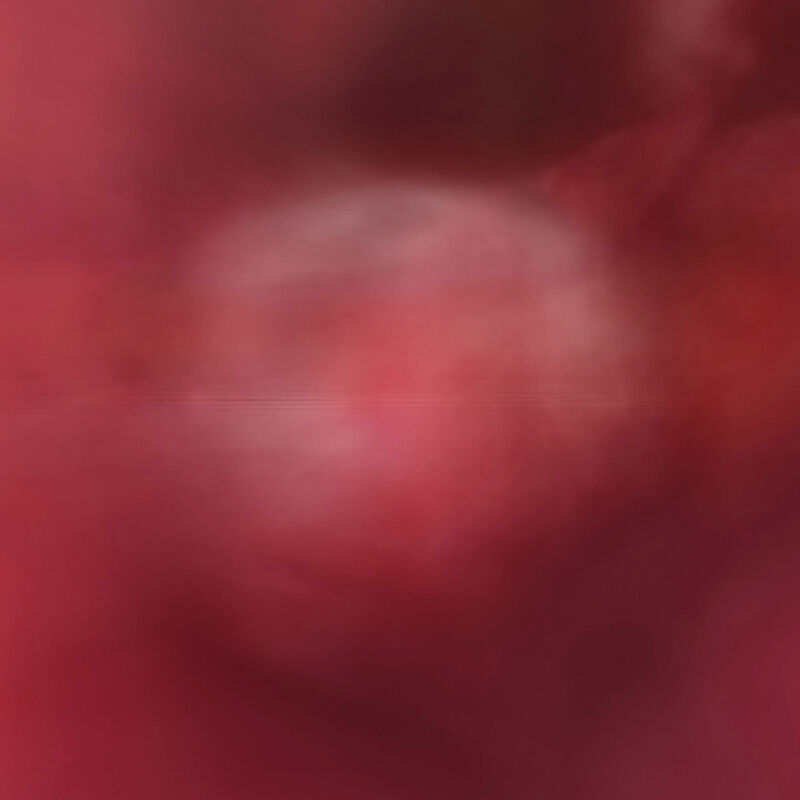}
\end{subfigure}
\begin{subfigure}{0.1 \textwidth}
\includegraphics[width=\textwidth]{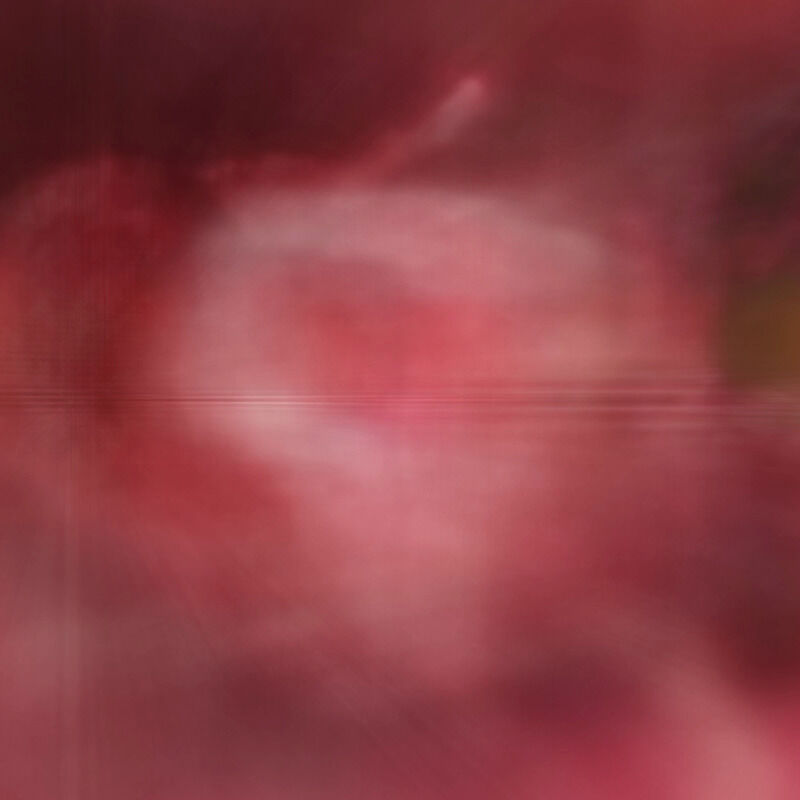}
\end{subfigure}
\begin{subfigure}{0.1 \textwidth}
\includegraphics[width=\textwidth]{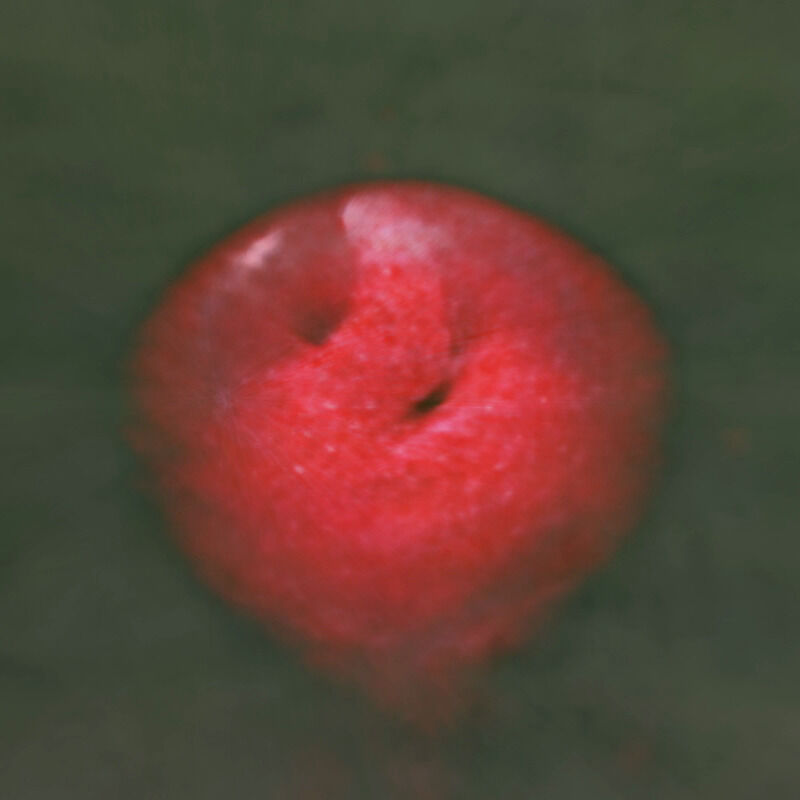}
\end{subfigure}
\begin{subfigure}{0.1 \textwidth}
\includegraphics[width=\textwidth]{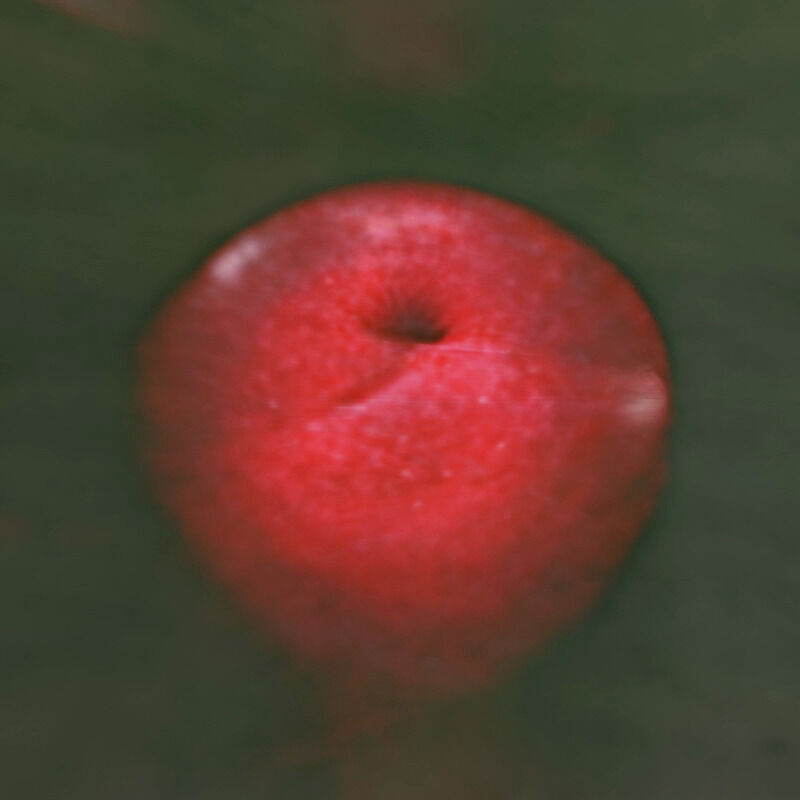}
\end{subfigure}
\begin{subfigure}{0.1 \textwidth}
\includegraphics[width=\textwidth]{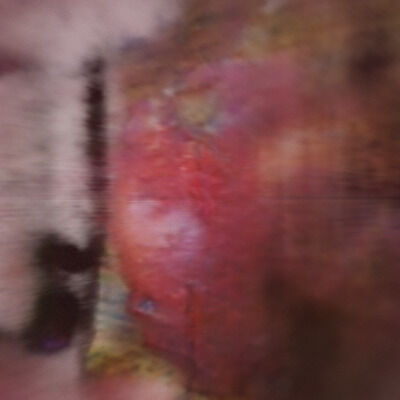}
\end{subfigure}
\begin{subfigure}{0.1 \textwidth}
\includegraphics[width=\textwidth]{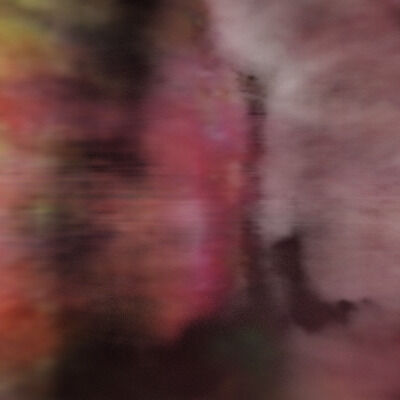}
\end{subfigure}
\begin{subfigure}{0.1 \textwidth}
\includegraphics[width=\textwidth]{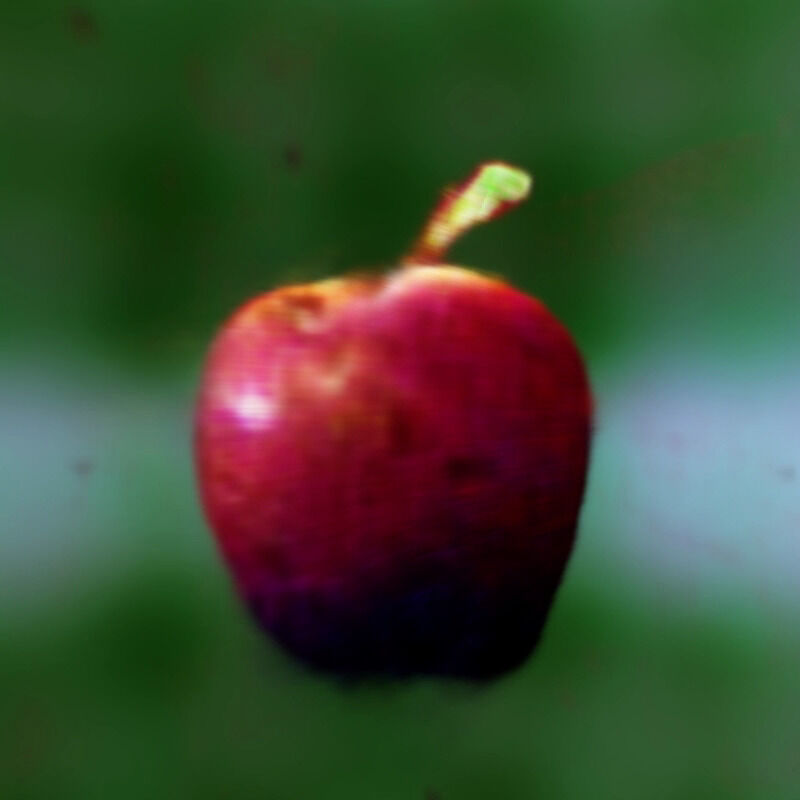}
\end{subfigure}
\begin{subfigure}{0.1 \textwidth}
\includegraphics[width=\textwidth]{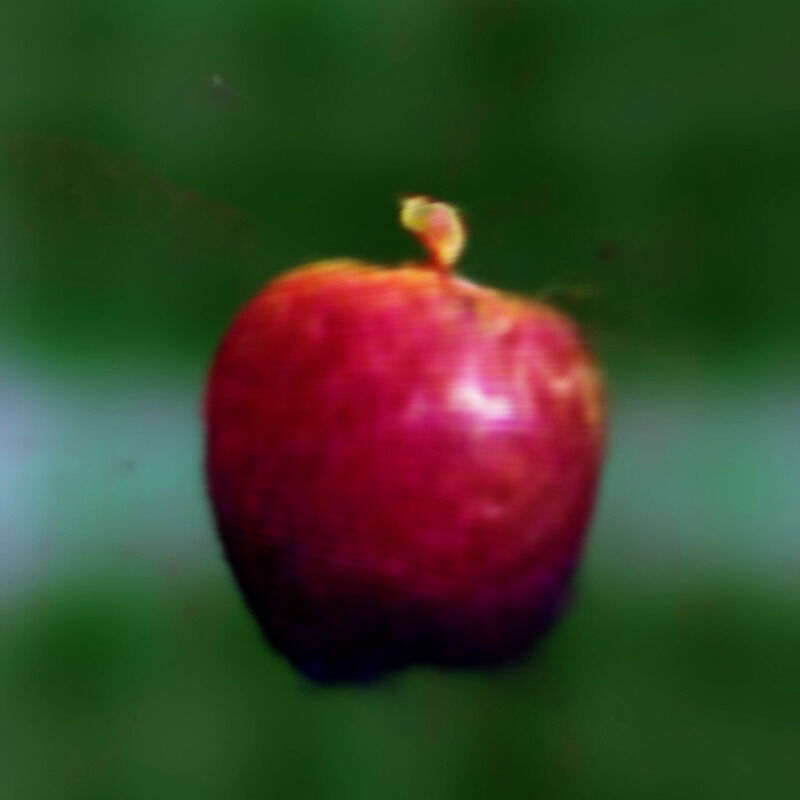}
\end{subfigure}
\begin{subfigure}{0.1 \textwidth}
\includegraphics[width=\textwidth]{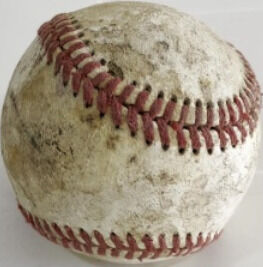}
\end{subfigure}
\begin{subfigure}{0.1 \textwidth}
\includegraphics[width=\textwidth]{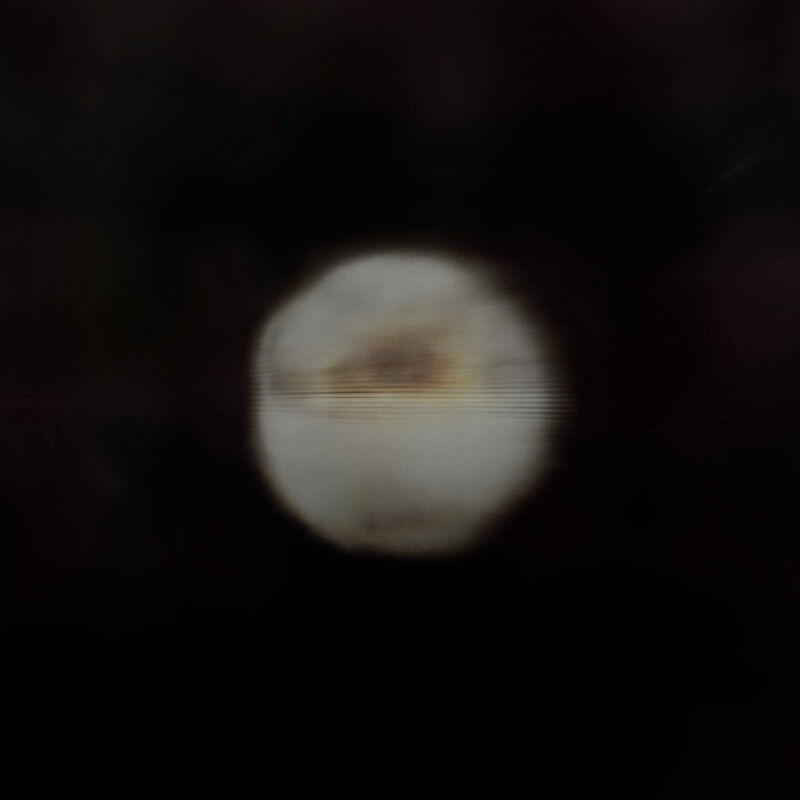}
\end{subfigure}
\begin{subfigure}{0.1 \textwidth}
\includegraphics[width=\textwidth]{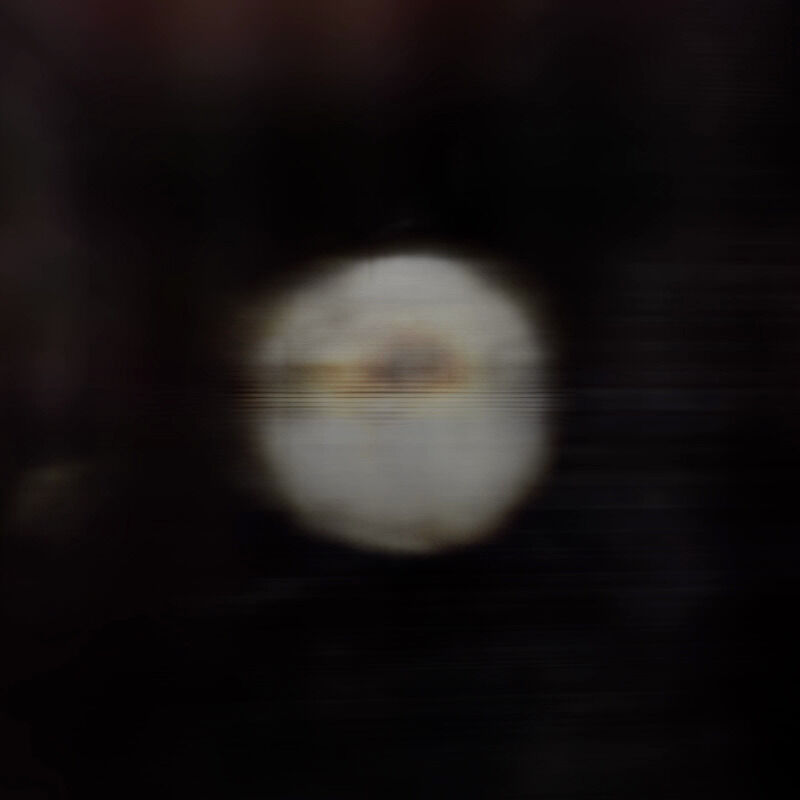}
\end{subfigure}
\begin{subfigure}{0.1 \textwidth}
\includegraphics[width=\textwidth]{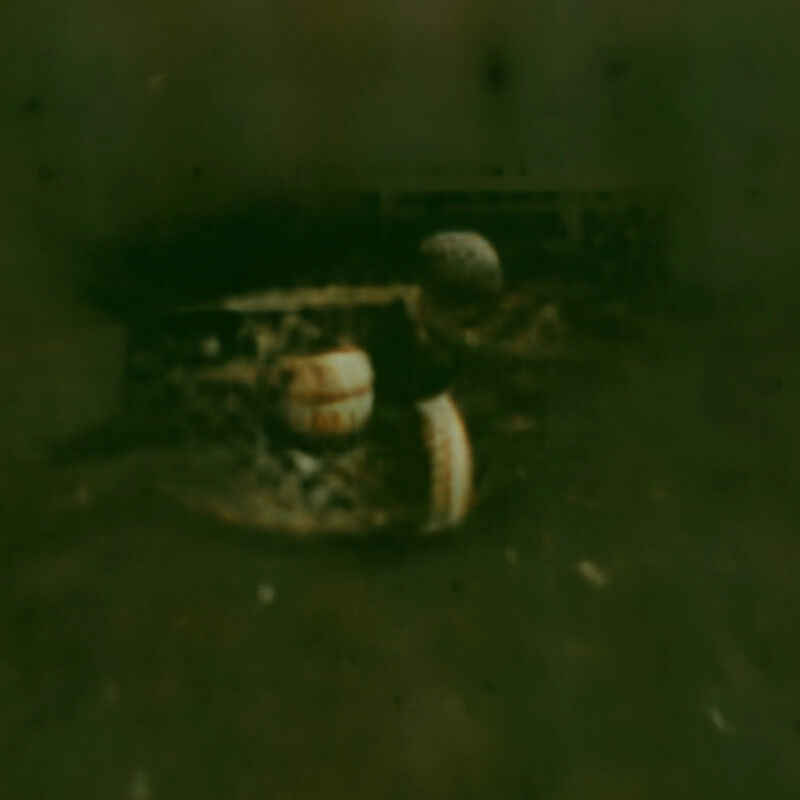}
\end{subfigure}
\begin{subfigure}{0.1 \textwidth}
\includegraphics[width=\textwidth]{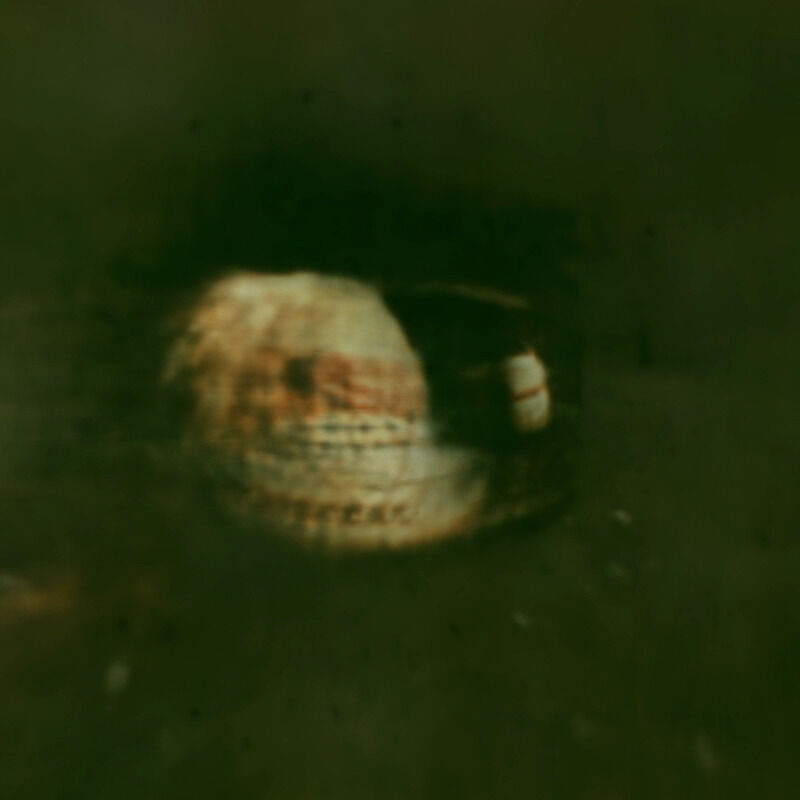}
\end{subfigure}
\begin{subfigure}{0.1 \textwidth}
\includegraphics[width=\textwidth]{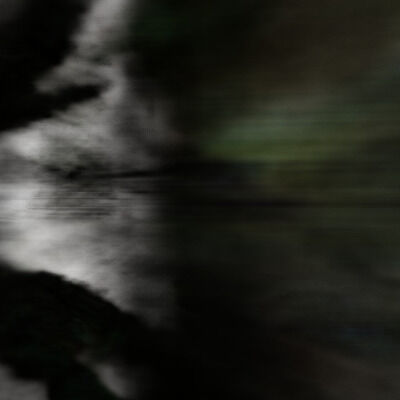}
\end{subfigure}
\begin{subfigure}{0.1 \textwidth}
\includegraphics[width=\textwidth]{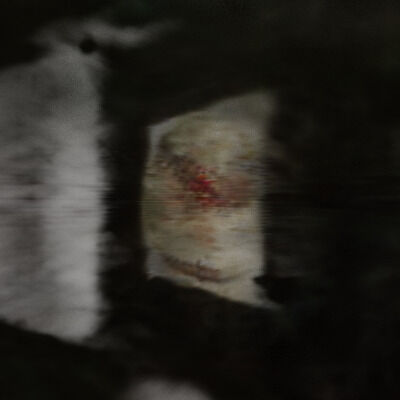}
\end{subfigure}
\begin{subfigure}{0.1 \textwidth}
\includegraphics[width=\textwidth]{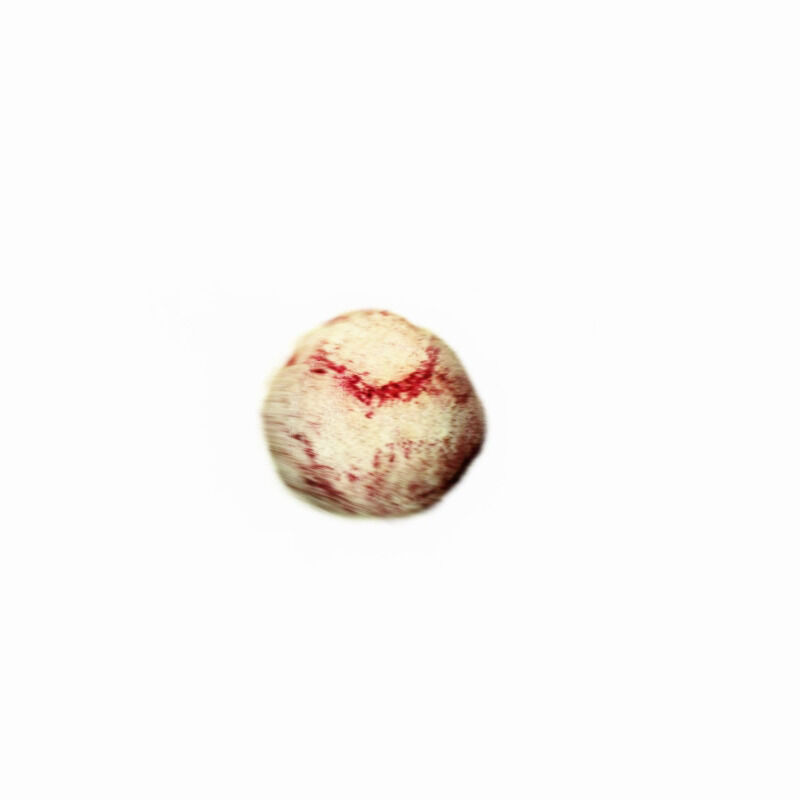}
\end{subfigure}
\begin{subfigure}{0.1 \textwidth}
\includegraphics[width=\textwidth]{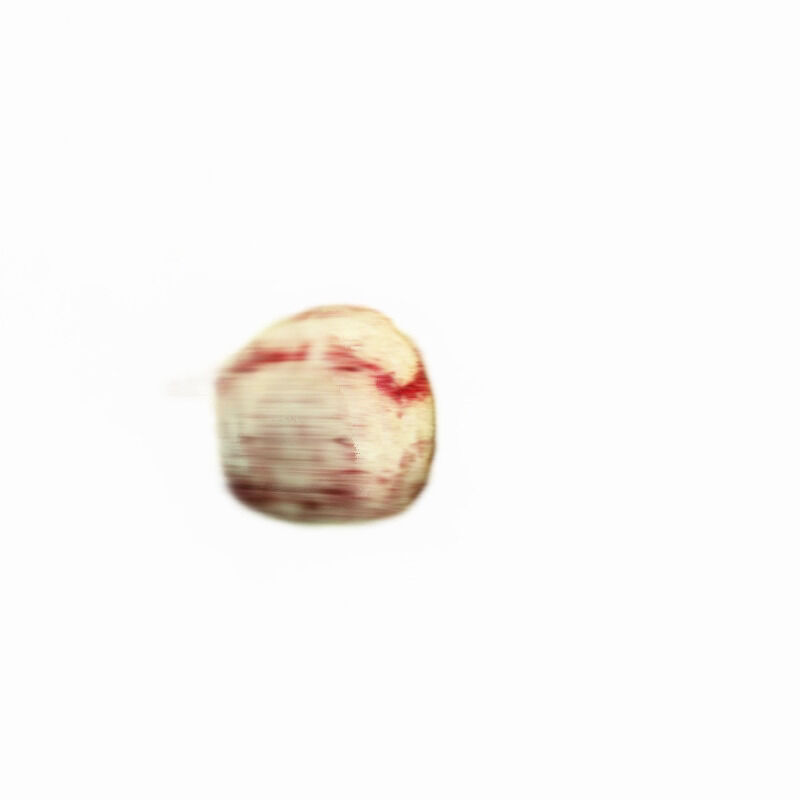}
\end{subfigure}
\\
\hspace*{.1\textwidth}
\begin{subfigure}{0.1 \textwidth}
\includegraphics[width=\textwidth]{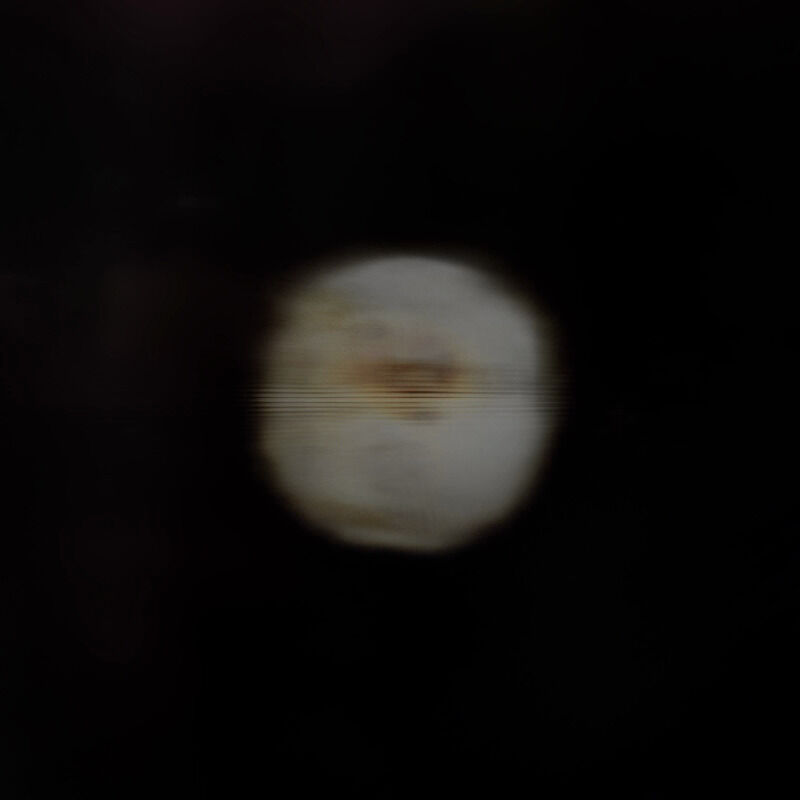}
\end{subfigure}
\begin{subfigure}{0.1 \textwidth}
\includegraphics[width=\textwidth]{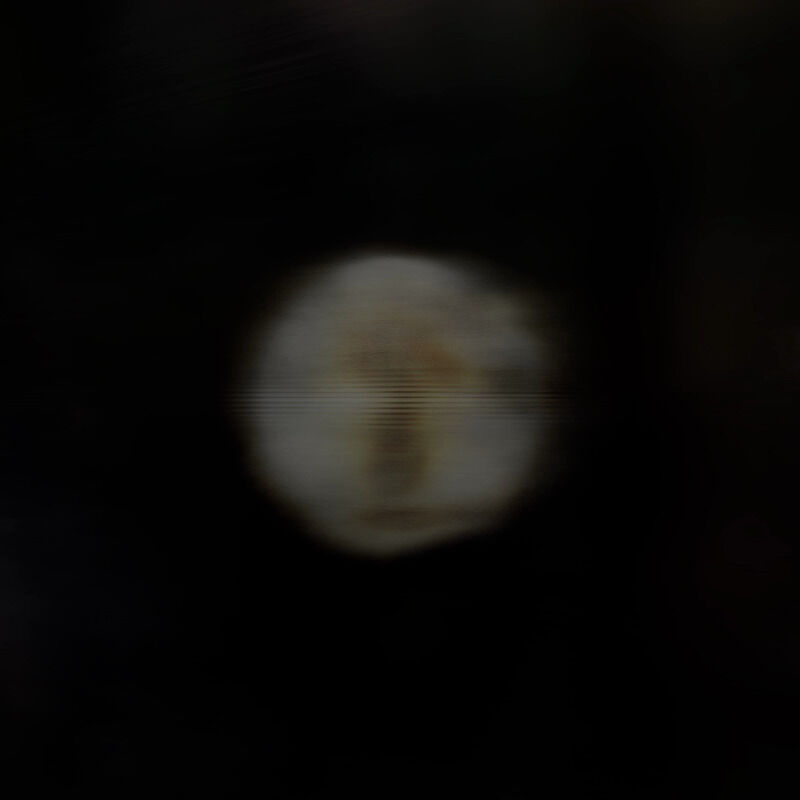}
\end{subfigure}
\begin{subfigure}{0.1 \textwidth}
\includegraphics[width=\textwidth]{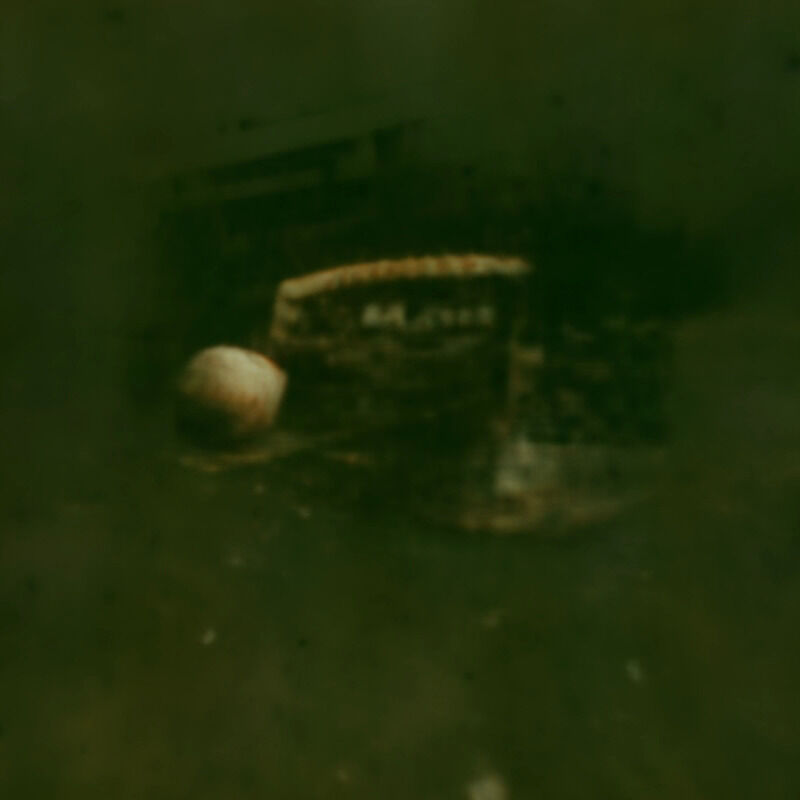}
\end{subfigure}
\begin{subfigure}{0.1 \textwidth}
\includegraphics[width=\textwidth]{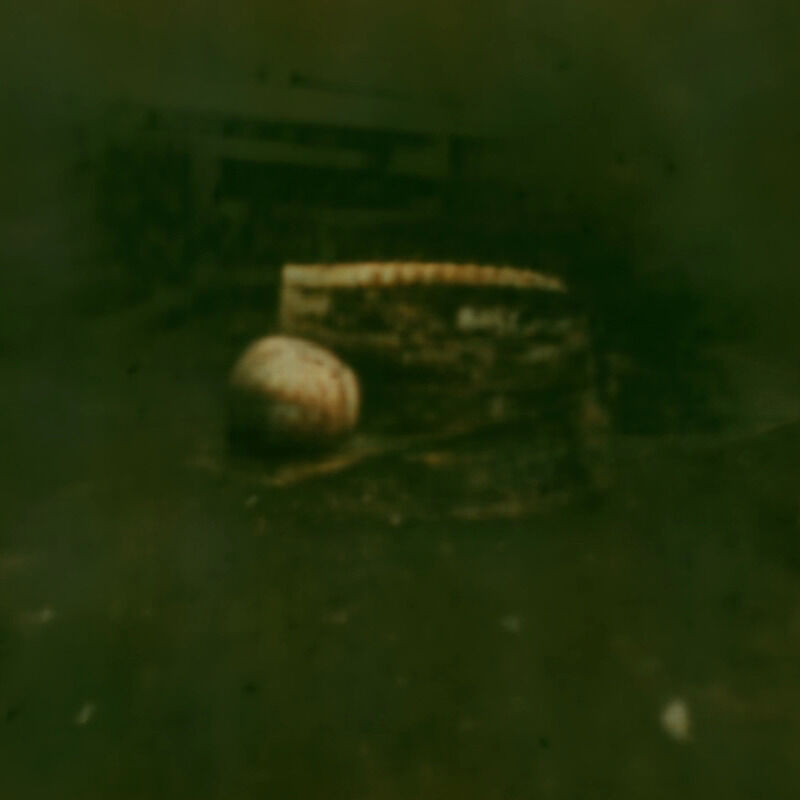}
\end{subfigure}
\begin{subfigure}{0.1 \textwidth}
\includegraphics[width=\textwidth]{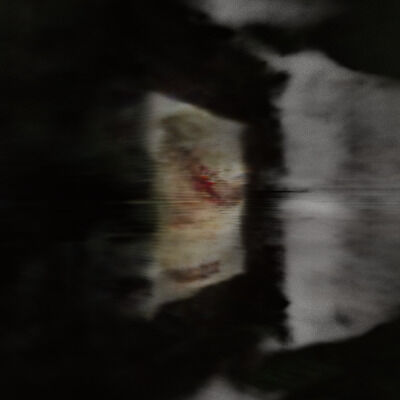}
\end{subfigure}
\begin{subfigure}{0.1 \textwidth}
\includegraphics[width=\textwidth]{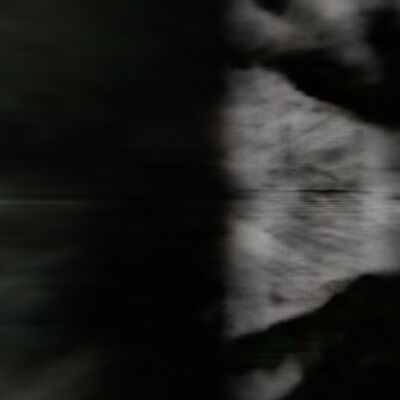}
\end{subfigure}
\begin{subfigure}{0.1 \textwidth}
\includegraphics[width=\textwidth]{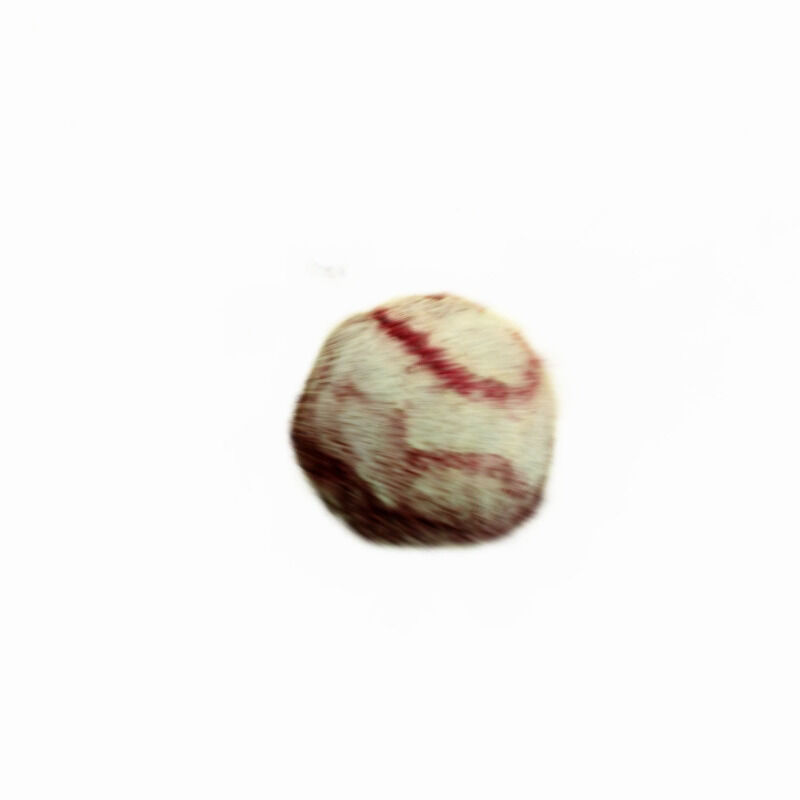}
\end{subfigure}
\begin{subfigure}{0.1 \textwidth}
\includegraphics[width=\textwidth]{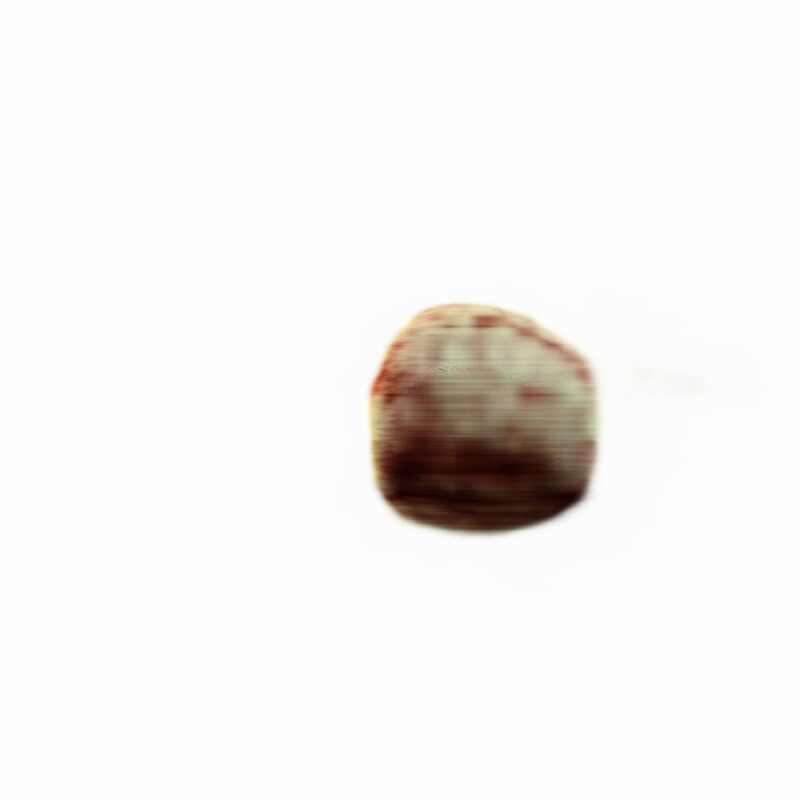}
\end{subfigure}
\begin{subfigure}{0.1 \textwidth}
\includegraphics[width=\textwidth]{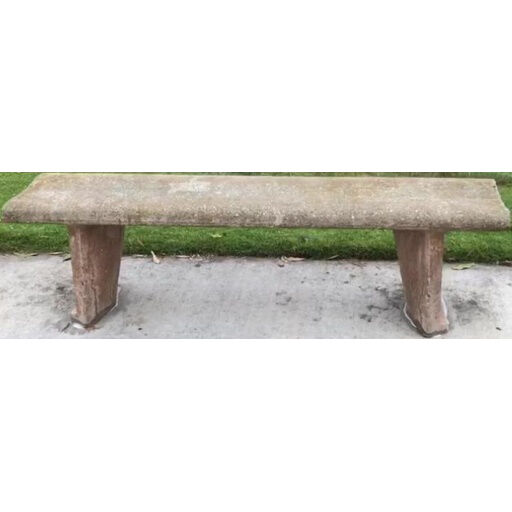}
\end{subfigure}
\begin{subfigure}{0.1 \textwidth}
\includegraphics[width=\textwidth]{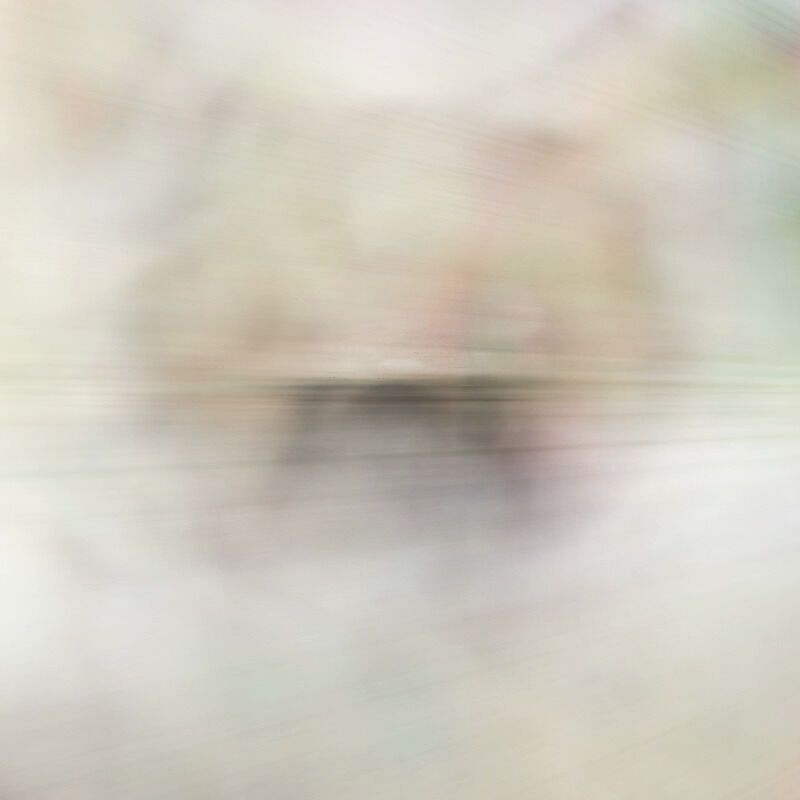}
\end{subfigure}
\begin{subfigure}{0.1 \textwidth}
\includegraphics[width=\textwidth]{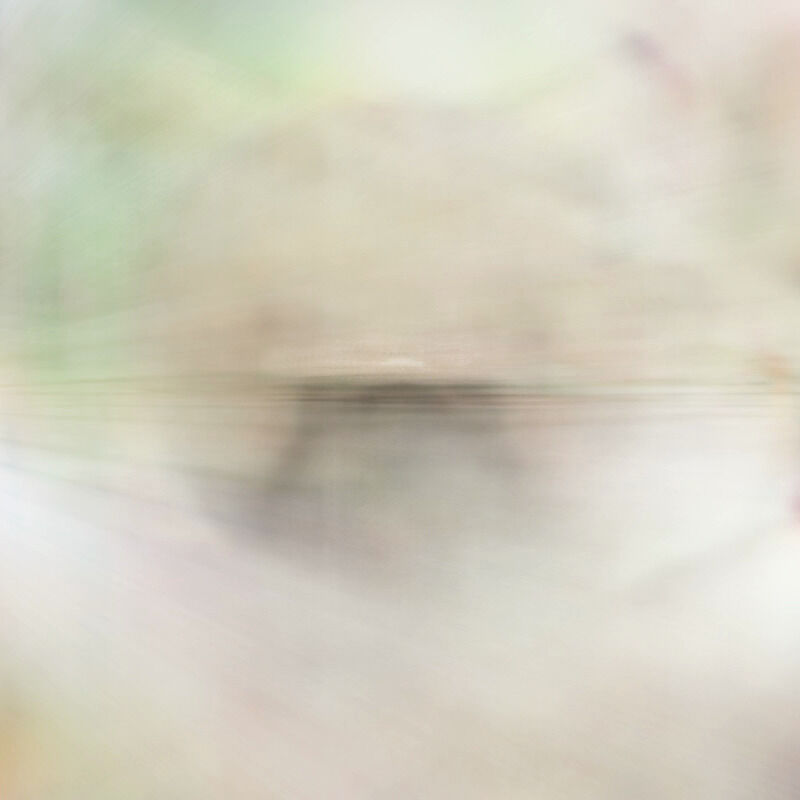}
\end{subfigure}
\begin{subfigure}{0.1 \textwidth}
\includegraphics[width=\textwidth]{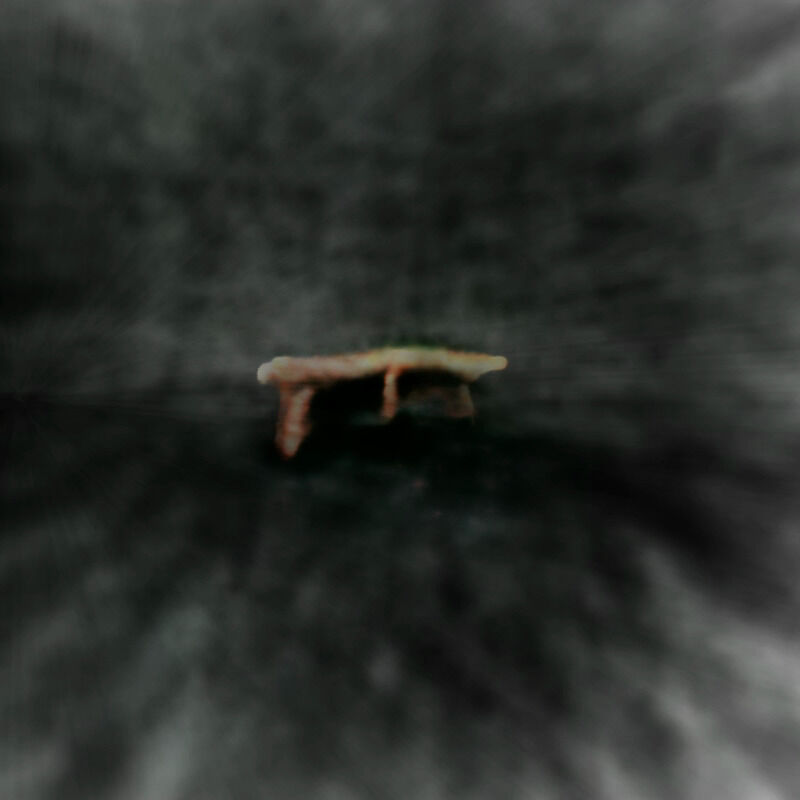}
\end{subfigure}
\begin{subfigure}{0.1 \textwidth}
\includegraphics[width=\textwidth]{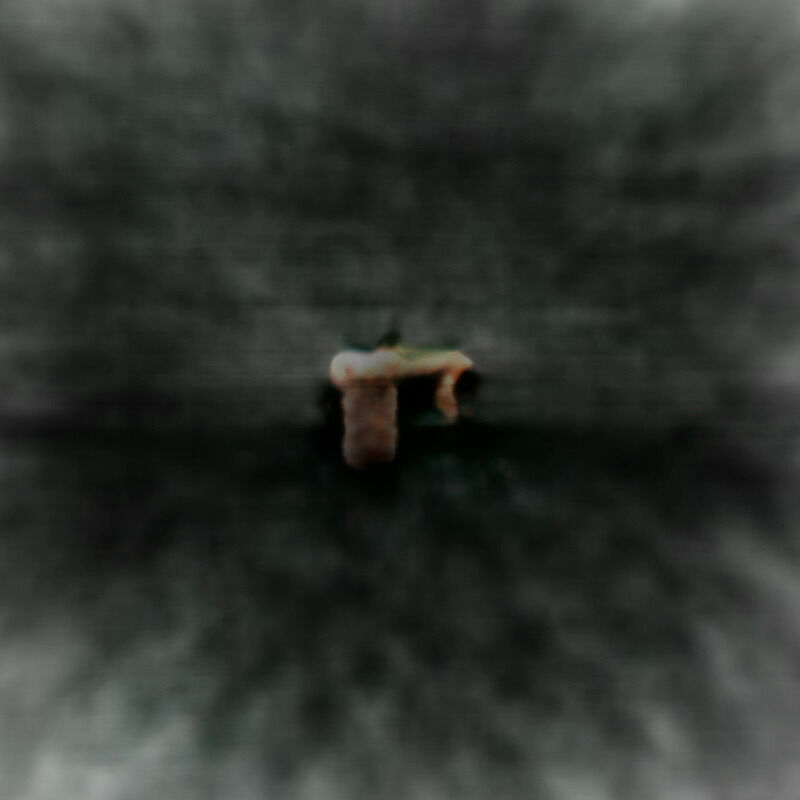}
\end{subfigure}
\begin{subfigure}{0.1 \textwidth}
\includegraphics[width=\textwidth]{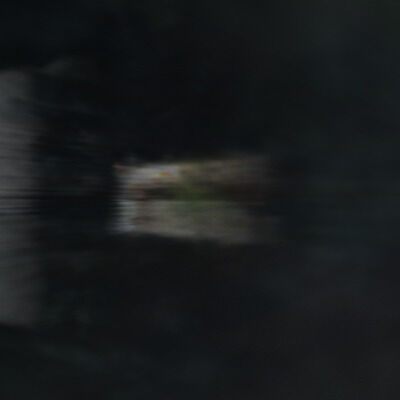}
\end{subfigure}
\begin{subfigure}{0.1 \textwidth}
\includegraphics[width=\textwidth]{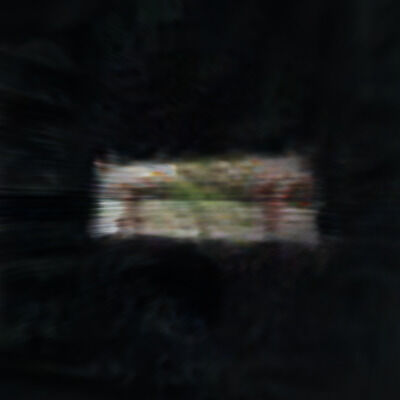}
\end{subfigure}
\begin{subfigure}{0.1 \textwidth}
\includegraphics[width=\textwidth]{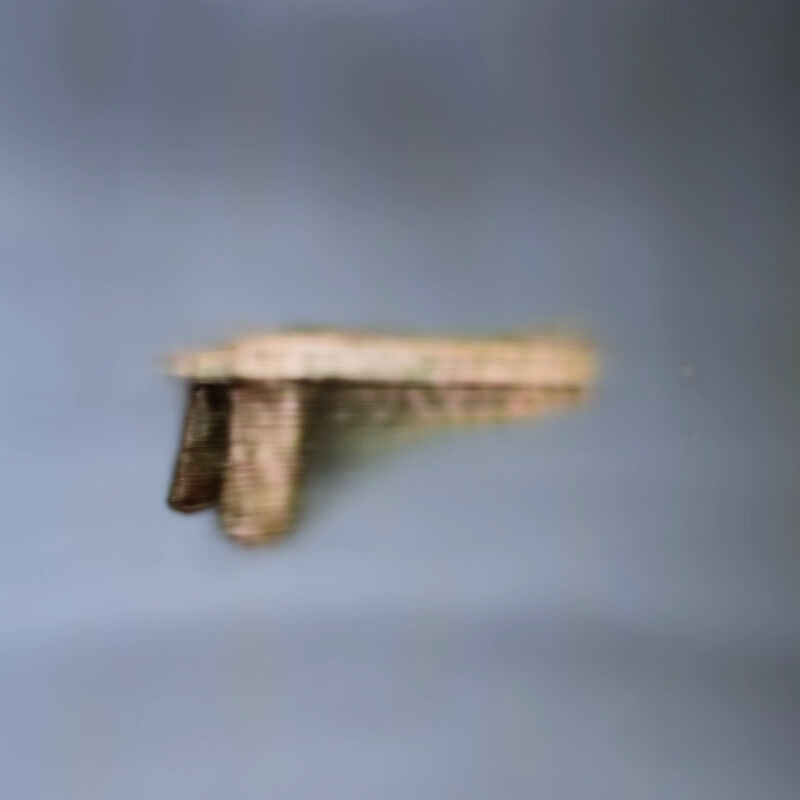}
\end{subfigure}
\begin{subfigure}{0.1 \textwidth}
\includegraphics[width=\textwidth]{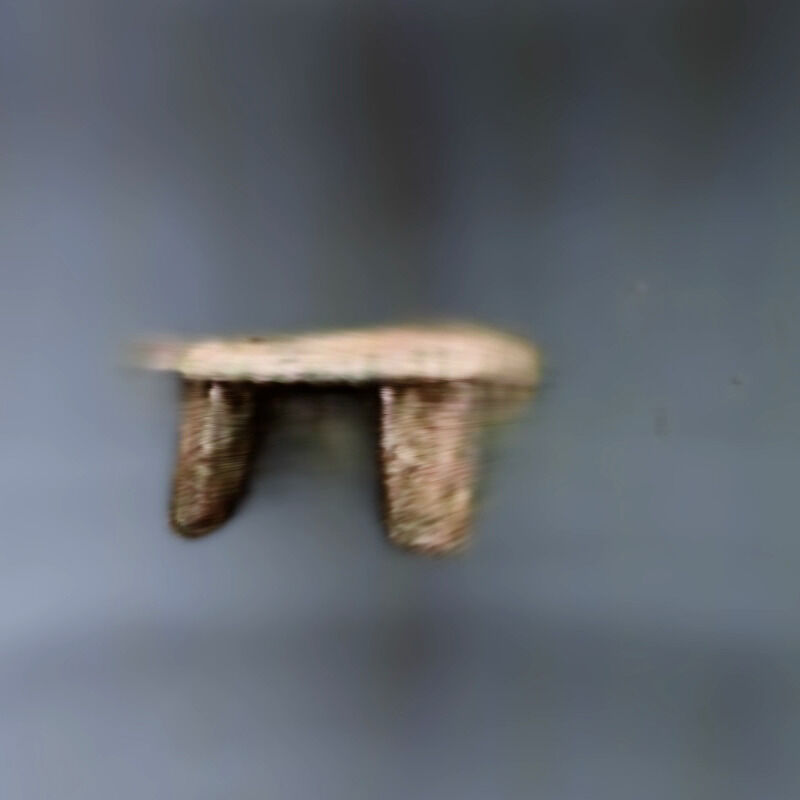}
\end{subfigure}
\\
\hspace*{.1\textwidth}
\begin{subfigure}{0.1 \textwidth}
\includegraphics[width=\textwidth]{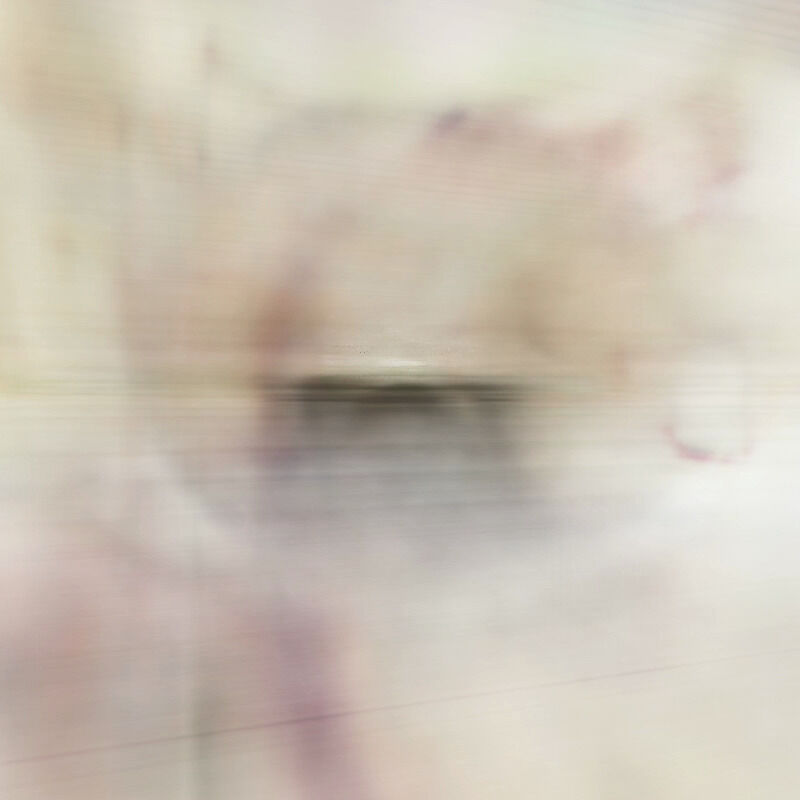}
\end{subfigure}
\begin{subfigure}{0.1 \textwidth}
\includegraphics[width=\textwidth]{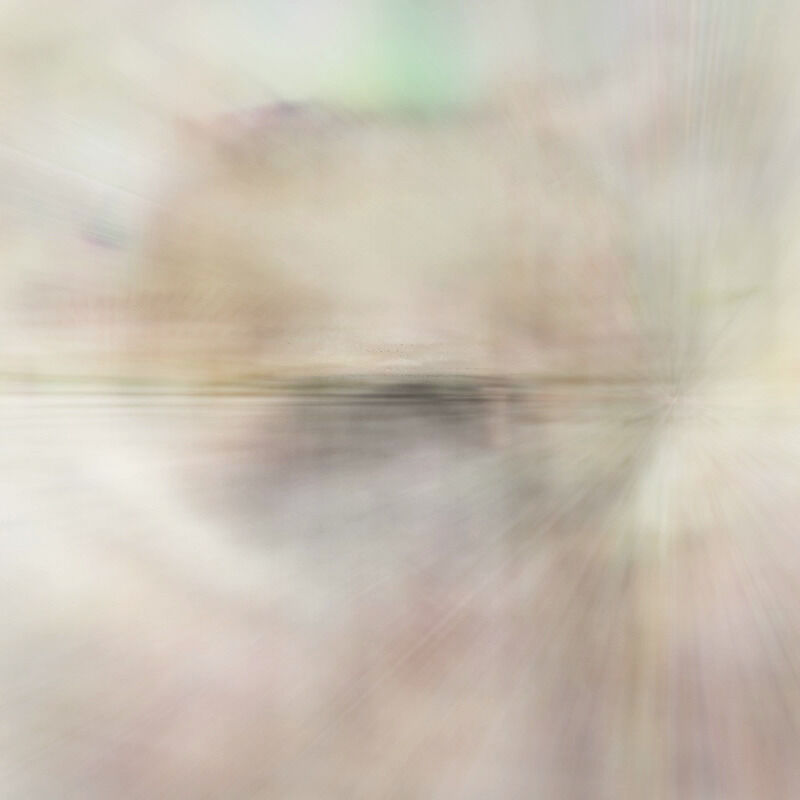}
\end{subfigure}
\begin{subfigure}{0.1 \textwidth}
\includegraphics[width=\textwidth]{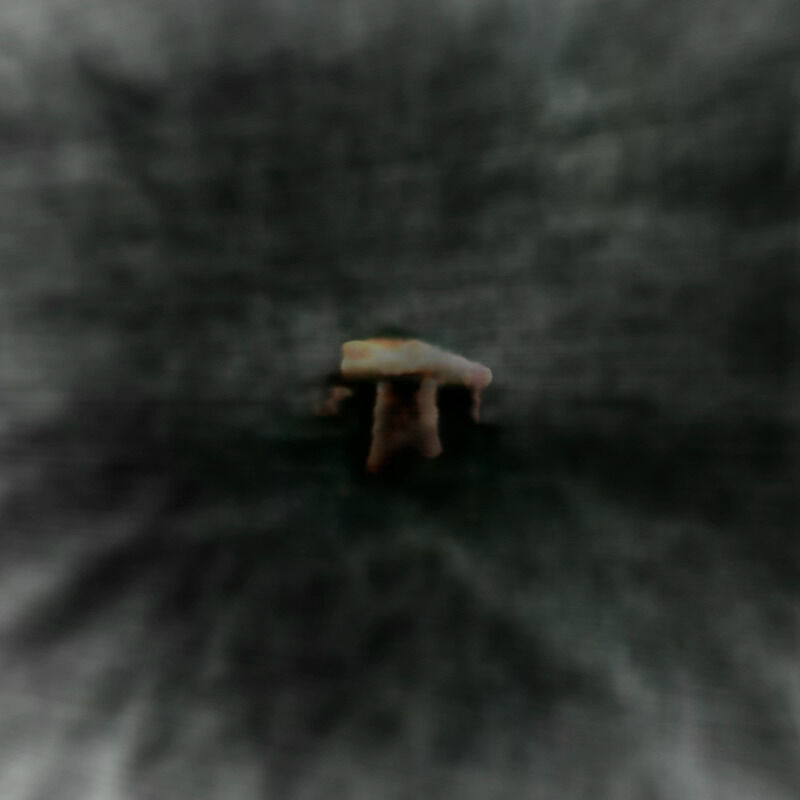}
\end{subfigure}
\begin{subfigure}{0.1 \textwidth}
\includegraphics[width=\textwidth]{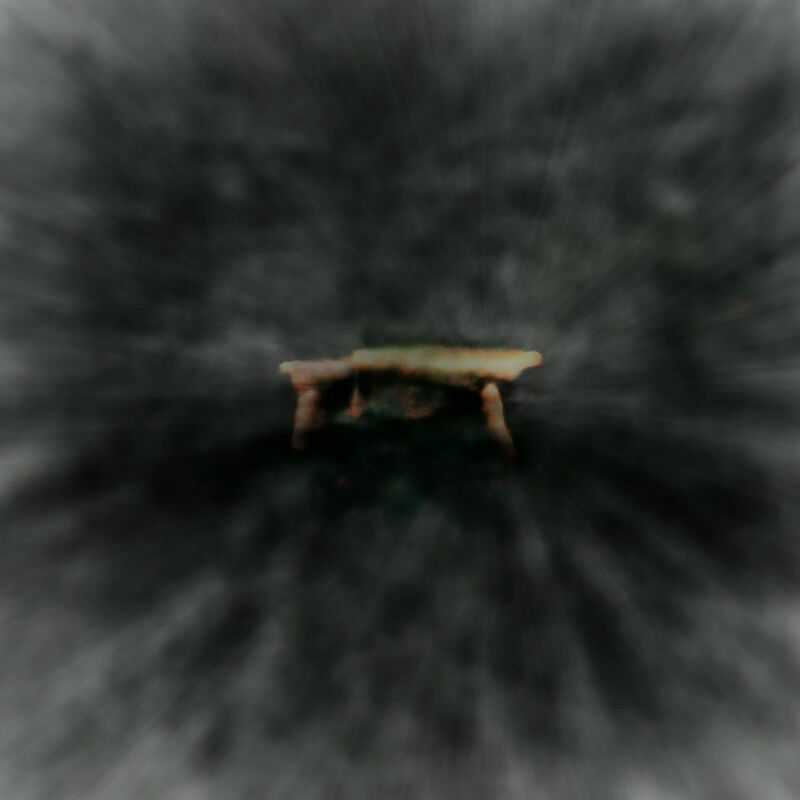}
\end{subfigure}
\begin{subfigure}{0.1 \textwidth}
\includegraphics[width=\textwidth]{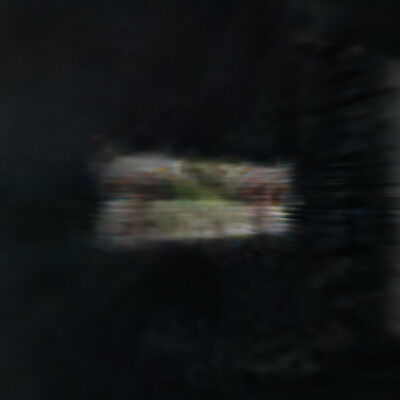}
\end{subfigure}
\begin{subfigure}{0.1 \textwidth}
\includegraphics[width=\textwidth]{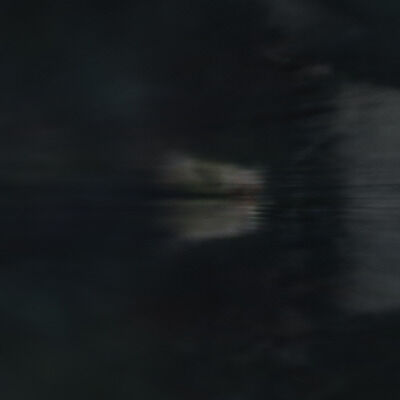}
\end{subfigure}
\begin{subfigure}{0.1 \textwidth}
\includegraphics[width=\textwidth]{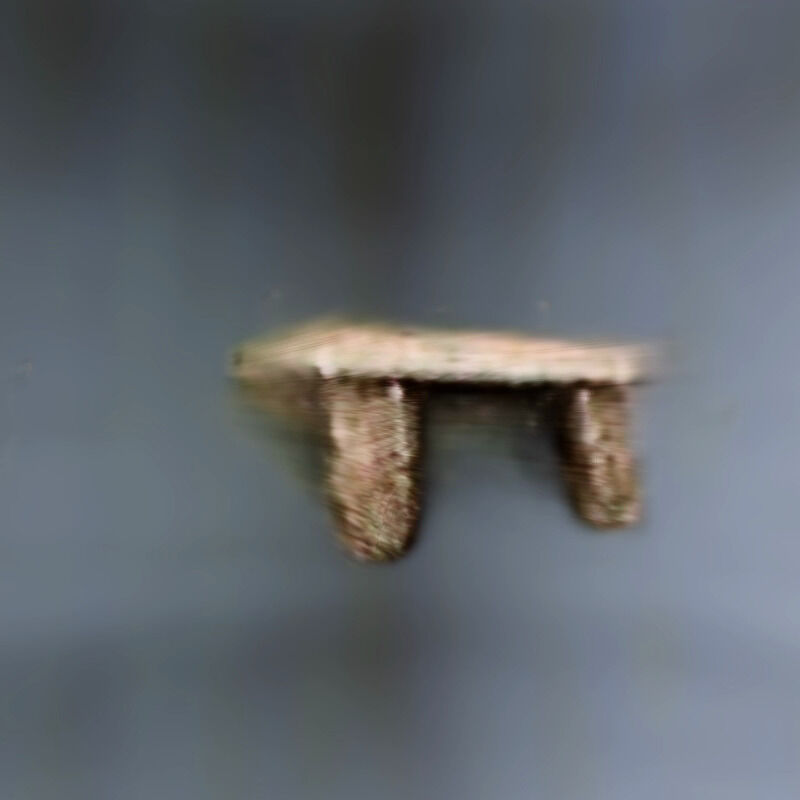}
\end{subfigure}
\begin{subfigure}{0.1 \textwidth}
\includegraphics[width=\textwidth]{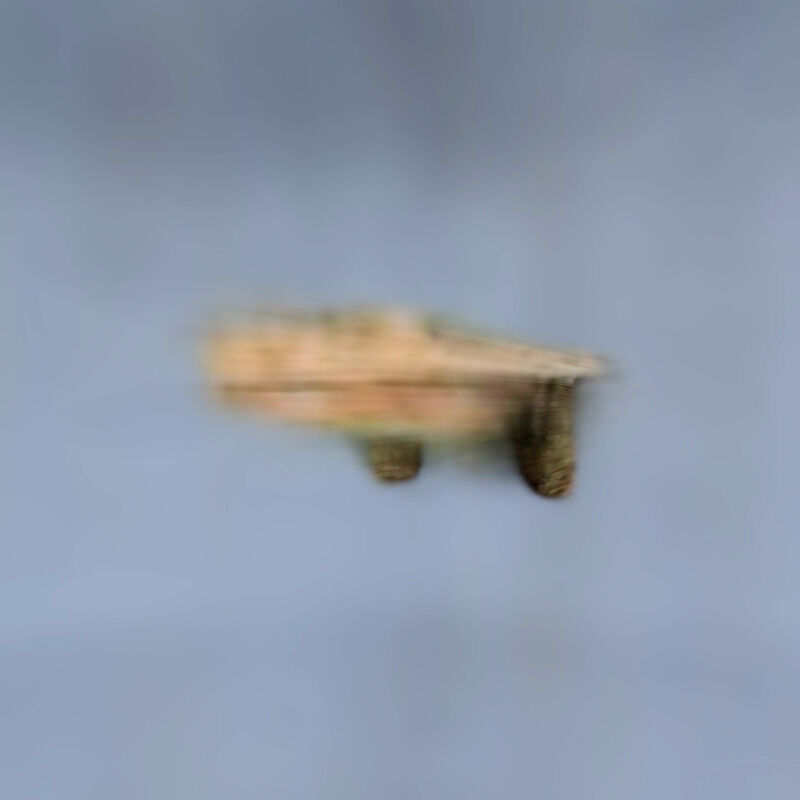}
\end{subfigure}
\begin{subfigure}{0.1 \textwidth}
\includegraphics[width=\textwidth]{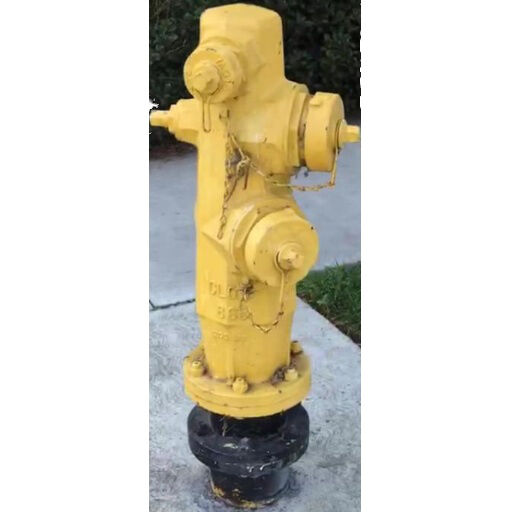}
\end{subfigure}
\begin{subfigure}{0.1 \textwidth}
\includegraphics[width=\textwidth]{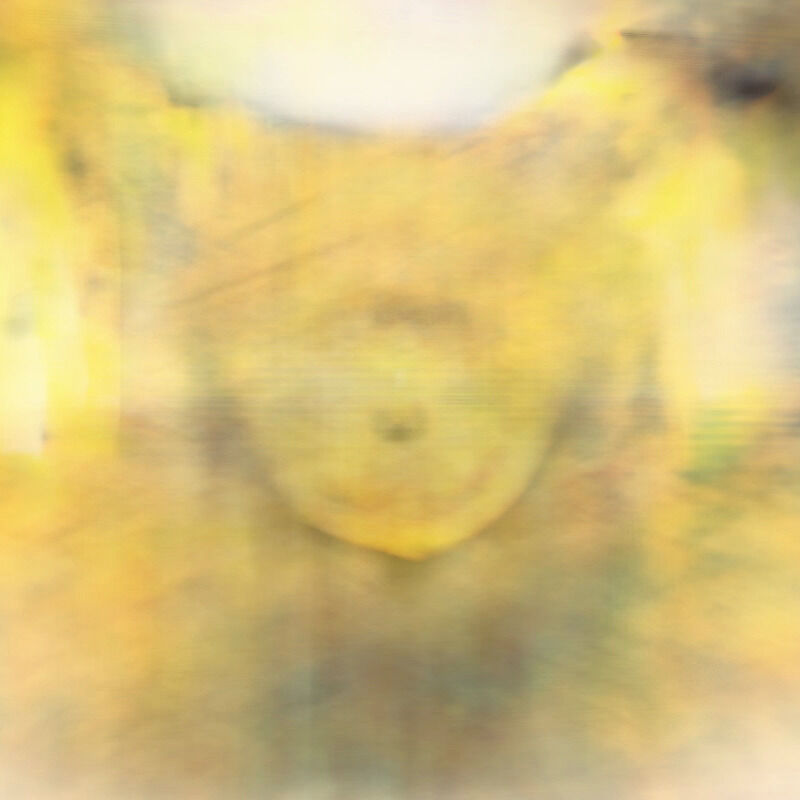}
\end{subfigure}
\begin{subfigure}{0.1 \textwidth}
\includegraphics[width=\textwidth]{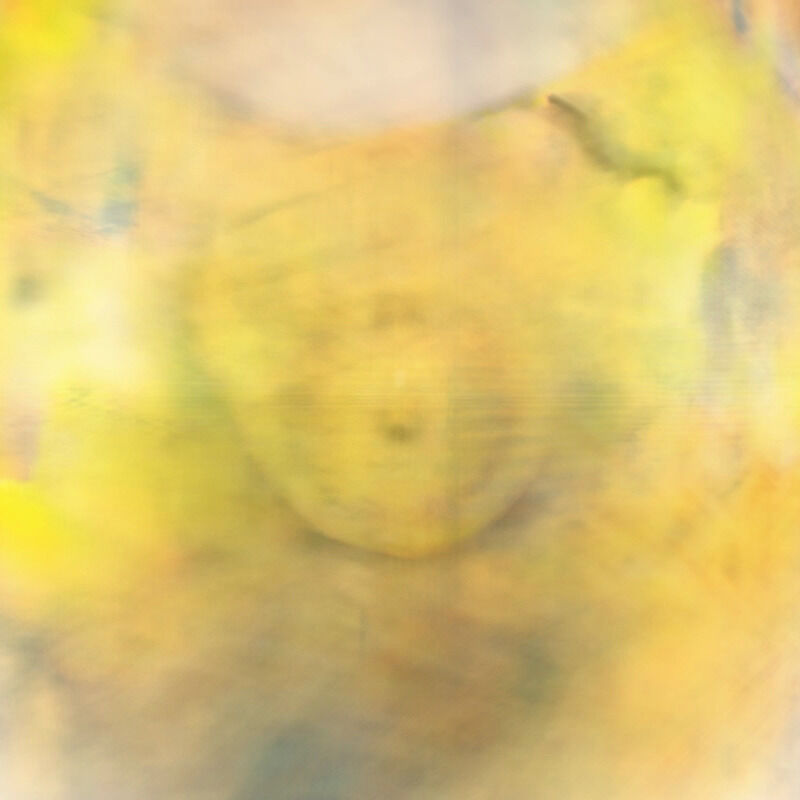}
\end{subfigure}
\begin{subfigure}{0.1 \textwidth}
\includegraphics[width=\textwidth]{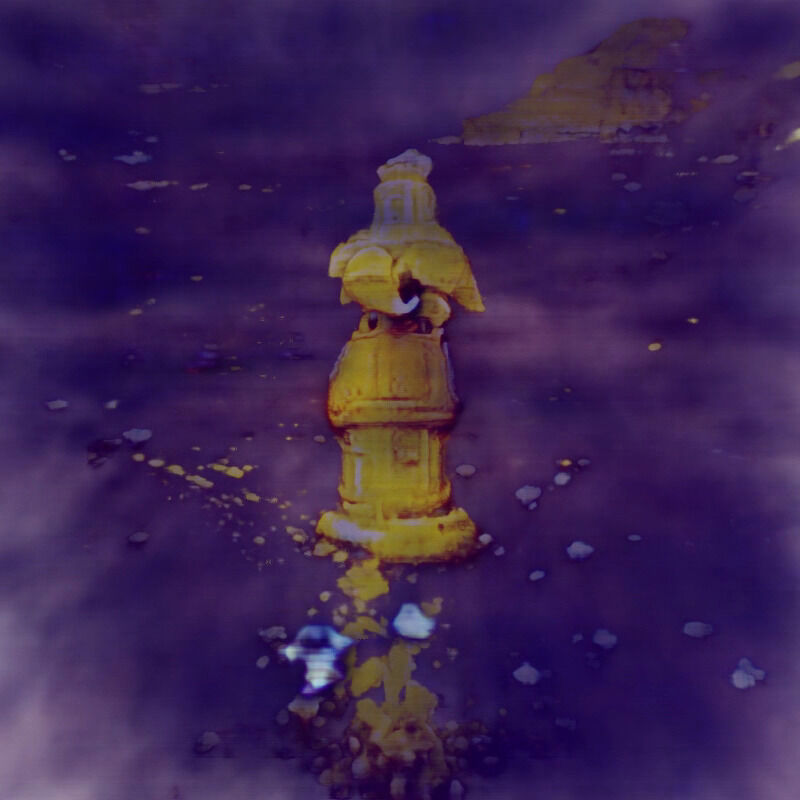}
\end{subfigure}
\begin{subfigure}{0.1 \textwidth}
\includegraphics[width=\textwidth]{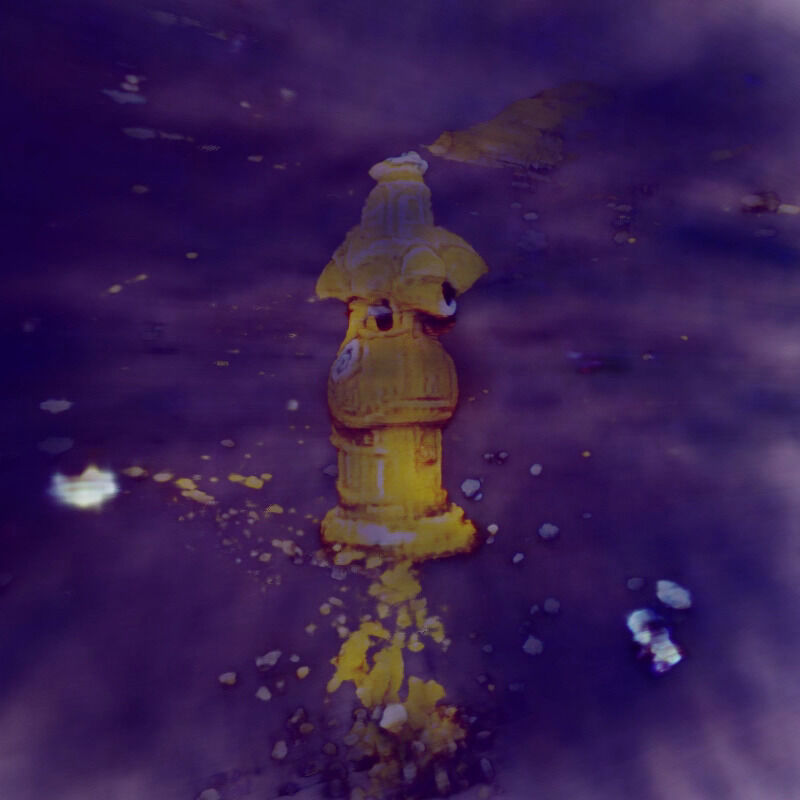}
\end{subfigure}
\begin{subfigure}{0.1 \textwidth}
\includegraphics[width=\textwidth]{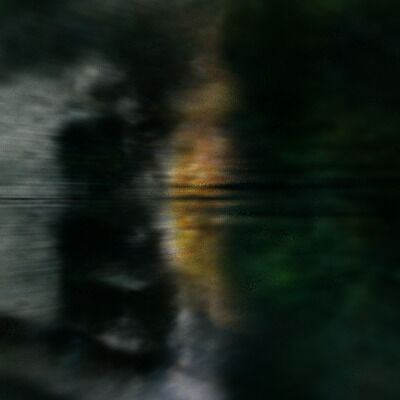}
\end{subfigure}
\begin{subfigure}{0.1 \textwidth}
\includegraphics[width=\textwidth]{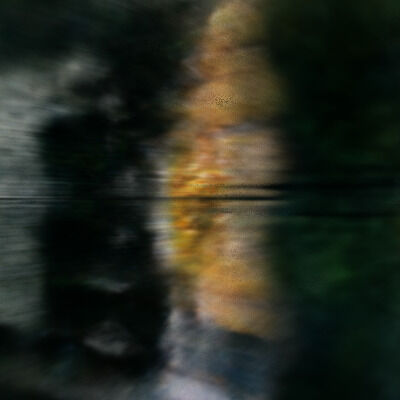}
\end{subfigure}
\begin{subfigure}{0.1 \textwidth}
\includegraphics[width=\textwidth]{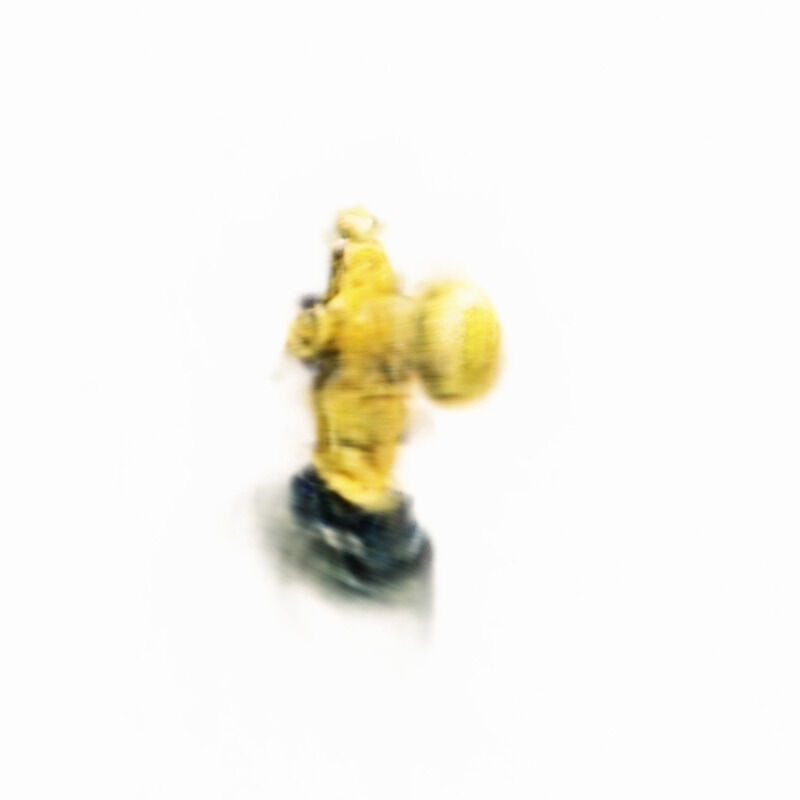}
\end{subfigure}
\begin{subfigure}{0.1 \textwidth}
\includegraphics[width=\textwidth]{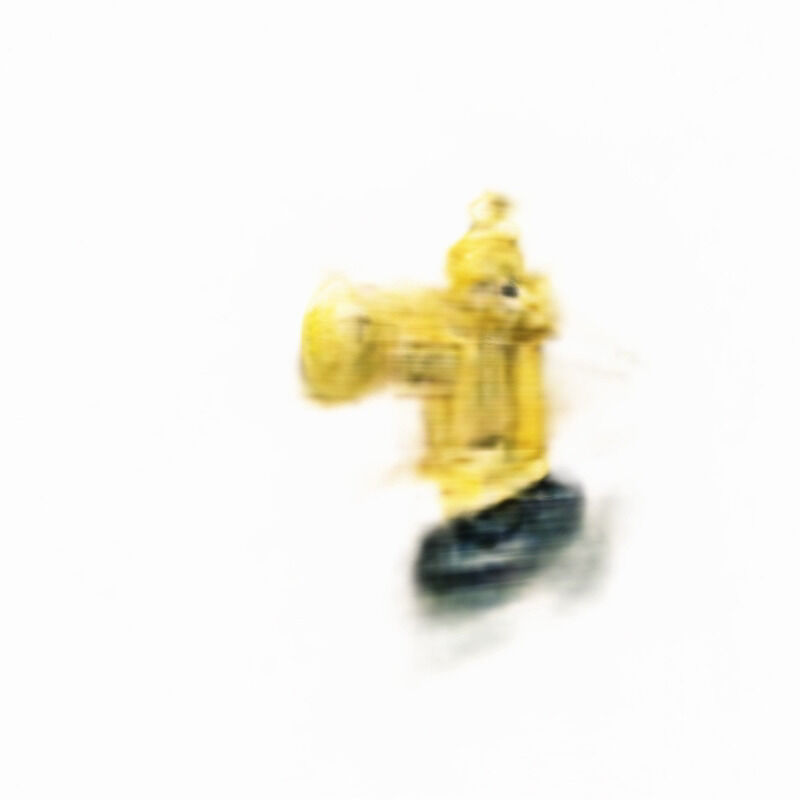}
\end{subfigure}
\\
\hspace*{.1\textwidth}
\begin{subfigure}{0.1 \textwidth}
\includegraphics[width=\textwidth]{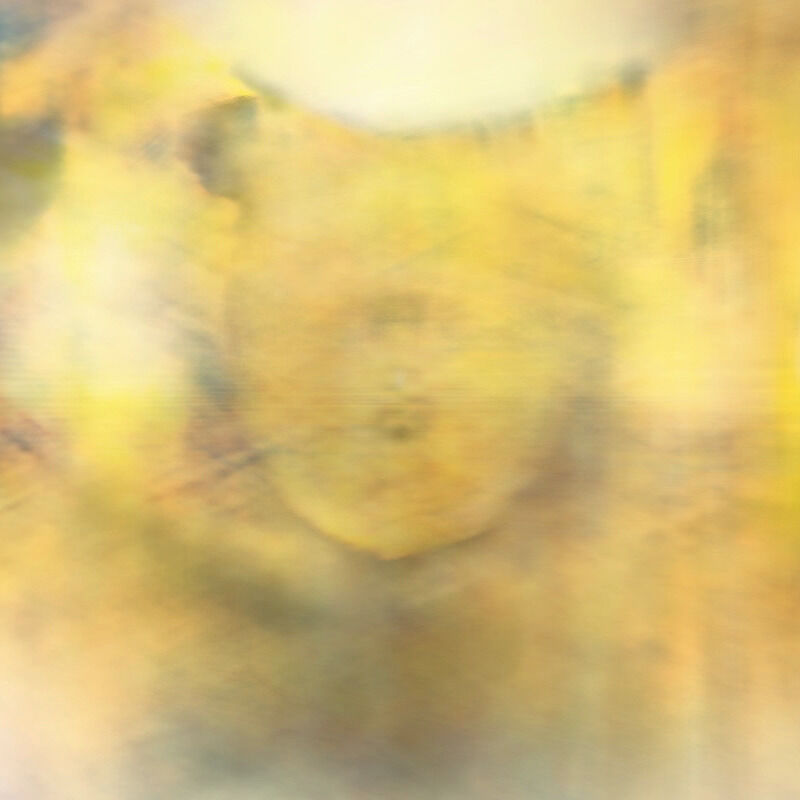}
\end{subfigure}
\begin{subfigure}{0.1 \textwidth}
\includegraphics[width=\textwidth]{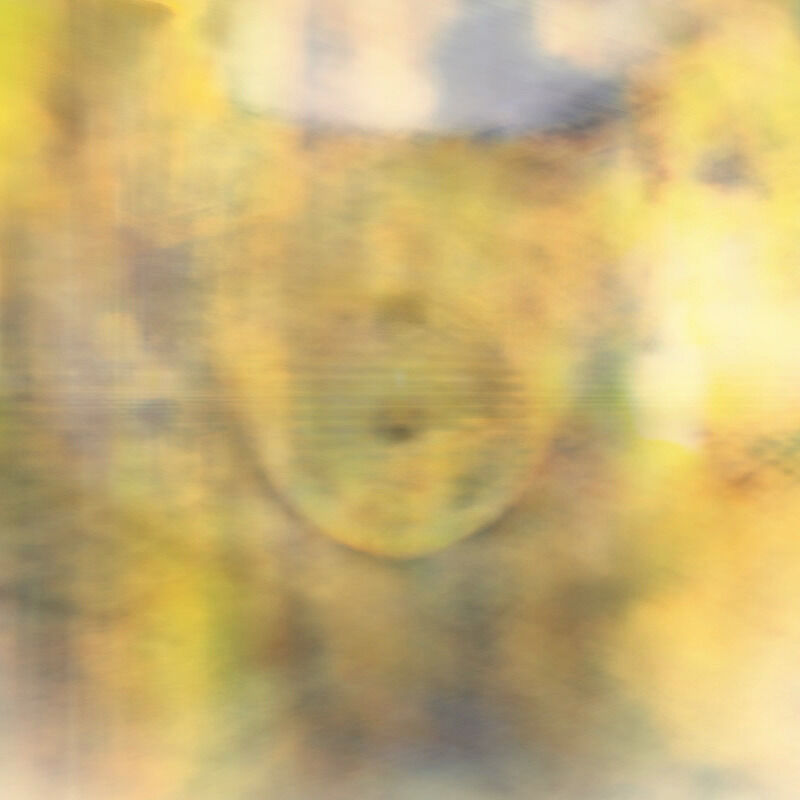}
\end{subfigure}
\begin{subfigure}{0.1 \textwidth}
\includegraphics[width=\textwidth]{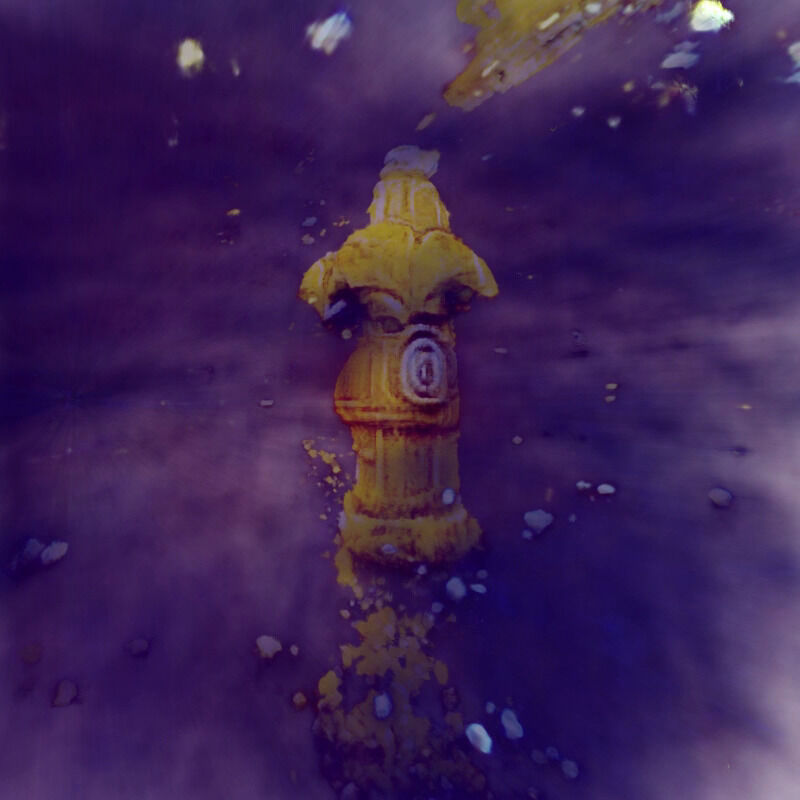}
\end{subfigure}
\begin{subfigure}{0.1 \textwidth}
\includegraphics[width=\textwidth]{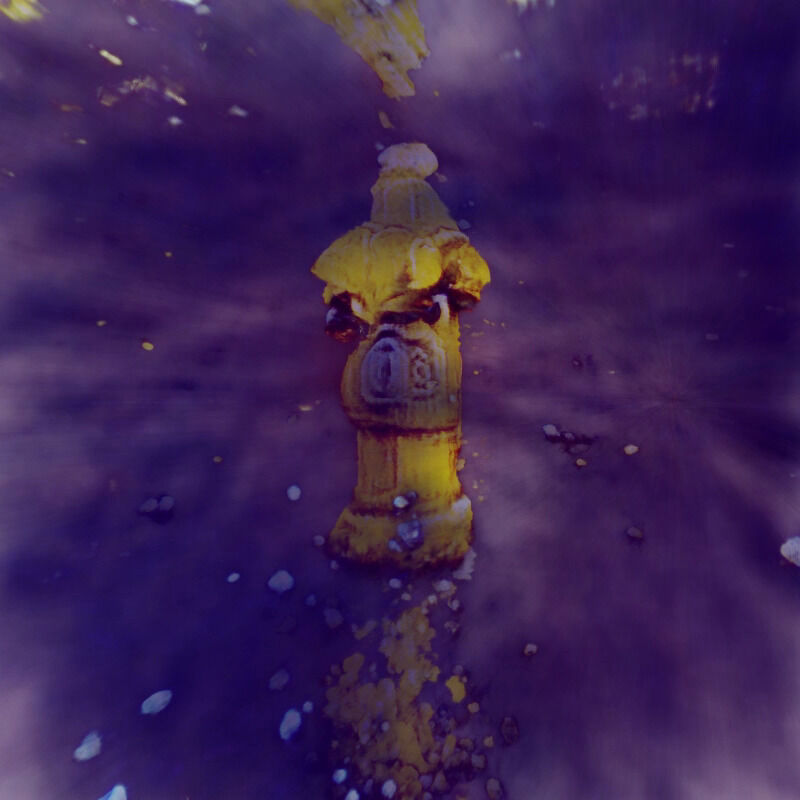}
\end{subfigure}
\begin{subfigure}{0.1 \textwidth}
\includegraphics[width=\textwidth]{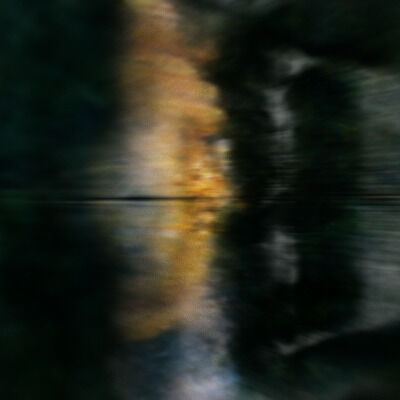}
\end{subfigure}
\begin{subfigure}{0.1 \textwidth}
\includegraphics[width=\textwidth]{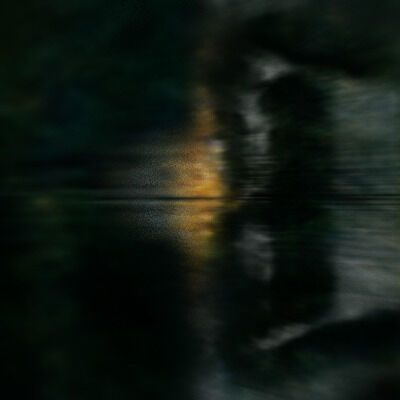}
\end{subfigure}
\begin{subfigure}{0.1 \textwidth}
\includegraphics[width=\textwidth]{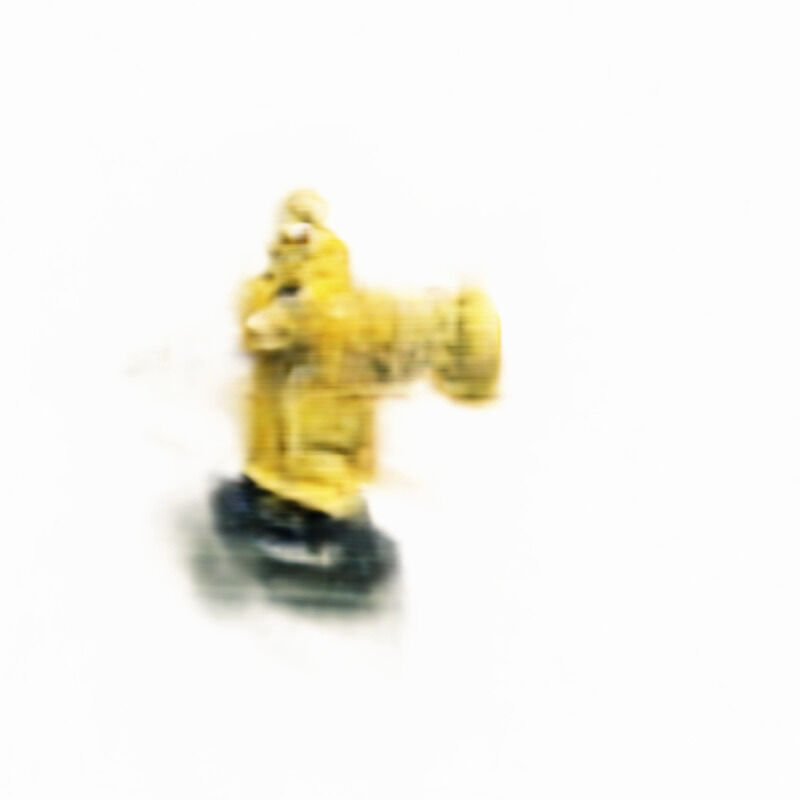}
\end{subfigure}
\begin{subfigure}{0.1 \textwidth}
\includegraphics[width=\textwidth]{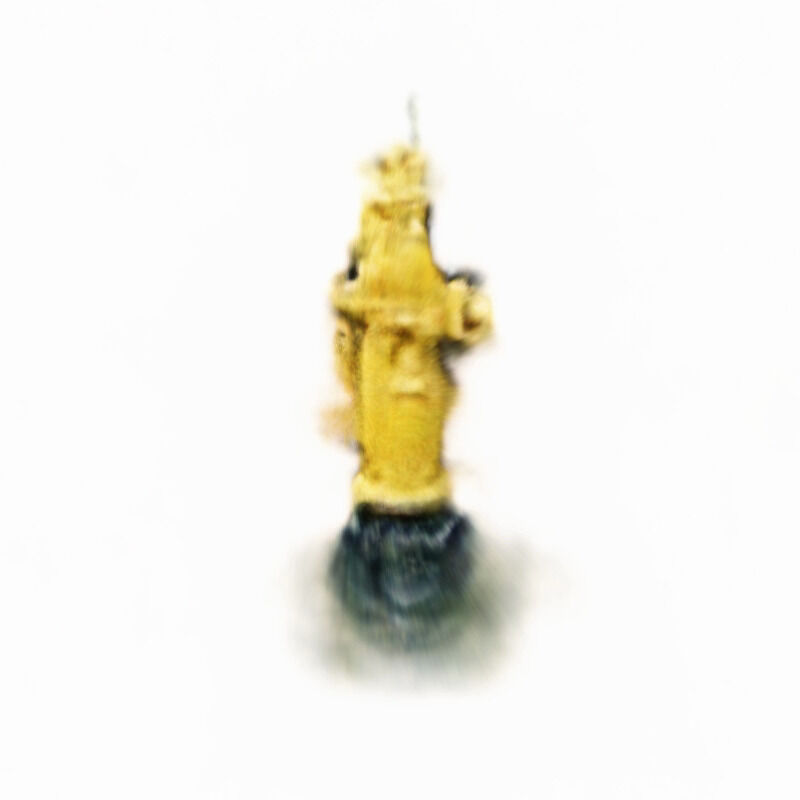}
\end{subfigure}
\begin{subfigure}{0.1 \textwidth}
\includegraphics[width=\textwidth]{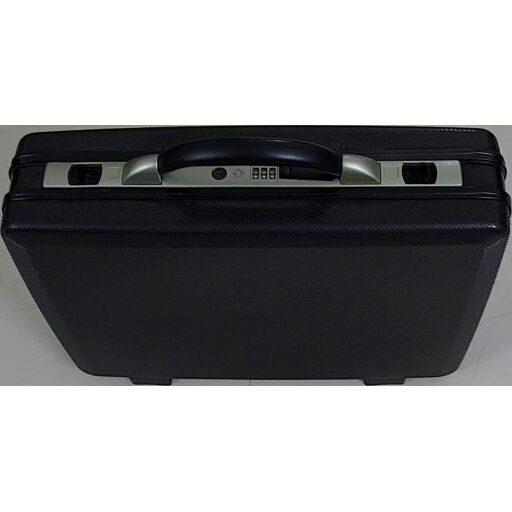}
\end{subfigure}
\begin{subfigure}{0.1 \textwidth}
\includegraphics[width=\textwidth]{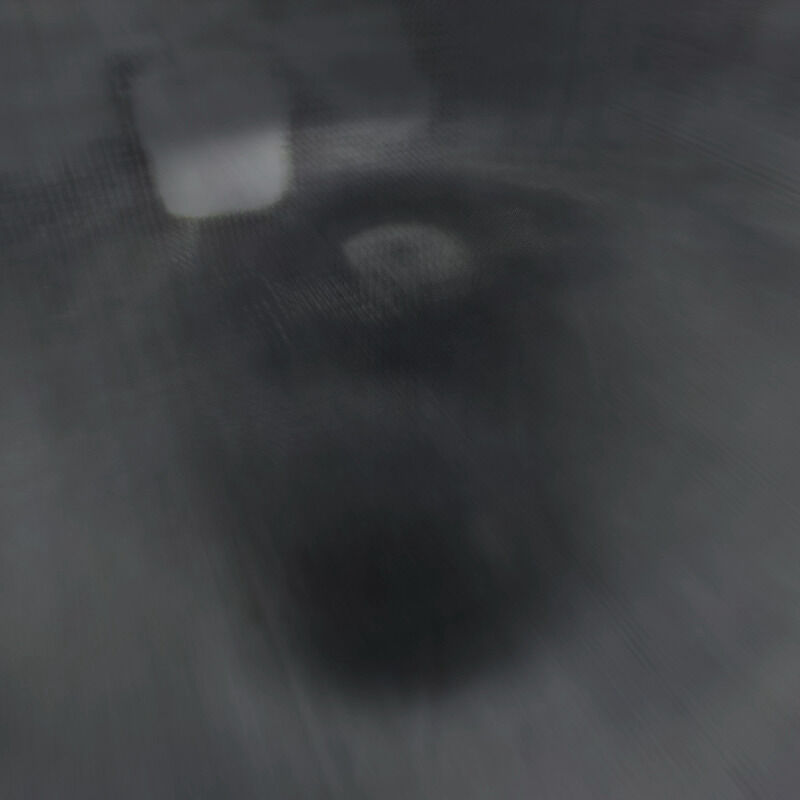}
\end{subfigure}
\begin{subfigure}{0.1 \textwidth}
\includegraphics[width=\textwidth]{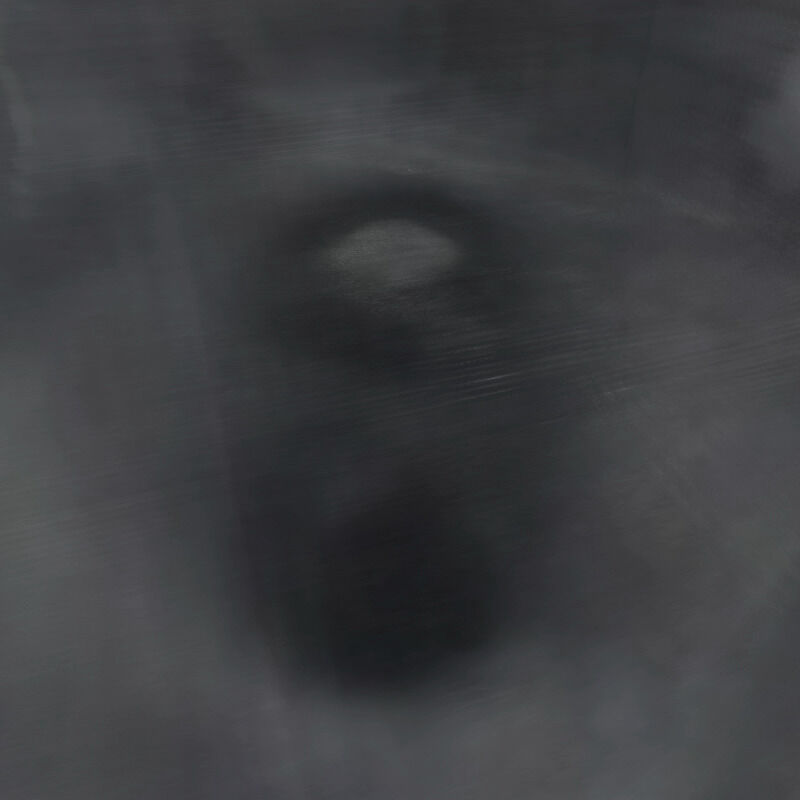}
\end{subfigure}
\begin{subfigure}{0.1 \textwidth}
\includegraphics[width=\textwidth]{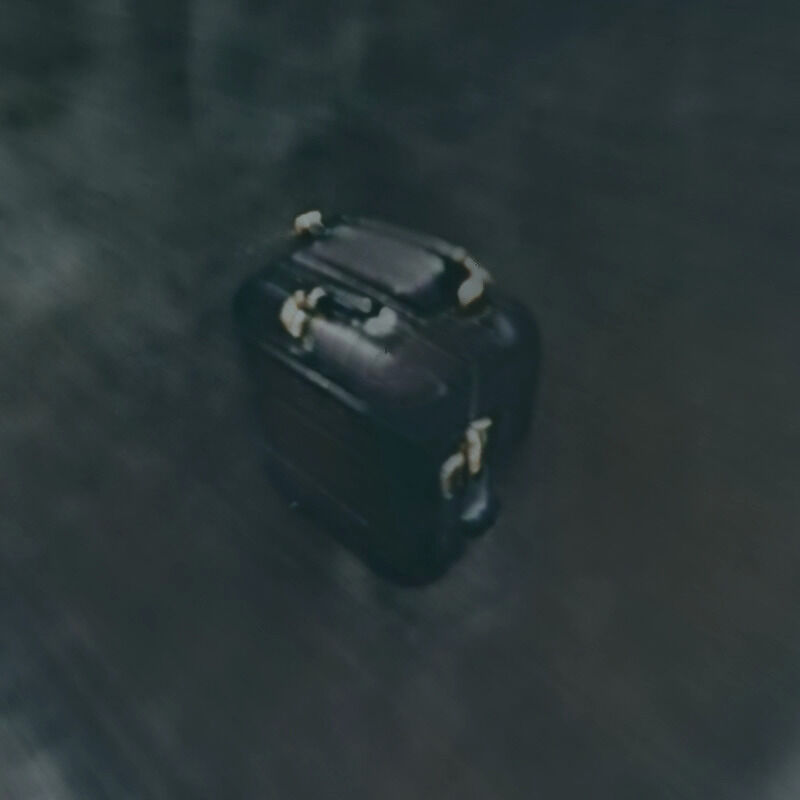}
\end{subfigure}
\begin{subfigure}{0.1 \textwidth}
\includegraphics[width=\textwidth]{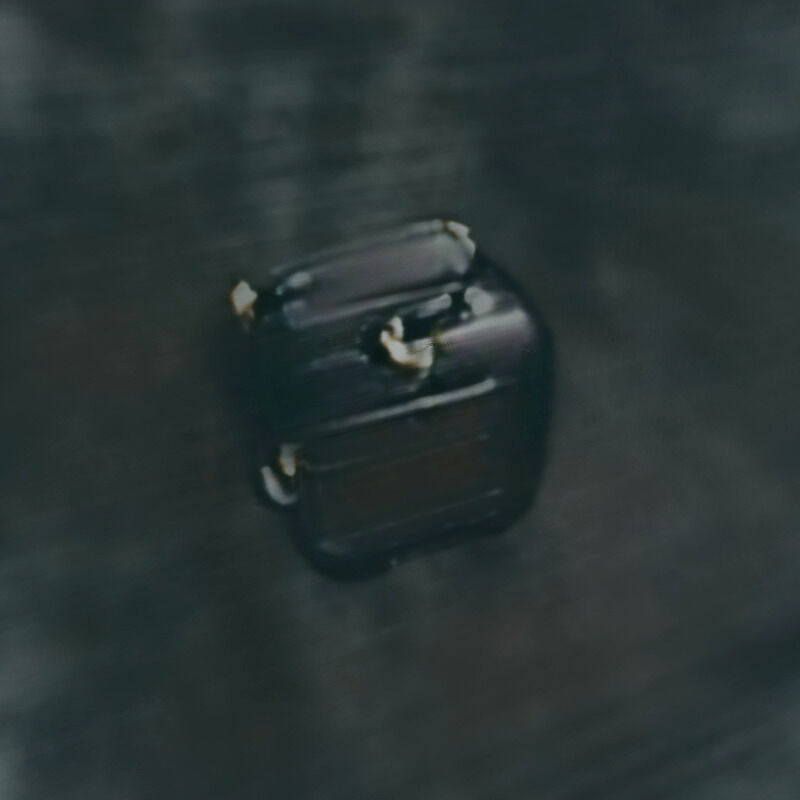}
\end{subfigure}
\begin{subfigure}{0.1 \textwidth}
\includegraphics[width=\textwidth]{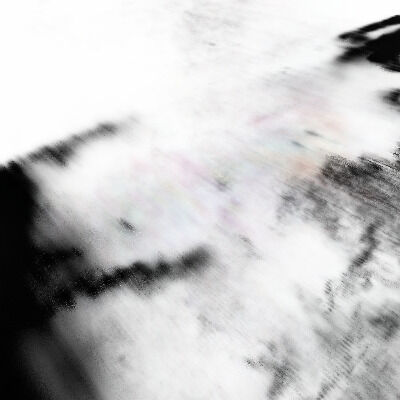}
\end{subfigure}
\begin{subfigure}{0.1 \textwidth}
\includegraphics[width=\textwidth]{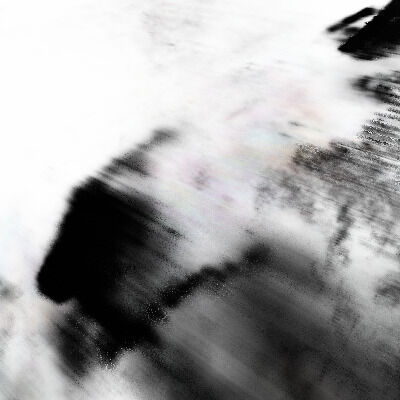}
\end{subfigure}
\begin{subfigure}{0.1 \textwidth}
\includegraphics[width=\textwidth]{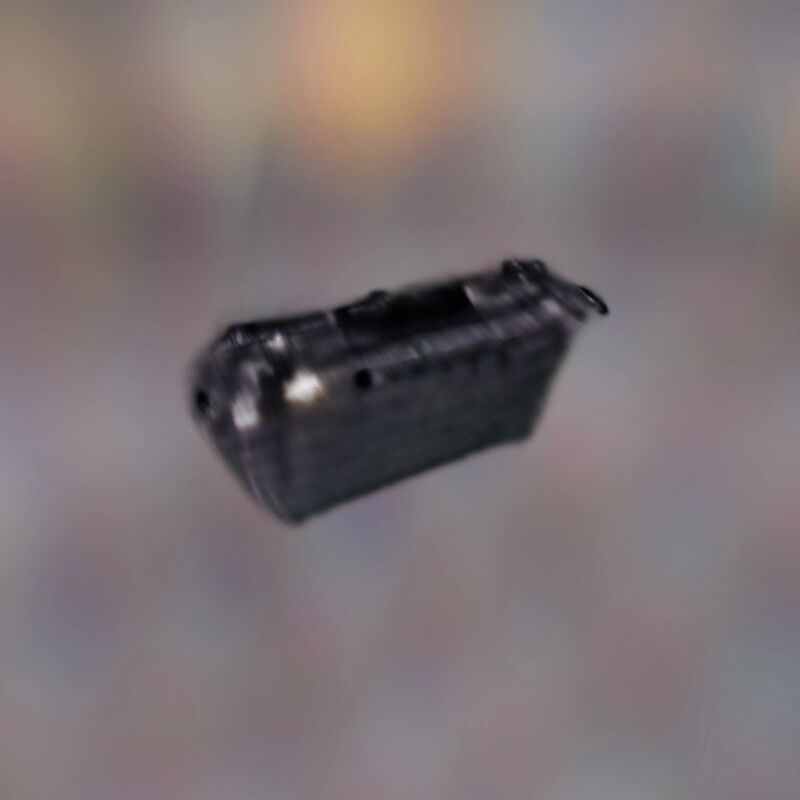}
\end{subfigure}
\begin{subfigure}{0.1 \textwidth}
\includegraphics[width=\textwidth]{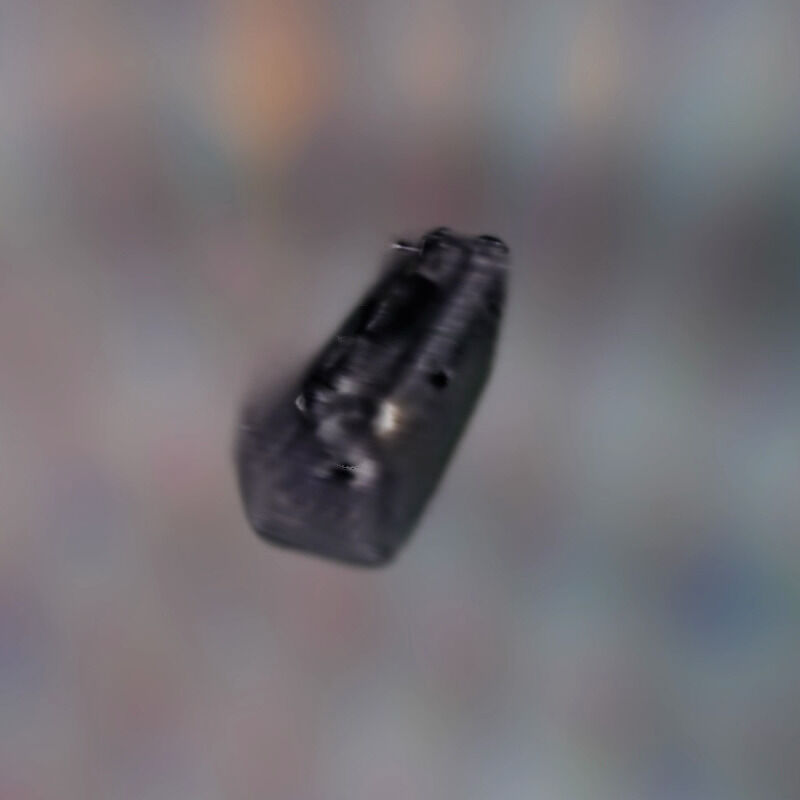}
\end{subfigure}
\\
\hspace*{.1\textwidth}
\begin{subfigure}{0.1 \textwidth}
\includegraphics[width=\textwidth]{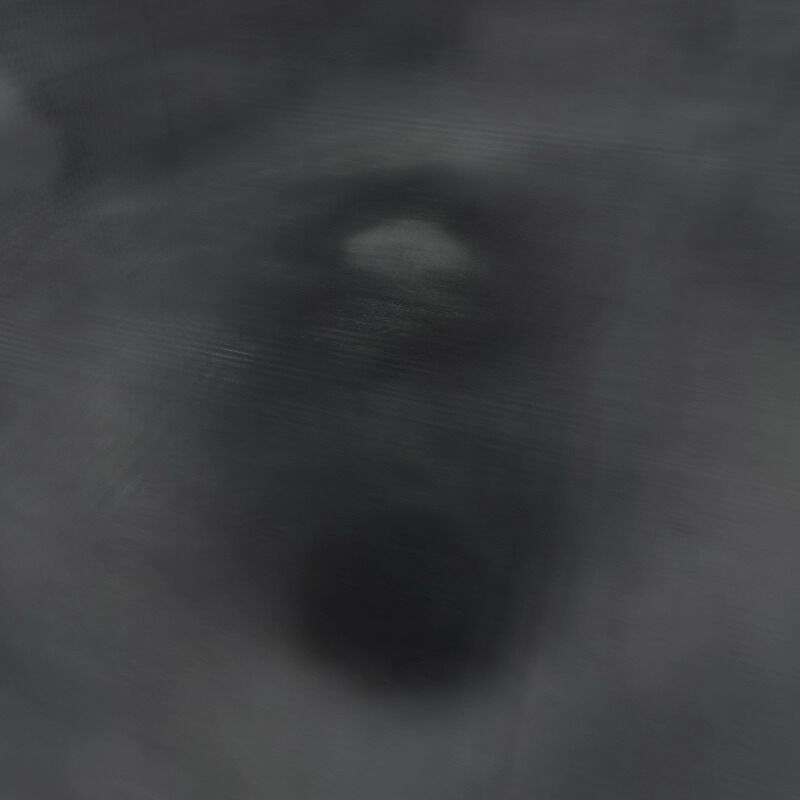}
\end{subfigure}
\begin{subfigure}{0.1 \textwidth}
\includegraphics[width=\textwidth]{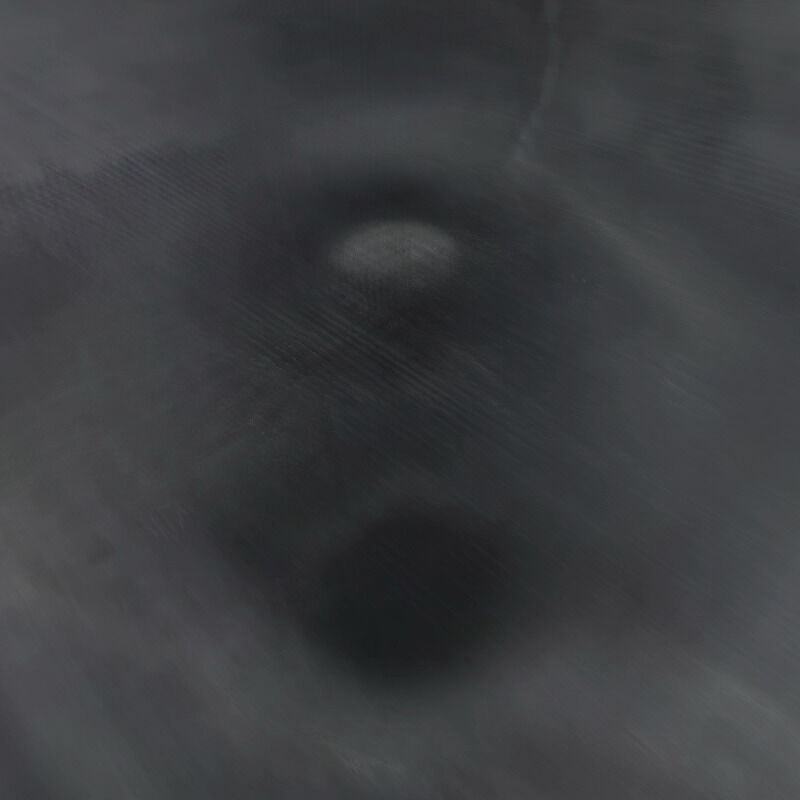}
\end{subfigure}
\begin{subfigure}{0.1 \textwidth}
\includegraphics[width=\textwidth]{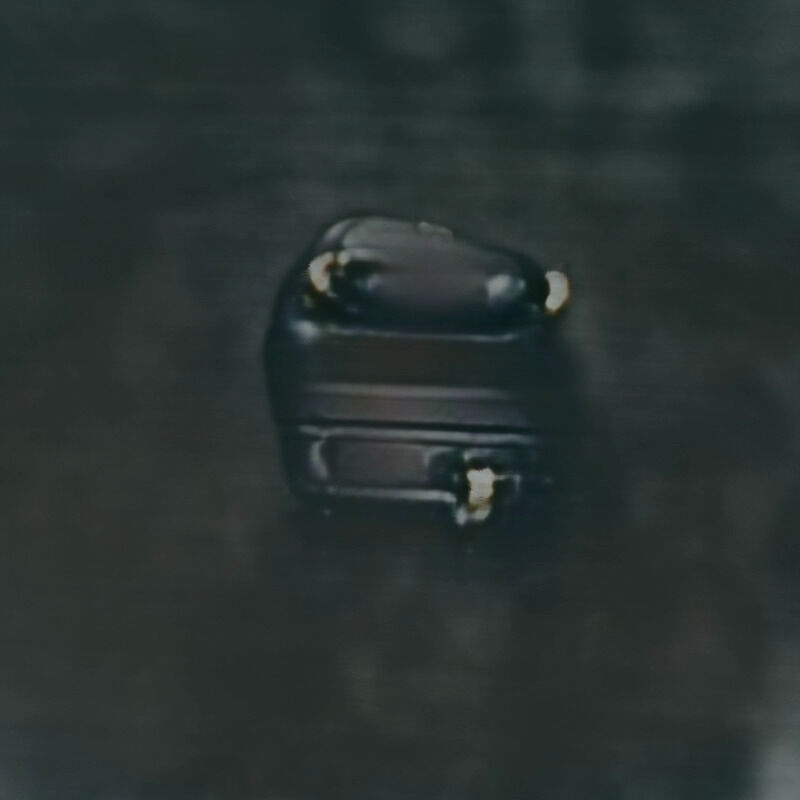}
\end{subfigure}
\begin{subfigure}{0.1 \textwidth}
\includegraphics[width=\textwidth]{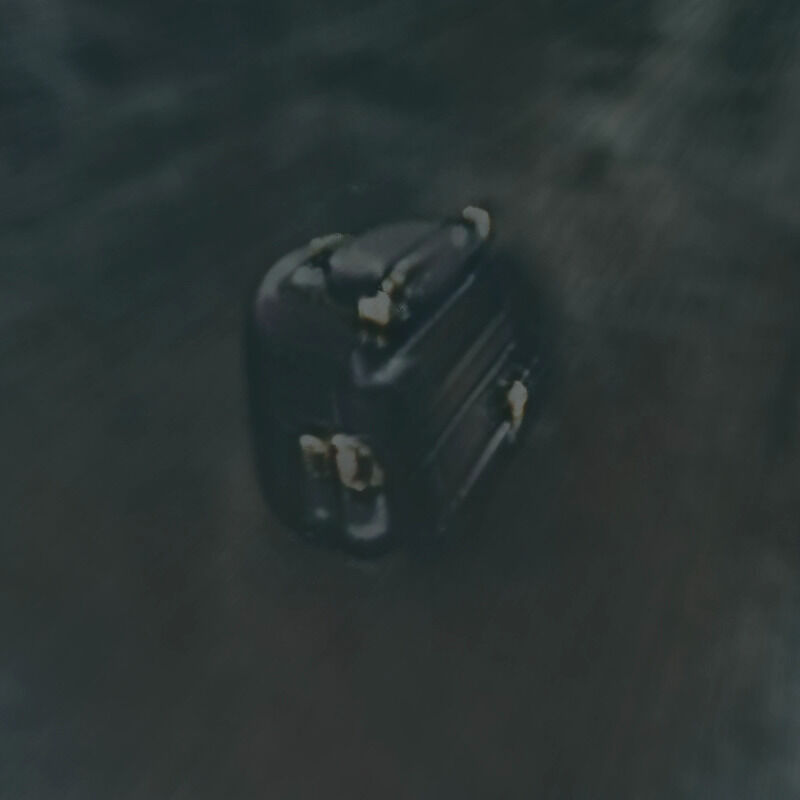}
\end{subfigure}
\begin{subfigure}{0.1 \textwidth}
\includegraphics[width=\textwidth]{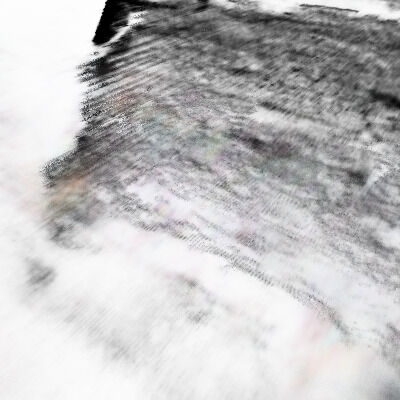}
\end{subfigure}
\begin{subfigure}{0.1 \textwidth}
\includegraphics[width=\textwidth]{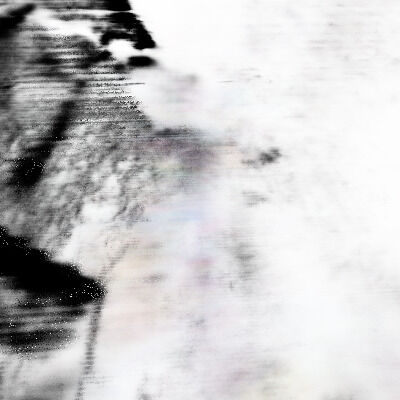}
\end{subfigure}
\begin{subfigure}{0.1 \textwidth}
\includegraphics[width=\textwidth]{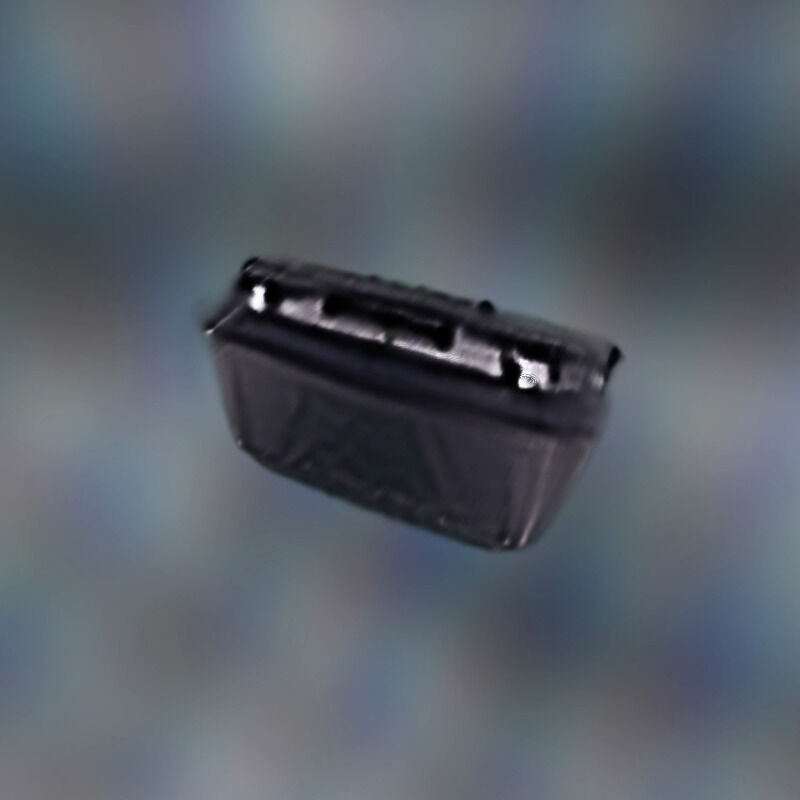}
\end{subfigure}
\begin{subfigure}{0.1 \textwidth}
\includegraphics[width=\textwidth]{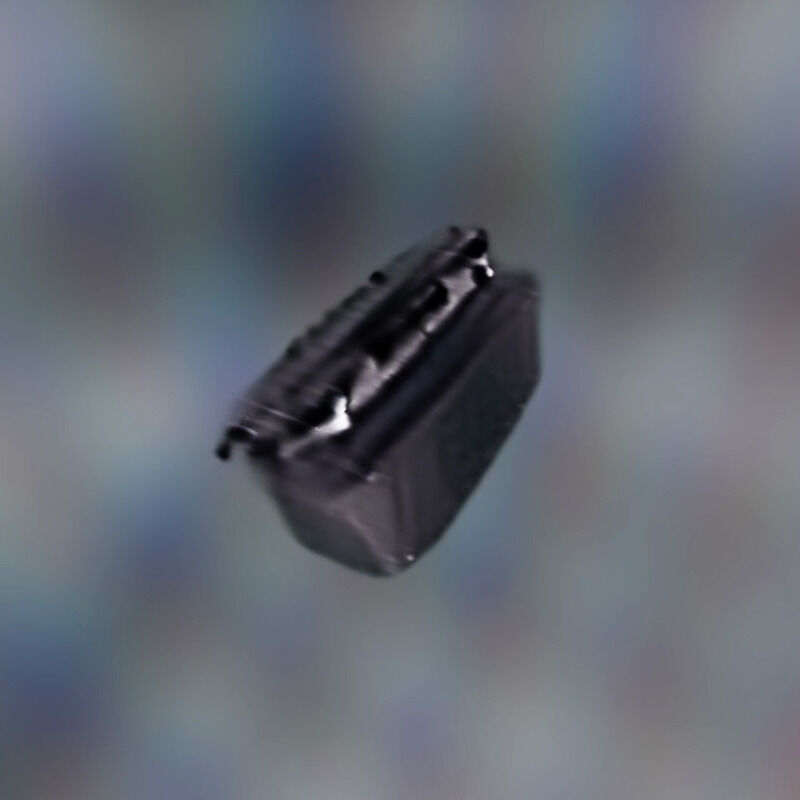}
\end{subfigure}

    \caption{Visual comparisons for novel view synthesis results. We provide \textbf{video comparisons} in supplementary.}
    \label{fig:results}
\end{figure*}

\vspace{-1mm}
\section{Experiments}
\vspace{-1mm}
In this section, we will first describe a unique training recipe that we propose to train our framework, then the implementation details.

\vspace{-1mm}
\subsection{Training Recipe}
\vspace{-1mm}
\label{sec:regularizer}

\paragraph{Diverse Camera Pose and Intrinsics Sampling}

During training, we observe that a diverse camera pose and intrinsics are essential for the model to learn a coherent 3D object. If the camera pose is not sampled from a diverse distribution, the 3D object will collapse towards the major viewing directions and thus generate unpleasant results. Specifically, we alter between two pose sampling strategies; (1) we have a fixed camera pose $\Mat{\Phi}_0$ (i.e., a Delta distribution) for the reference view, (2) we uniformly sample the camera rotation on the surface of a unit sphere and then uniformly sample the radius in $[0.4, 1.0]$ to locate the camera. We start with the Delta distribution, and as training progresses, we gradually shift to the other sophisticated camera pose distribution. The camera intrinsics provide additional augmentation on the viewing direction. By uniformly sampling the field of view in $[50, 70]$, we zoom in and out to get different resolution projections for the scene. We also 
add Gaussian noise to the camera pose sampling phase. Such perturbations help us render more diverse images during training.

\paragraph{Foreground Aware Diffusion Prior}

\looseness=-1
ERNIE-ViLG~\cite{feng2022ernie} utilizes a weighting mechanism during the training of diffusion models to let the model generate.
Using an additional object detection model, their weighted loss forces the model to focus on salient regions.
Similarly, we hope to strengthen the guidance on the foreground object region. We propose to weight the score matching distance with the predicted density. 
Specifically, we penalize the regions with higher density so that our NeRF gains more guidance on the foreground regions. Hence, for each rendered image, we first obtain the bounding box for the non-zero density regions. Then, we extend the $w(t)$ in  $\mathcal{L}_\text{diff}$ (Eq.~\ref{eqn:final_loss}) into a spatial function:
\begin{equation}
W_{i,j} = \begin{cases}
2, & \text{if } (i, j) \text{ is inside the bounding box} \\
1, & \text{otherwise}.
\end{cases}
\end{equation}

\paragraph{Geometry Regularization}
We also include a series of geometry regularizations to avoid undesired artifacts in our 3D object. 
To avoid foggy and semi-transparent artifacts at the back of the scene, we utilize the orient loss from Ref-NeRF~\cite{verbin2022ref} to prevent the surface normals from facing backward from the camera.
We also incorporate depth smoothness loss~\cite{wang2018learning}, sparsity loss, and distortion loss~\cite{SunSC22_2,barron2022mip} to avoid the NeRF filling empty space with floaters.
Full details for these regularizations are provided in the supplementary material.

\vspace{-3mm}
\paragraph{Timestep Annealing}
At the start of training, the quality of the rendered images is poor, and we need the diffusion prior to provide more thorough guidance. As a result, we need to add more noise to perturb the rendered image. As the training progresses, the quality of rendered images gradually improves, and we choose to reduce the timestep range accordingly. By annealing the timestep where we add noise, we are lowering the noise level of the perturbation, and the noise estimate becomes more accurate.

\begin{table*}[]
\centering
\caption{Quantitative evaluation by comparing the proposed method with previous state-of-the-art approaches.}
\resizebox{0.9\textwidth}{!}{
\begin{tabular}{l|cccccccc|c}
\hline
CLIP Distance $\downarrow$ & Apple  & Bench  & Baseball & Hydrant & Remote & Suitcase & Statue & Sandwich & Average \\ \hline
DSNeRF~\cite{deng2022depth}                     & 0.5309 & \cellcolor{orange!25} 0.5447 & \cellcolor{yellow!25} 0.5506   & \cellcolor{yellow!25} 0.6226  & \cellcolor{yellow!25} 0.6441 & \cellcolor{yellow!25}0.5533   & 0.5992 & 0.6110   & \cellcolor{yellow!25} 0.5821       \\
DietNeRF~\cite{jain2021putting}                   & \cellcolor{orange!25} 0.4245 & \cellcolor{yellow!25} 0.5579 & \cellcolor{orange!25} 0.4545   & \cellcolor{orange!25} 0.5190  &\cellcolor{orange!25} 0.5718 & \cellcolor{orange!25} 0.4890 & \cellcolor{yellow!25}   0.5947 & \cellcolor{orange!25} 0.5179   & \cellcolor{orange!25} 0.5162       \\
SinNeRF~\cite{xu2022sinnerf}                    & \cellcolor{yellow!25}  0.5192 & 0.5769 & 0.6230   & 0.6343  & 0.6496 & 0.5705 & \cellcolor{orange!25} 0.5605 & \cellcolor{yellow!25} 0.6015   &  0.5918       \\
\textbf{NeuralLift-360 (Ours)}               & \cellcolor{red!25} 0.3914 & \cellcolor{red!25} 0.4785 & \cellcolor{red!25} 0.3974   & \cellcolor{red!25} 0.5146  & \cellcolor{red!25} 0.4916 & \cellcolor{red!25} 0.4186   & \cellcolor{red!25} 0.4411       & \cellcolor{red!25} 0.4655   &  \cellcolor{red!25} 0.4498      \\ \hline
\end{tabular}
}

\vspace{-2mm}
\label{tab:clip}
\end{table*}

\begin{figure*}
    \centering
   \begin{tabular}{P{0.14\textwidth}P{0.14\textwidth}P{0.2\textwidth}P{0.2\textwidth}P{0.18\textwidth}}
    \small Reference& \small w/o $\mathcal{L}_\text{diff}$ & \small w/o fine-tuning  & \small w/o $\mathcal{L}_\text{ranking}$ & \small Full Model\\
    \end{tabular}
\begin{subfigure}{0.1 \textwidth}
\includegraphics[width=\textwidth]{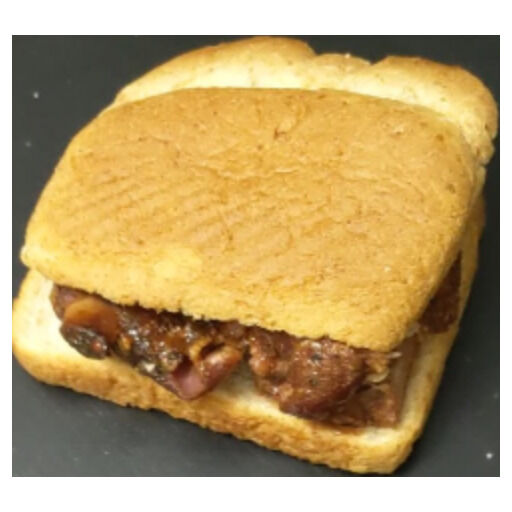}
\end{subfigure}
\begin{subfigure}{0.1 \textwidth}
\includegraphics[width=\textwidth]{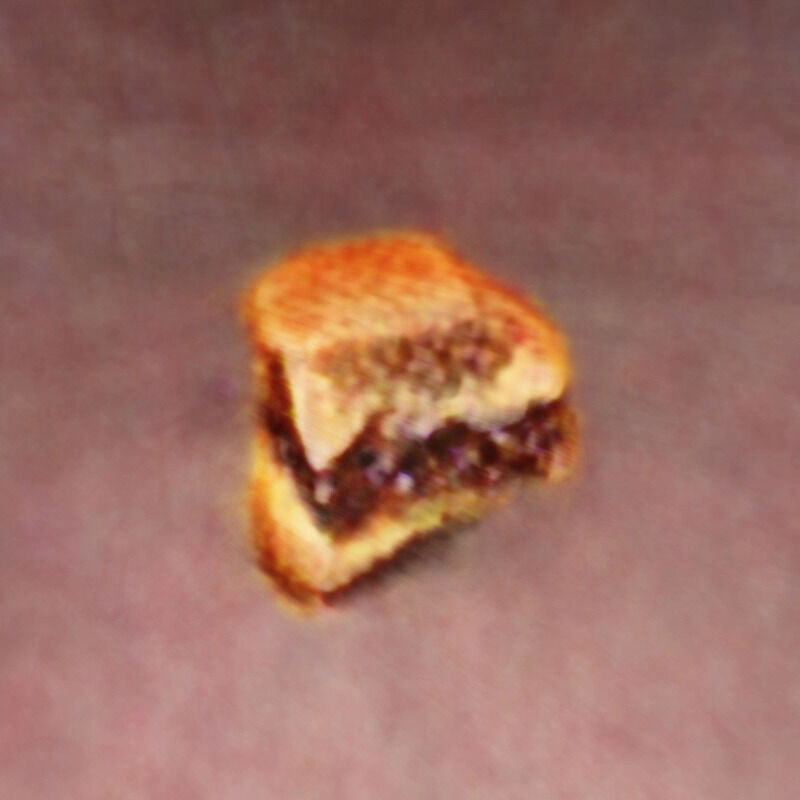}
\end{subfigure}
\begin{subfigure}{0.1 \textwidth}
\includegraphics[width=\textwidth]{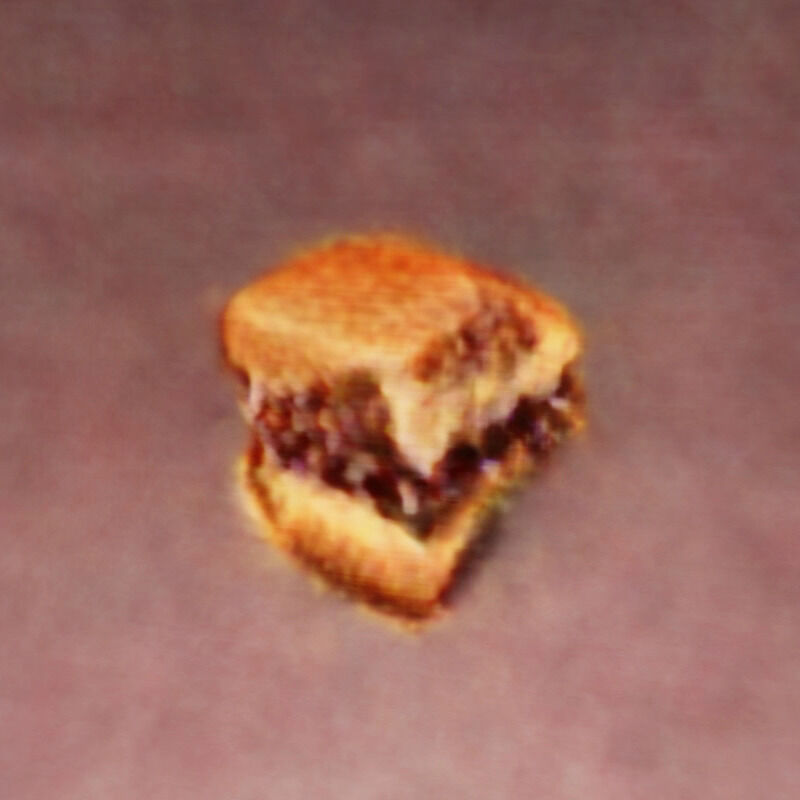}
\end{subfigure}
\begin{subfigure}{0.1 \textwidth}
\includegraphics[width=\textwidth]{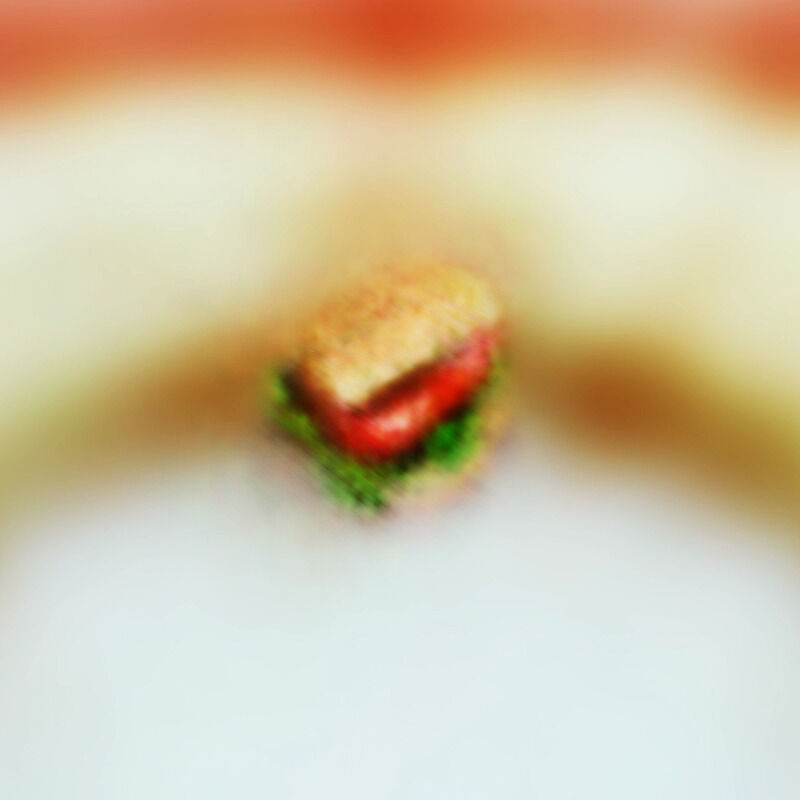}
\end{subfigure}
\begin{subfigure}{0.1 \textwidth}
\includegraphics[width=\textwidth]{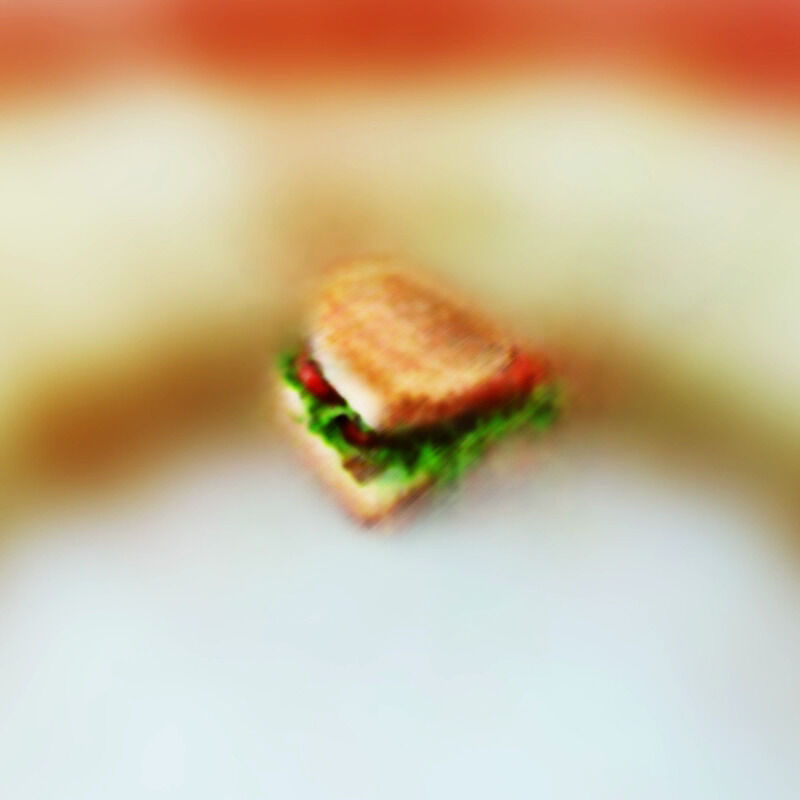}
\end{subfigure}
\begin{subfigure}{0.1 \textwidth}
\includegraphics[width=\textwidth]{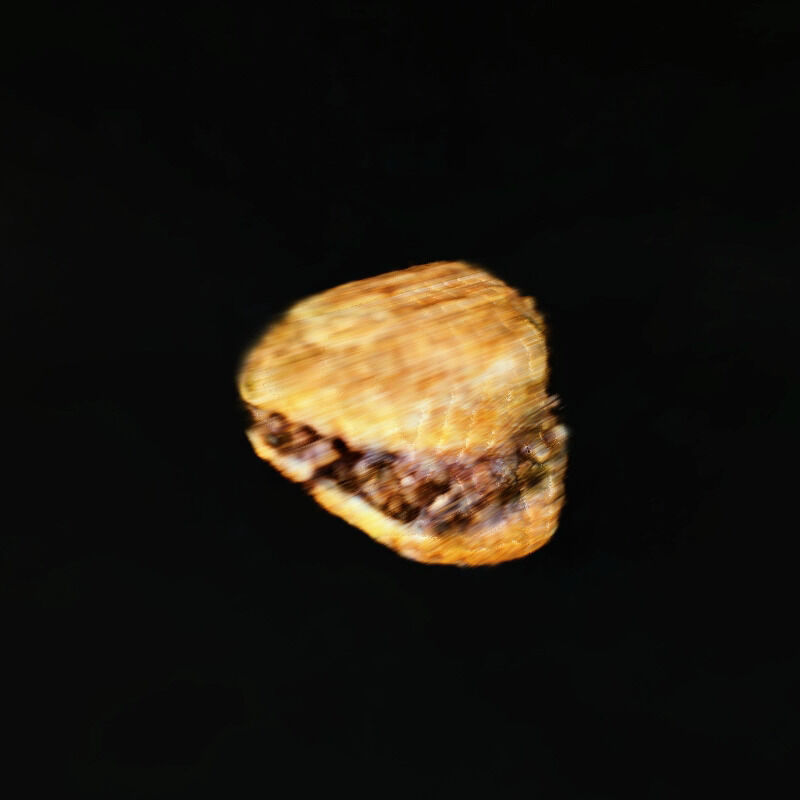}
\end{subfigure}
\begin{subfigure}{0.1 \textwidth}
\includegraphics[width=\textwidth]{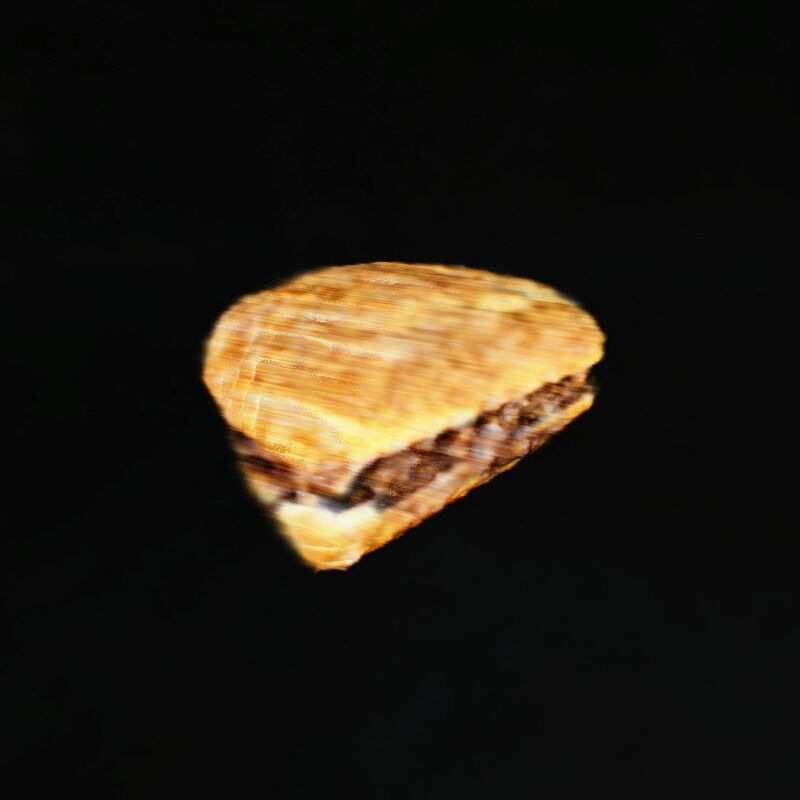}
\end{subfigure}
\begin{subfigure}{0.1 \textwidth}
\includegraphics[width=\textwidth]{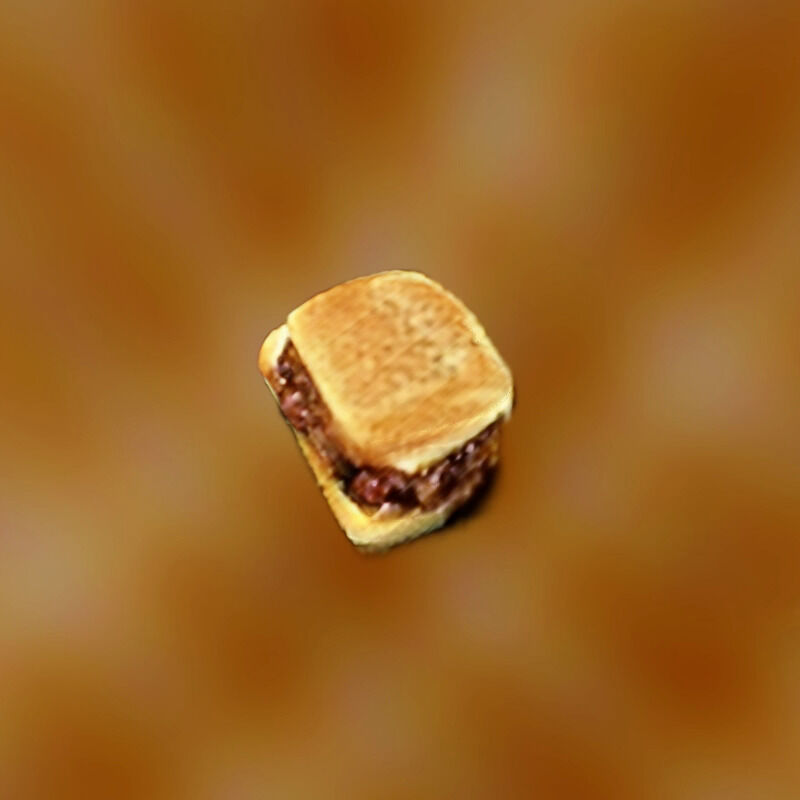}
\end{subfigure}
\begin{subfigure}{0.1 \textwidth}
\includegraphics[width=\textwidth]{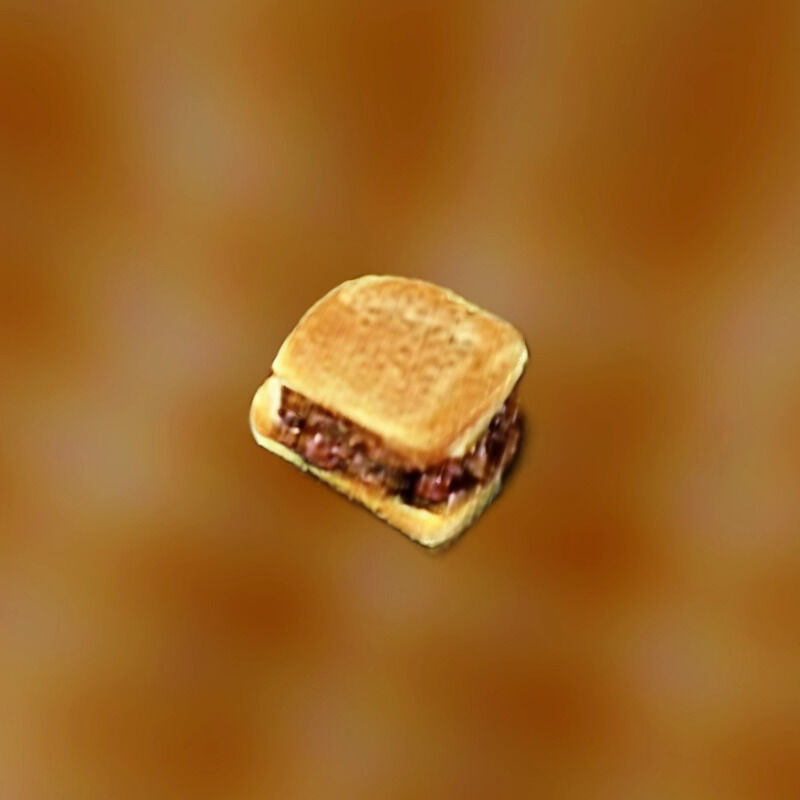}
\end{subfigure}
\\
\hspace*{.1\textwidth}
\begin{subfigure}{0.1 \textwidth}
\includegraphics[width=\textwidth]{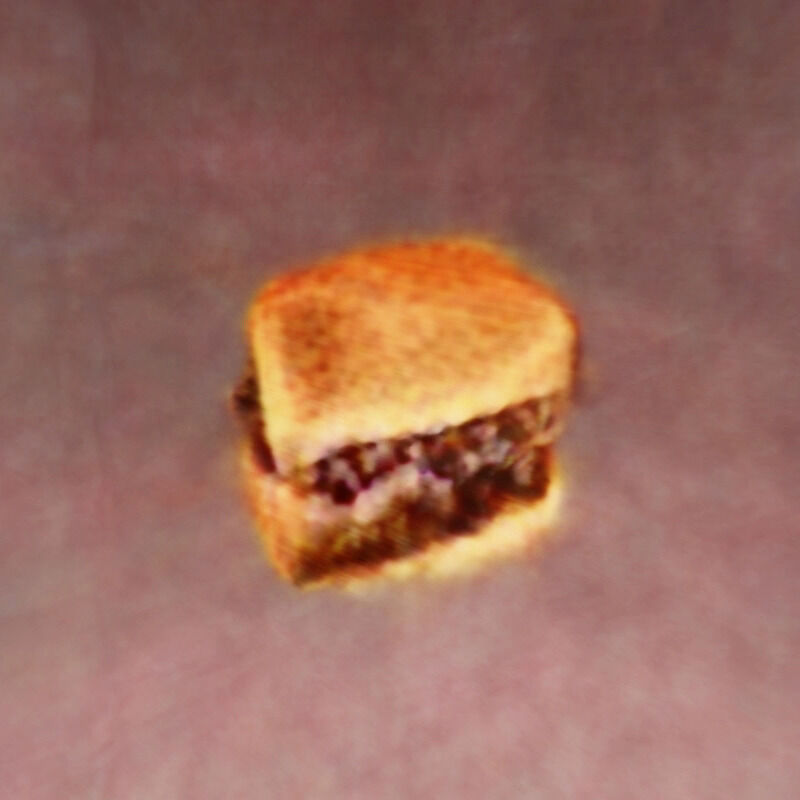}
\end{subfigure}
\begin{subfigure}{0.1 \textwidth}
\includegraphics[width=\textwidth]{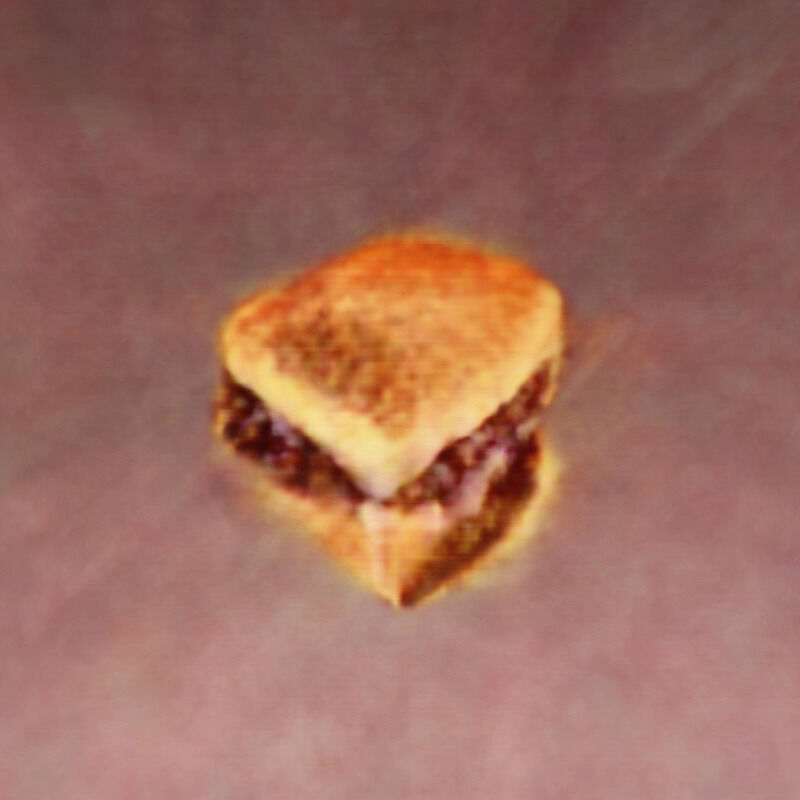}
\end{subfigure}
\begin{subfigure}{0.1 \textwidth}
\includegraphics[width=\textwidth]{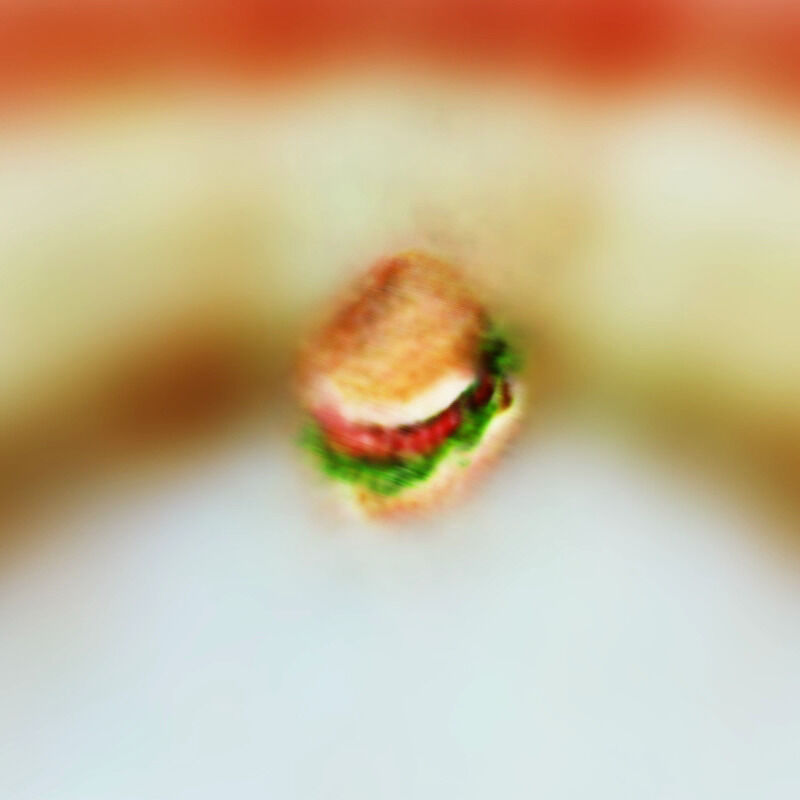}
\end{subfigure}
\begin{subfigure}{0.1 \textwidth}
\includegraphics[width=\textwidth]{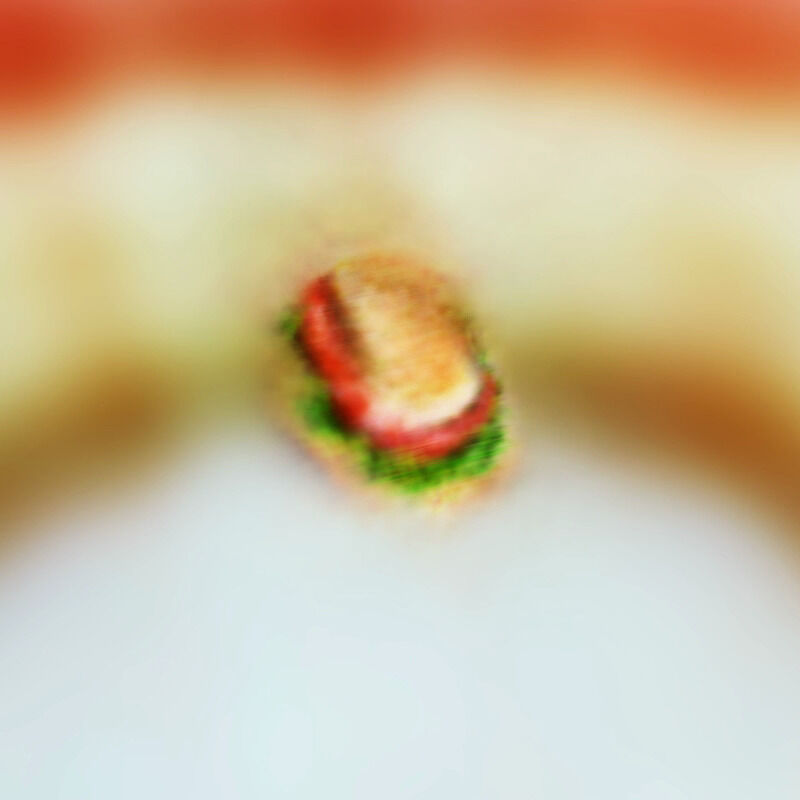}
\end{subfigure}
\begin{subfigure}{0.1 \textwidth}
\includegraphics[width=\textwidth]{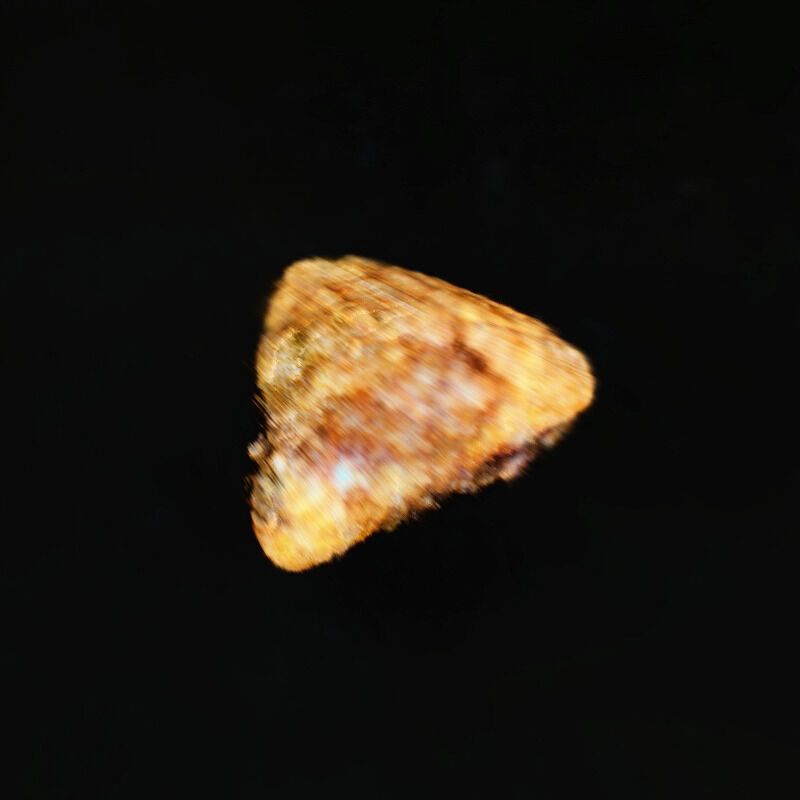}
\end{subfigure}
\begin{subfigure}{0.1 \textwidth}
\includegraphics[width=\textwidth]{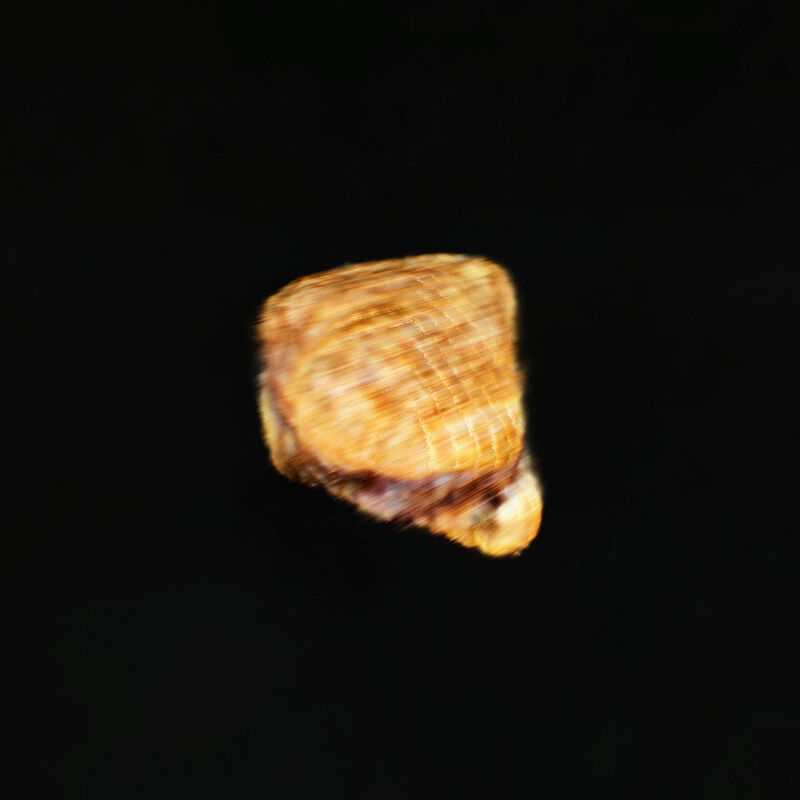}
\end{subfigure}
\begin{subfigure}{0.1 \textwidth}
\includegraphics[width=\textwidth]{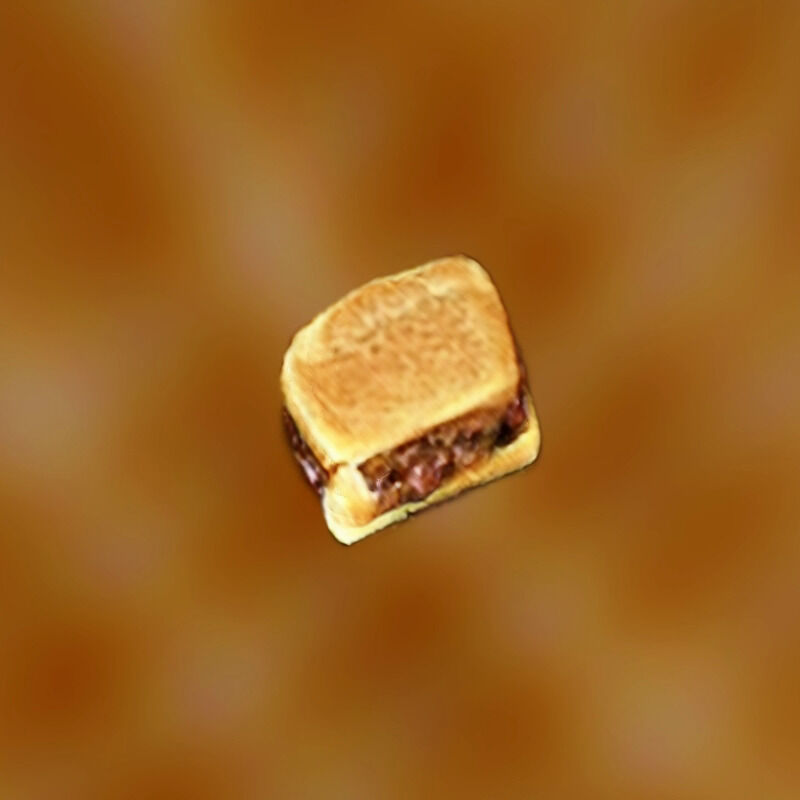}
\end{subfigure}
\begin{subfigure}{0.1 \textwidth}
\includegraphics[width=\textwidth]{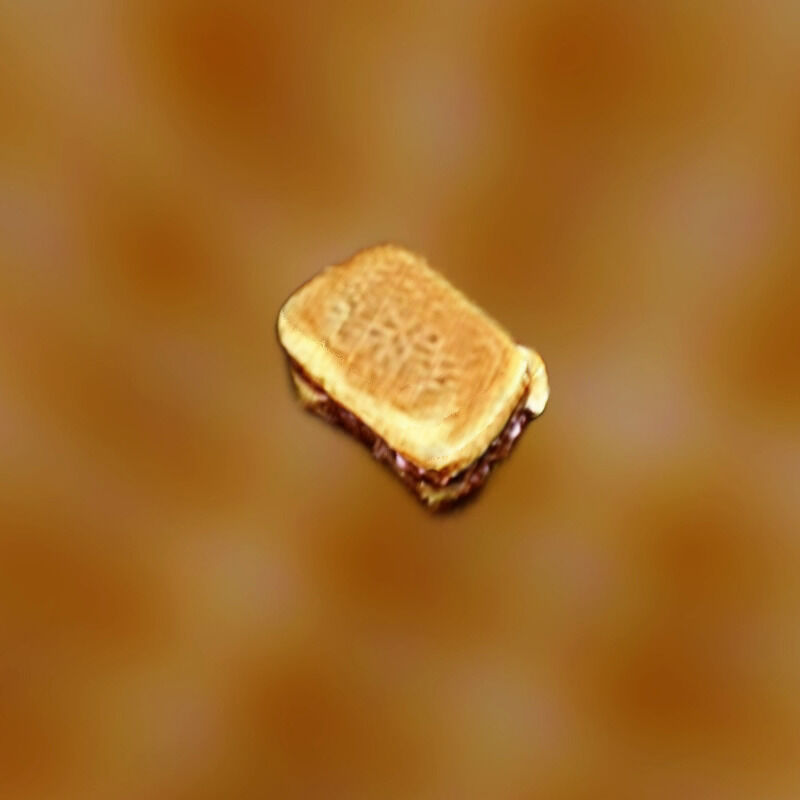}
\end{subfigure}
\vspace{-1mm}
\caption{Ablation study on the baseline variants. Given a single reference image, we run different variants to render its 360 $^\circ$ views.}
\vspace{-4mm}
\label{fig:ablation_sd}
\end{figure*}

\begin{figure}
    \centering
   \begin{tabular}{P{0.1\textwidth}P{0.15\textwidth}P{0.15\textwidth}}
    \small Reference& \small w/o CLIP guidance  & \small Full Model\\
    \end{tabular}
\begin{subfigure}{0.08 \textwidth}
\includegraphics[width=\textwidth]{images/reference/remote.jpg}
\end{subfigure}
\begin{subfigure}{0.08 \textwidth}
\includegraphics[width=\textwidth]{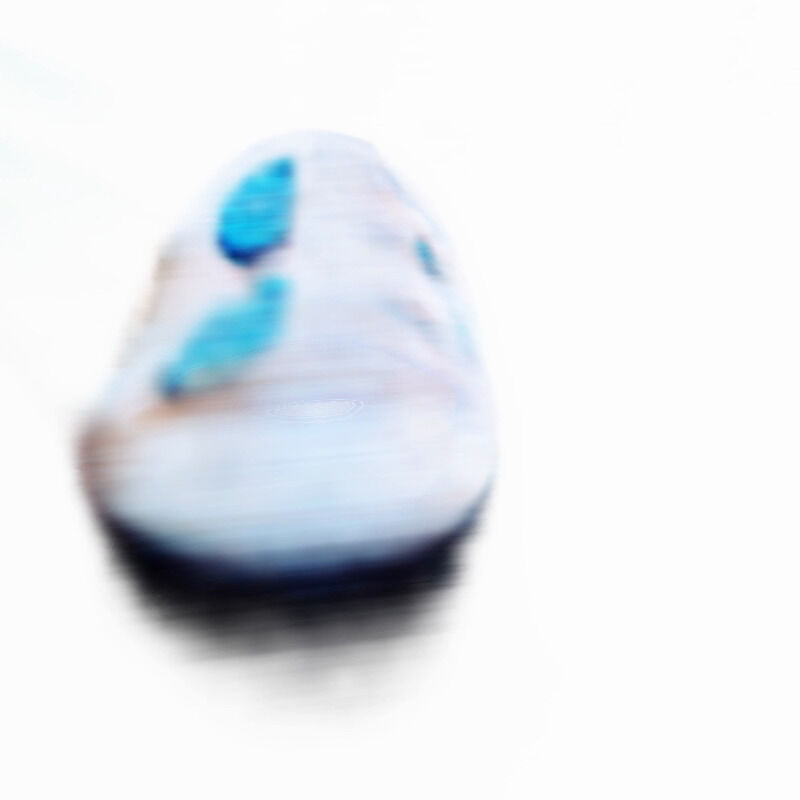}
\end{subfigure}
\begin{subfigure}{0.08 \textwidth}
\includegraphics[width=\textwidth]{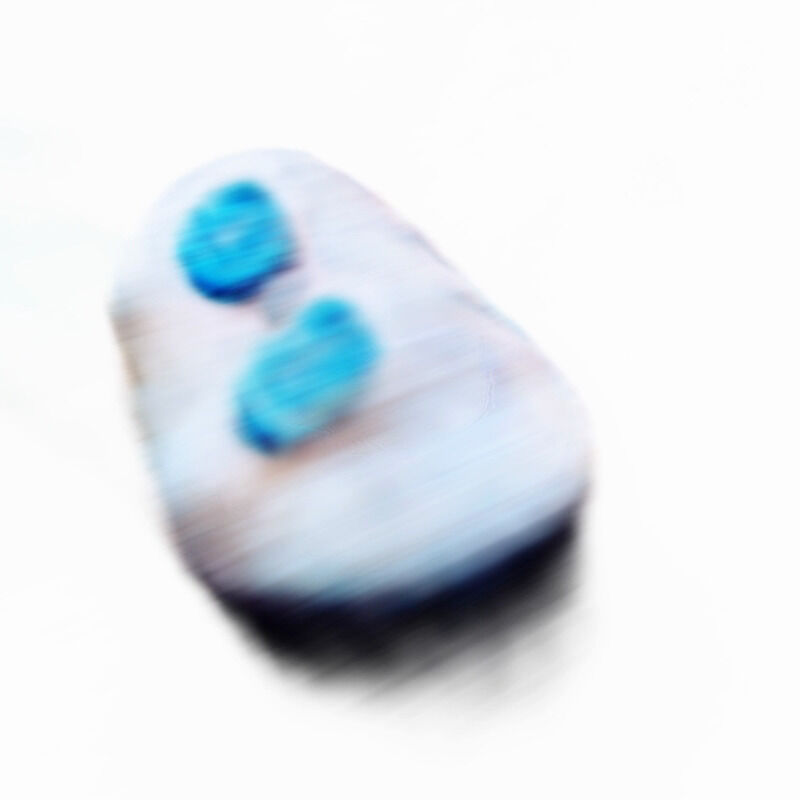}
\end{subfigure}
\begin{subfigure}{0.08 \textwidth}
\includegraphics[width=\textwidth]{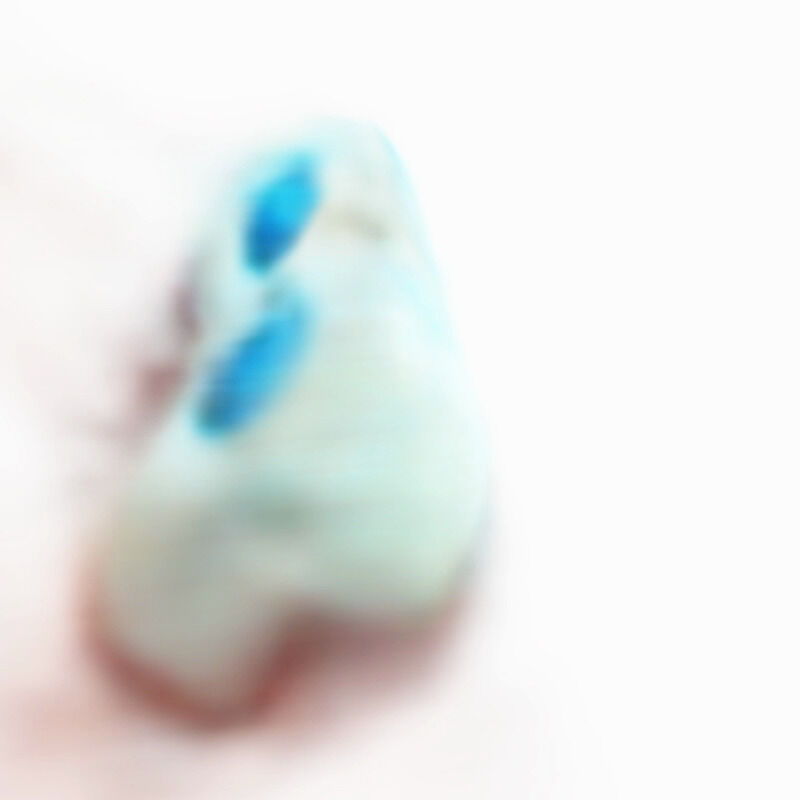}
\end{subfigure}
\begin{subfigure}{0.08 \textwidth}
\includegraphics[width=\textwidth]{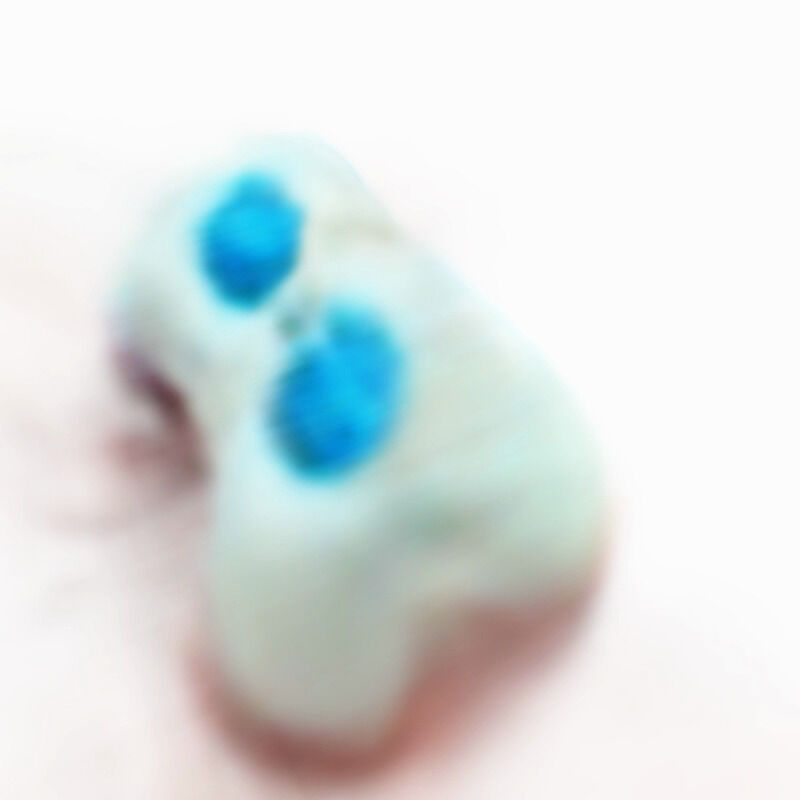}
\end{subfigure}
\\
\hspace*{0.08\textwidth}
\begin{subfigure}{0.08 \textwidth}
\includegraphics[width=\textwidth]{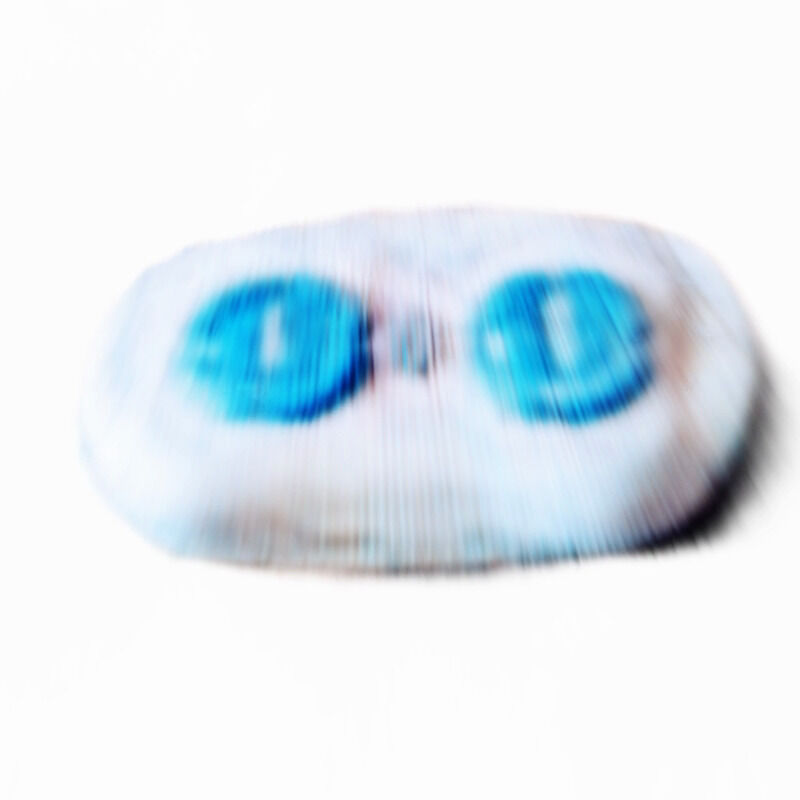}
\end{subfigure}
\begin{subfigure}{0.08 \textwidth}
\includegraphics[width=\textwidth]{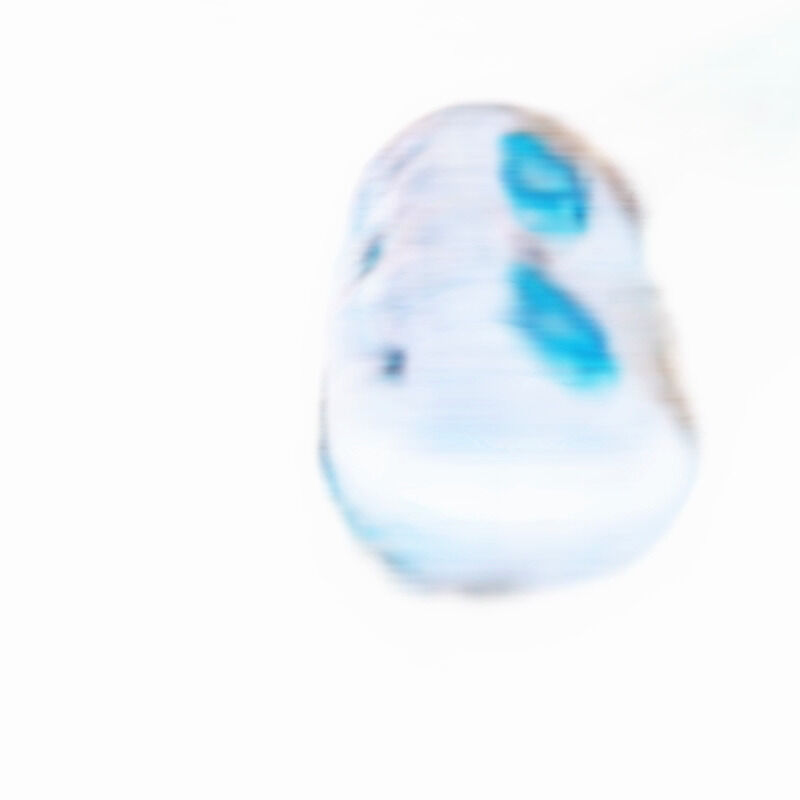}
\end{subfigure}
\begin{subfigure}{0.08 \textwidth}
\includegraphics[width=\textwidth]{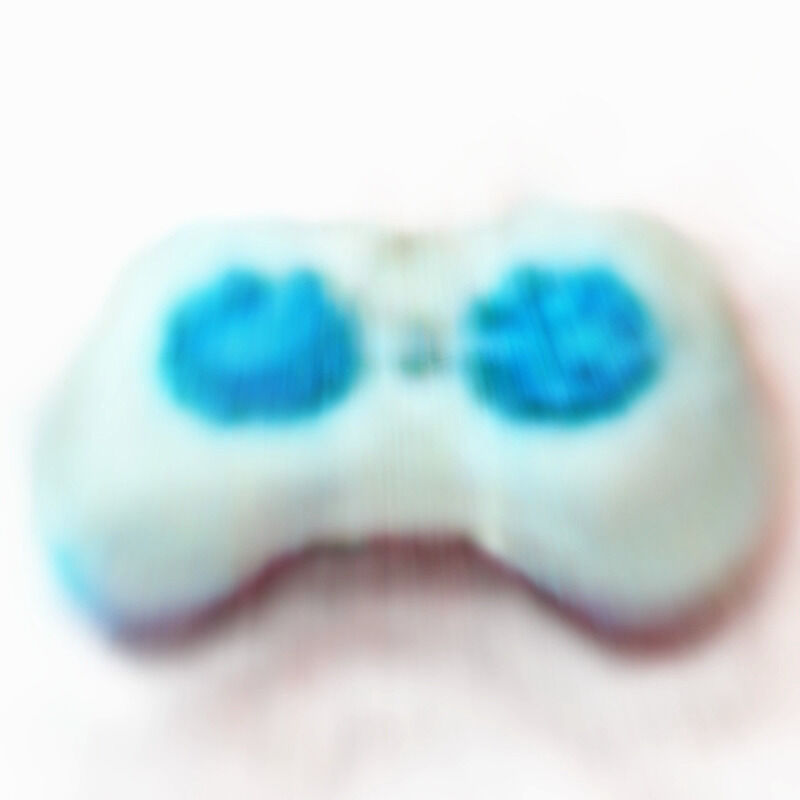}
\end{subfigure}
\begin{subfigure}{0.08 \textwidth}
\includegraphics[width=\textwidth]{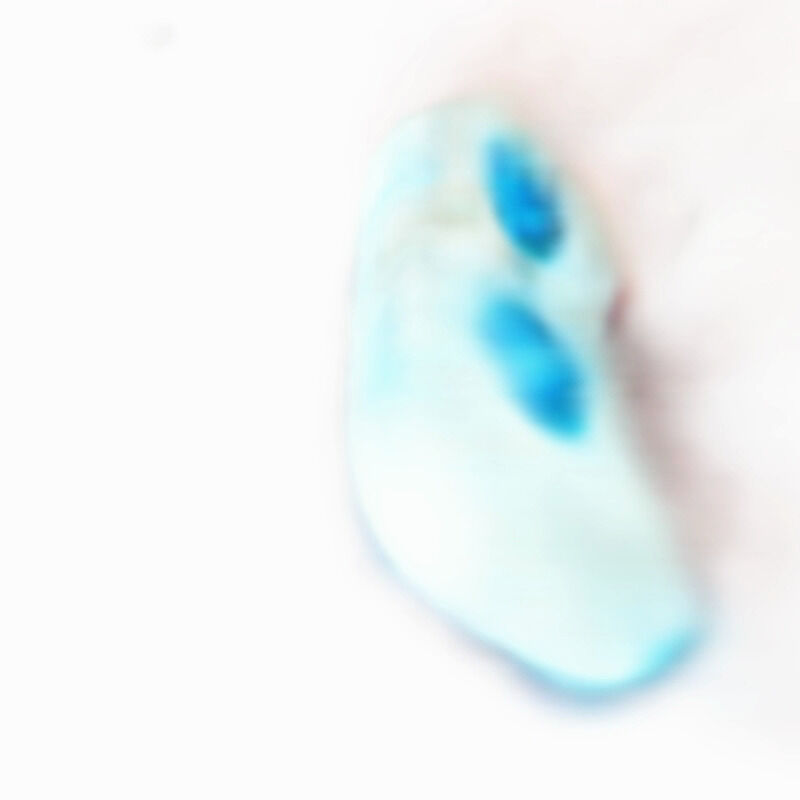}
\end{subfigure}
\begin{subfigure}{0.08 \textwidth}
\includegraphics[width=\textwidth]{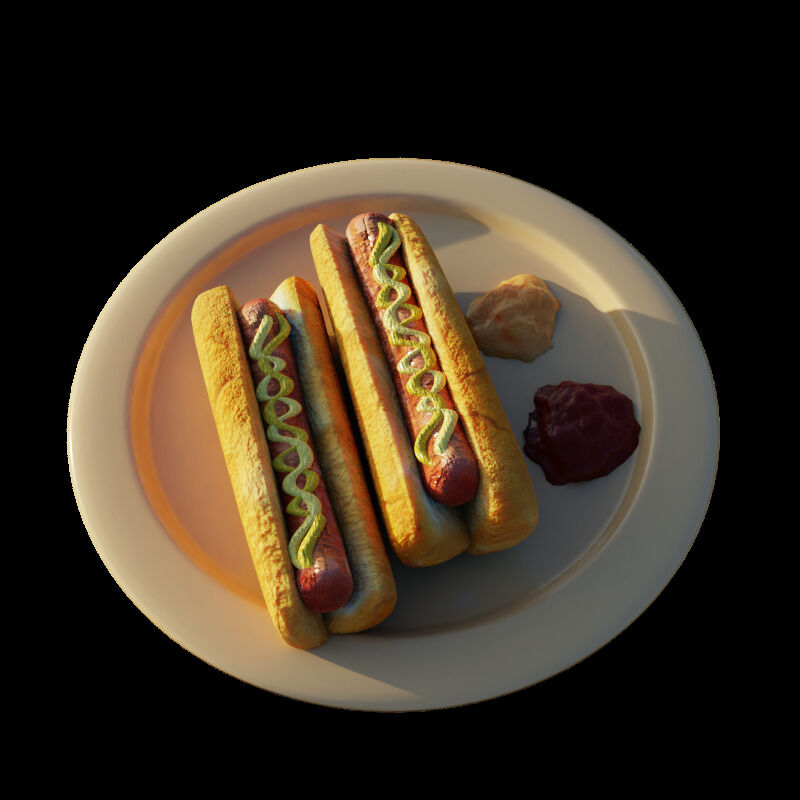}
\end{subfigure}
\begin{subfigure}{0.08 \textwidth}
\includegraphics[width=\textwidth]{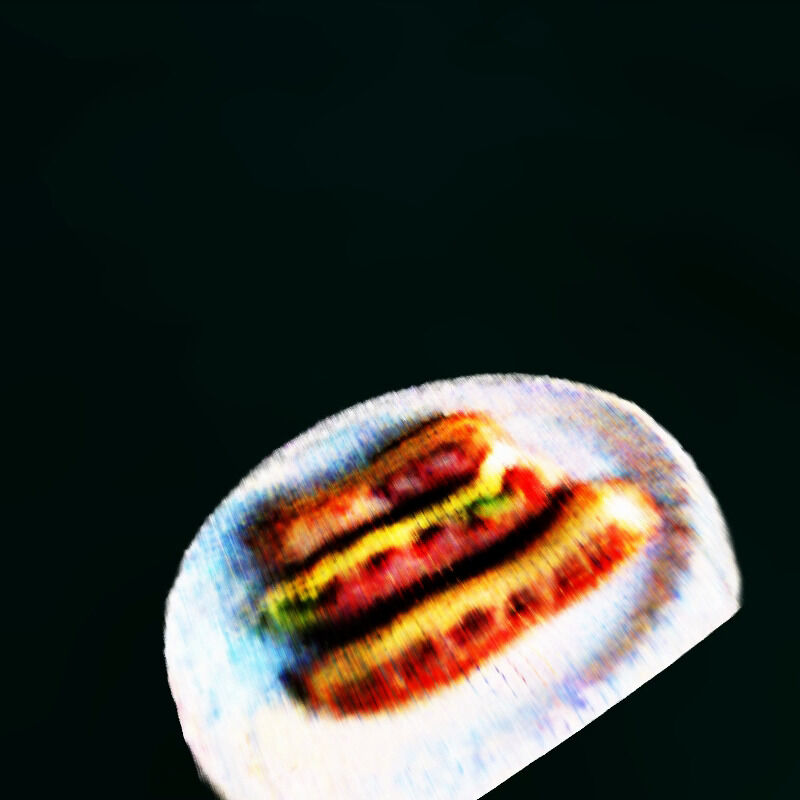}
\end{subfigure}
\begin{subfigure}{0.08 \textwidth}
\includegraphics[width=\textwidth]{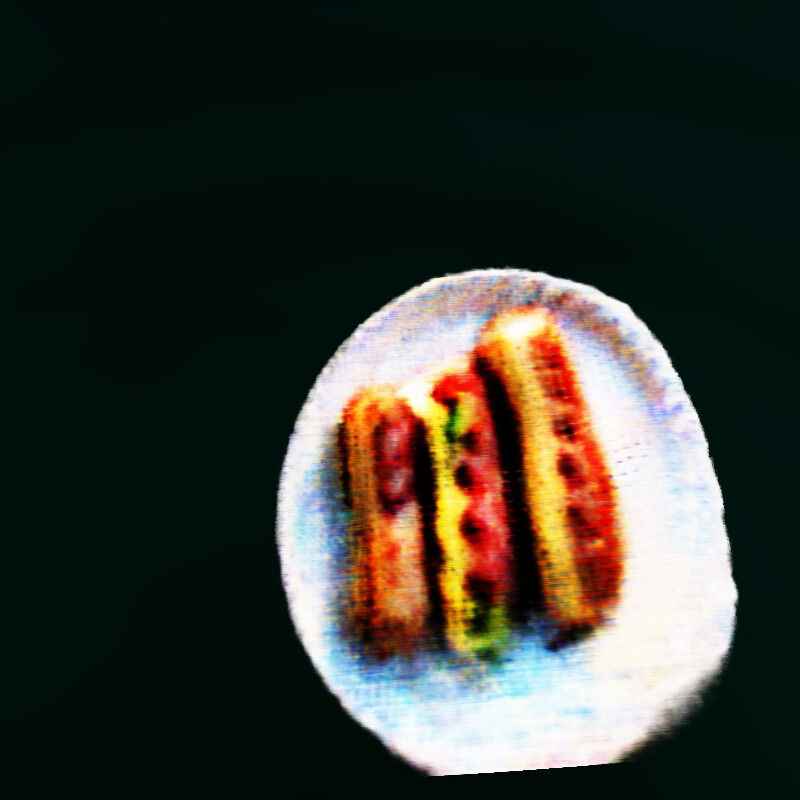}
\end{subfigure}
\begin{subfigure}{0.08 \textwidth}
\includegraphics[width=\textwidth]{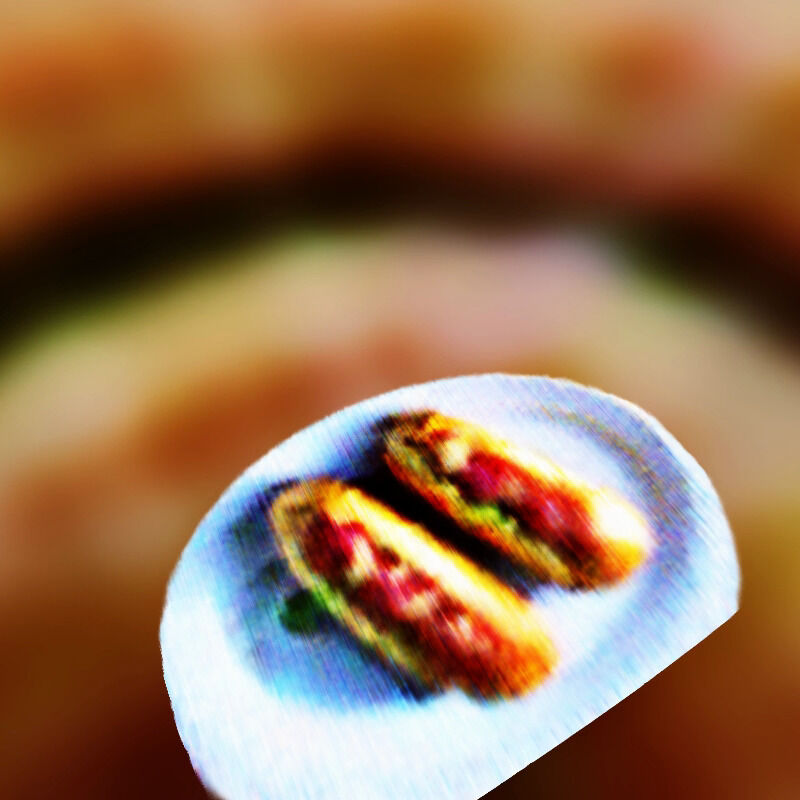}
\end{subfigure}
\begin{subfigure}{0.08 \textwidth}
\includegraphics[width=\textwidth]{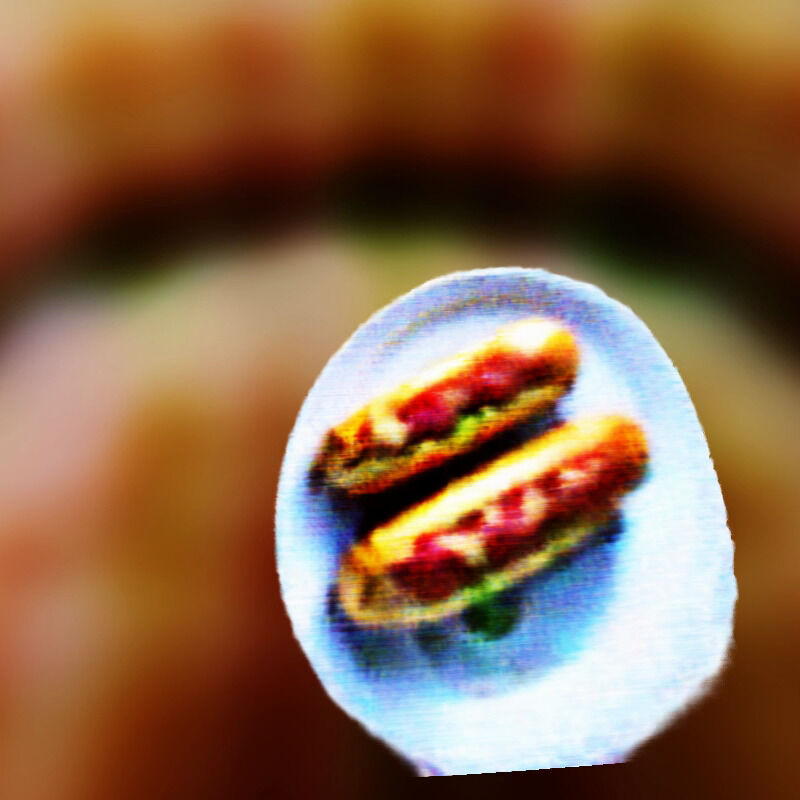}
\end{subfigure}
\\
\hspace*{0.08\textwidth}
\begin{subfigure}{0.08 \textwidth}
\includegraphics[width=\textwidth]{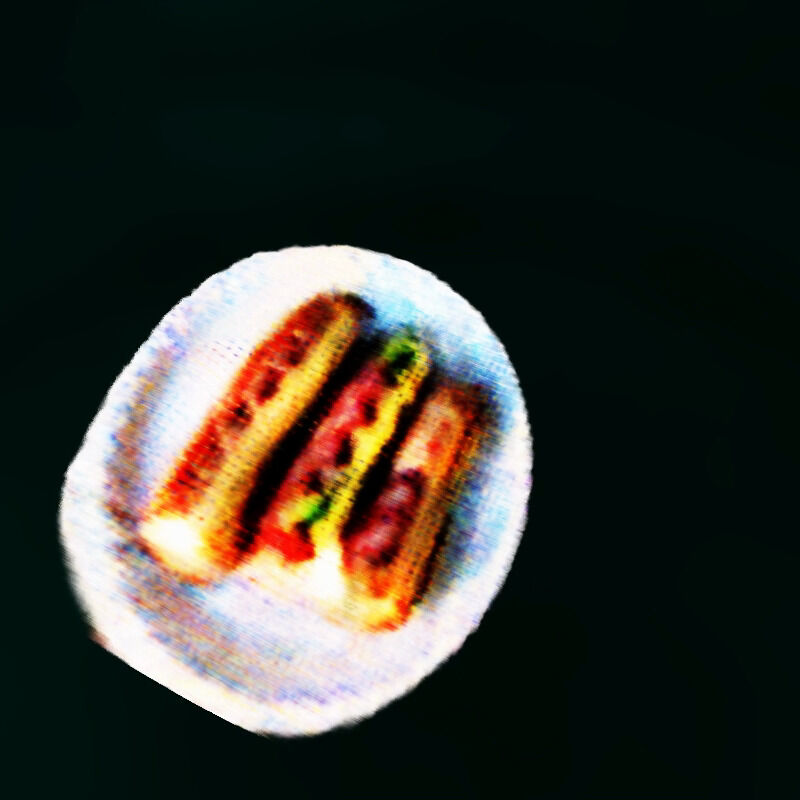}
\end{subfigure}
\begin{subfigure}{0.08 \textwidth}
\includegraphics[width=\textwidth]{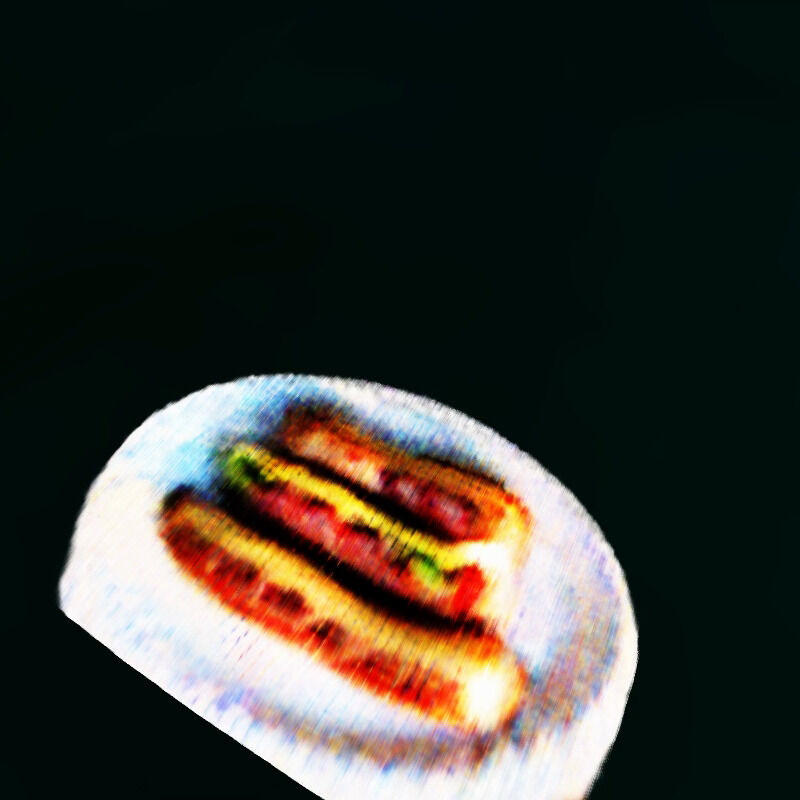}
\end{subfigure}
\begin{subfigure}{0.08 \textwidth}
\includegraphics[width=\textwidth]{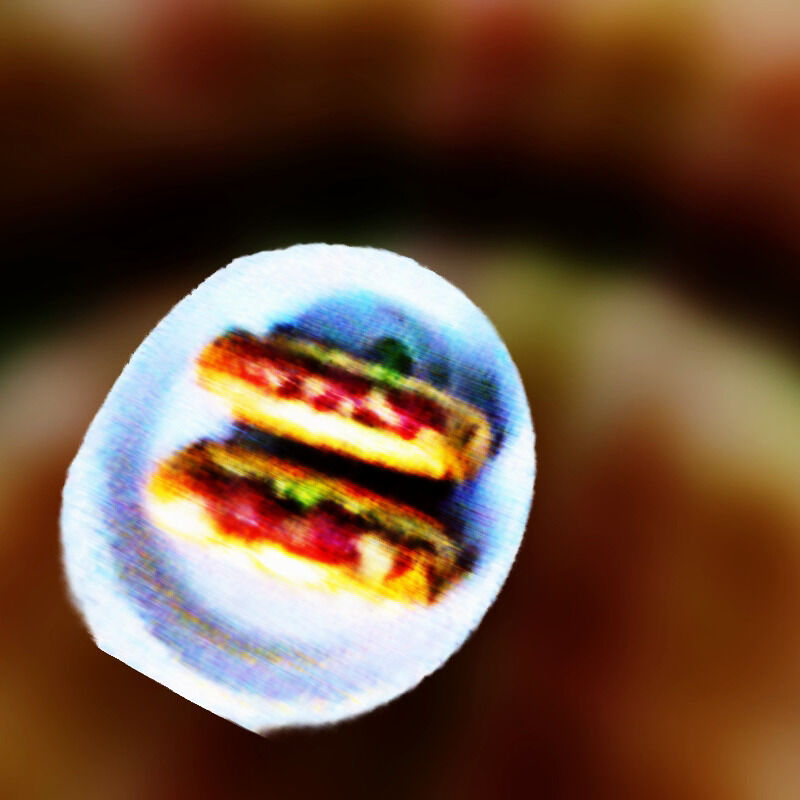}
\end{subfigure}
\begin{subfigure}{0.08 \textwidth}
\includegraphics[width=\textwidth]{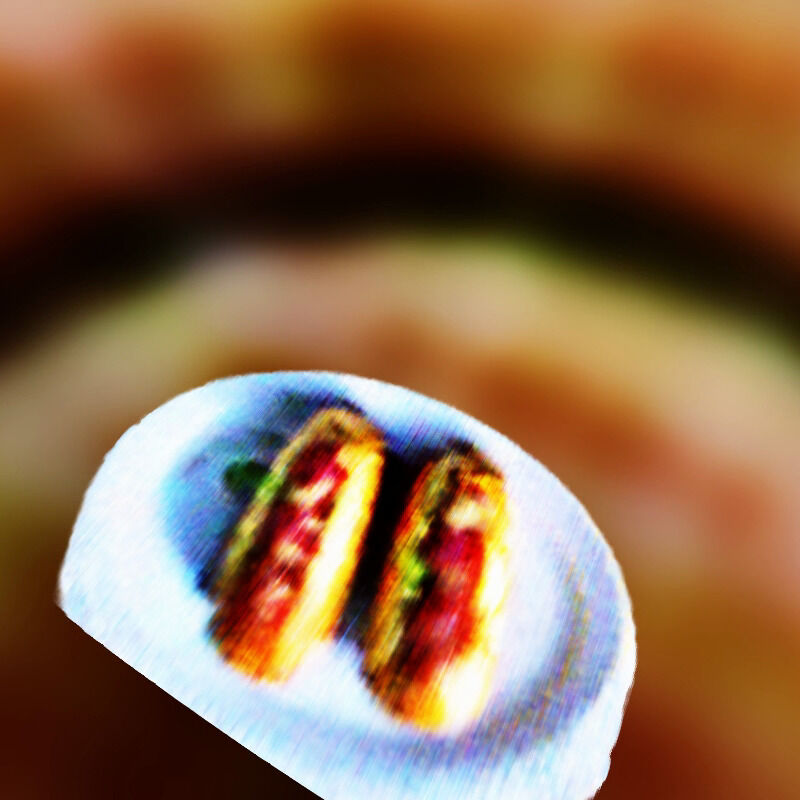}
\end{subfigure}
\begin{subfigure}{0.08 \textwidth}
\includegraphics[width=\textwidth]{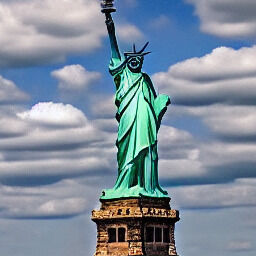}
\end{subfigure}
\begin{subfigure}{0.08 \textwidth}
\includegraphics[width=\textwidth]{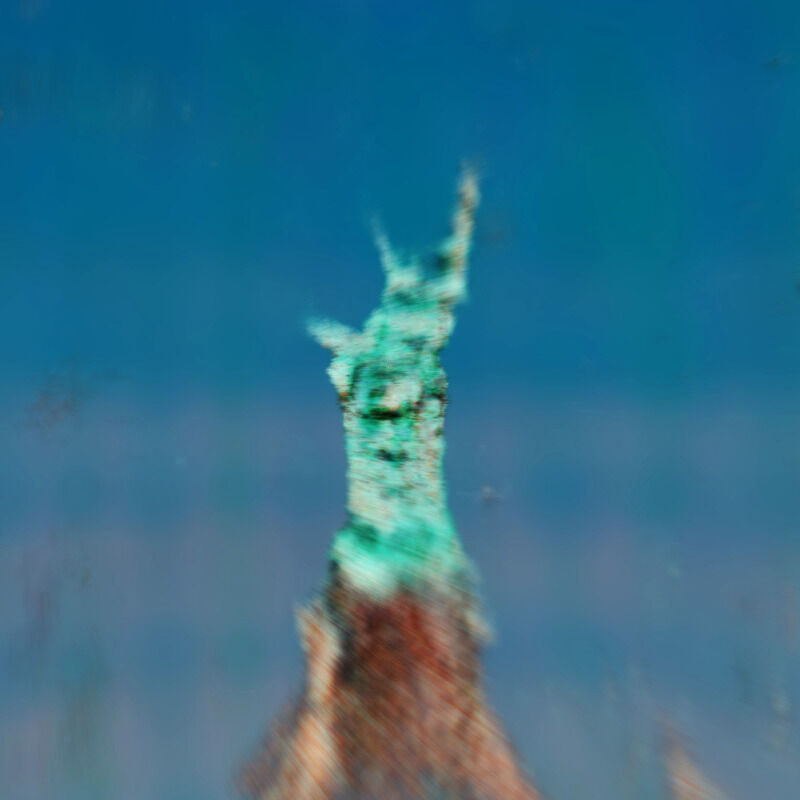}
\end{subfigure}
\begin{subfigure}{0.08 \textwidth}
\includegraphics[width=\textwidth]{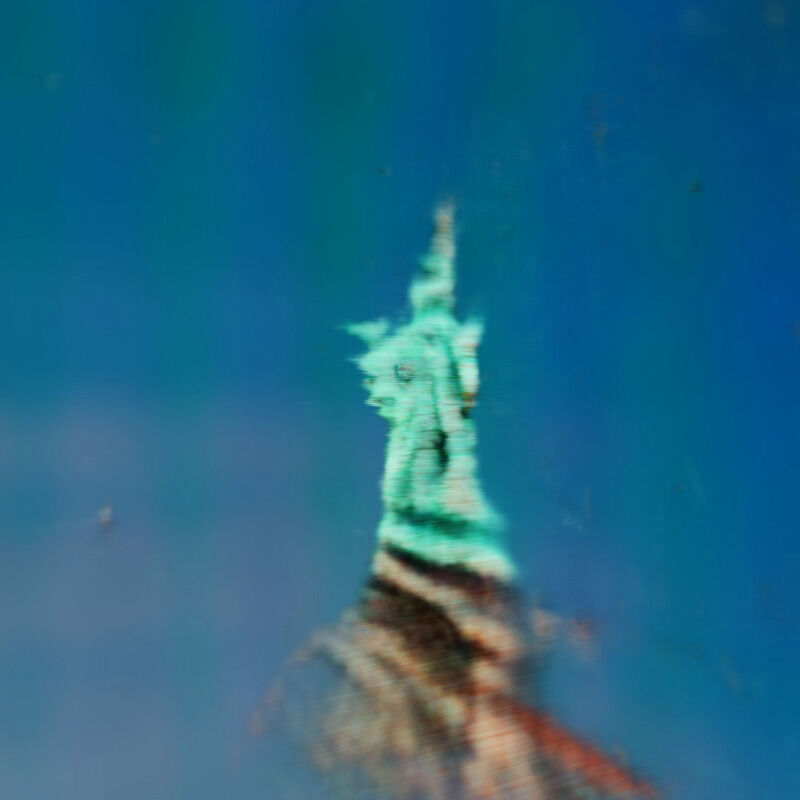}
\end{subfigure}
\begin{subfigure}{0.08 \textwidth}
\includegraphics[width=\textwidth]{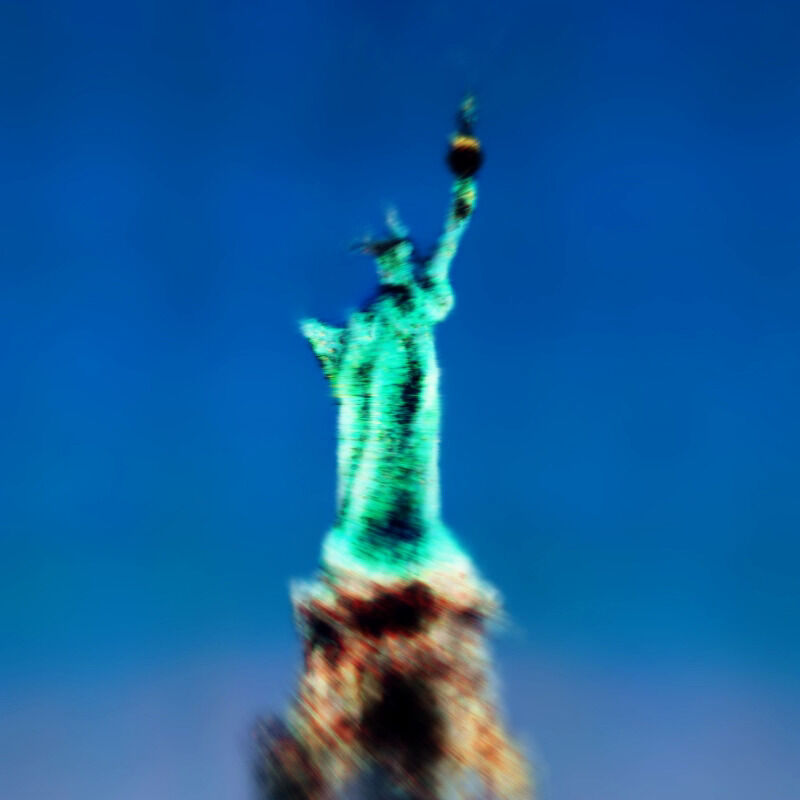}
\end{subfigure}
\begin{subfigure}{0.08 \textwidth}
\includegraphics[width=\textwidth]{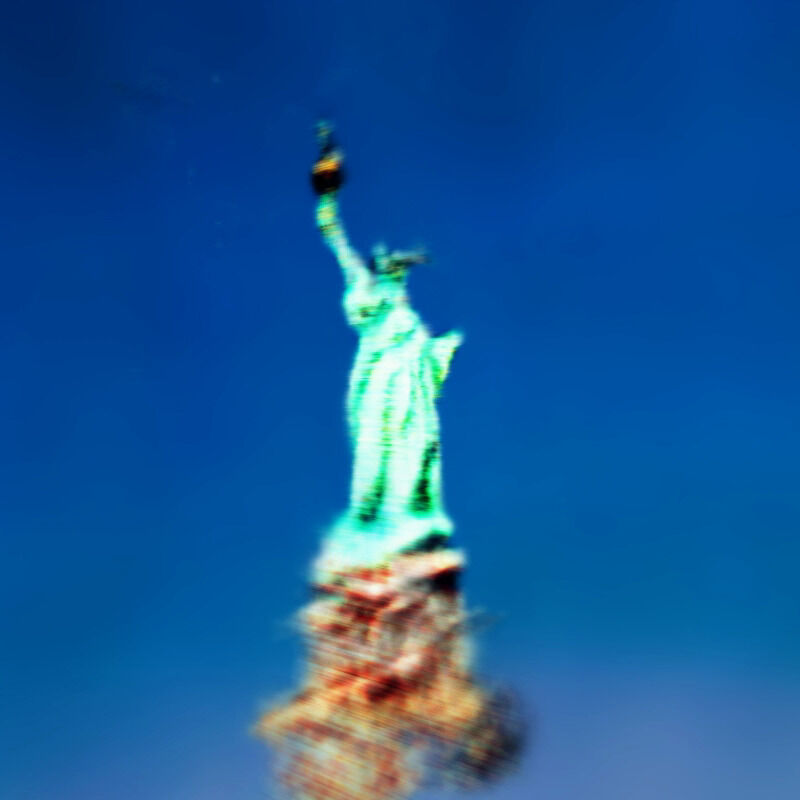}
\end{subfigure}
\\
\hspace*{0.08\textwidth}
\begin{subfigure}{0.08 \textwidth}
\includegraphics[width=\textwidth]{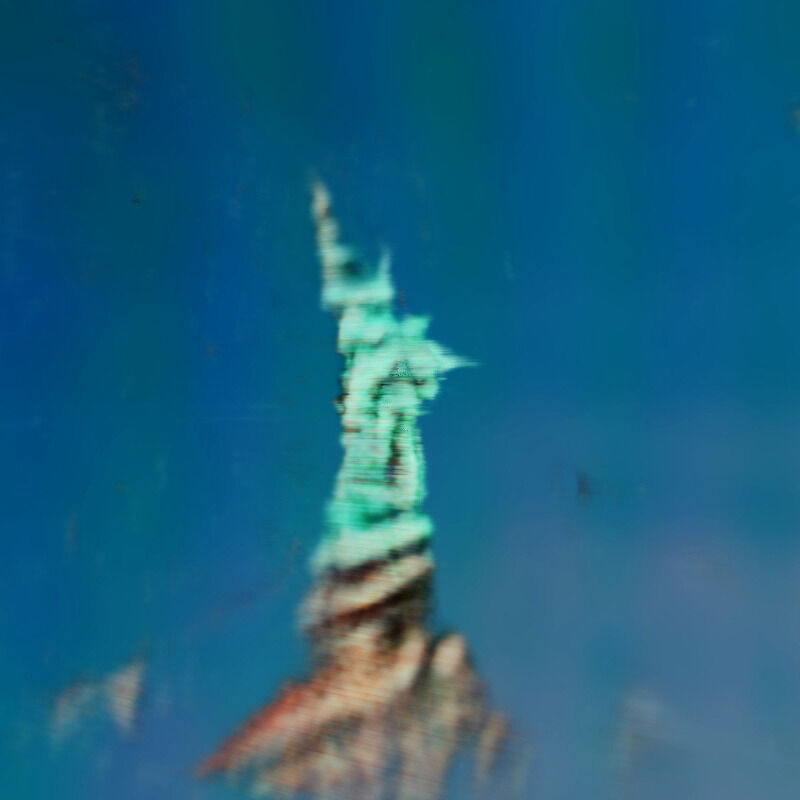}
\end{subfigure}
\begin{subfigure}{0.08 \textwidth}
\includegraphics[width=\textwidth]{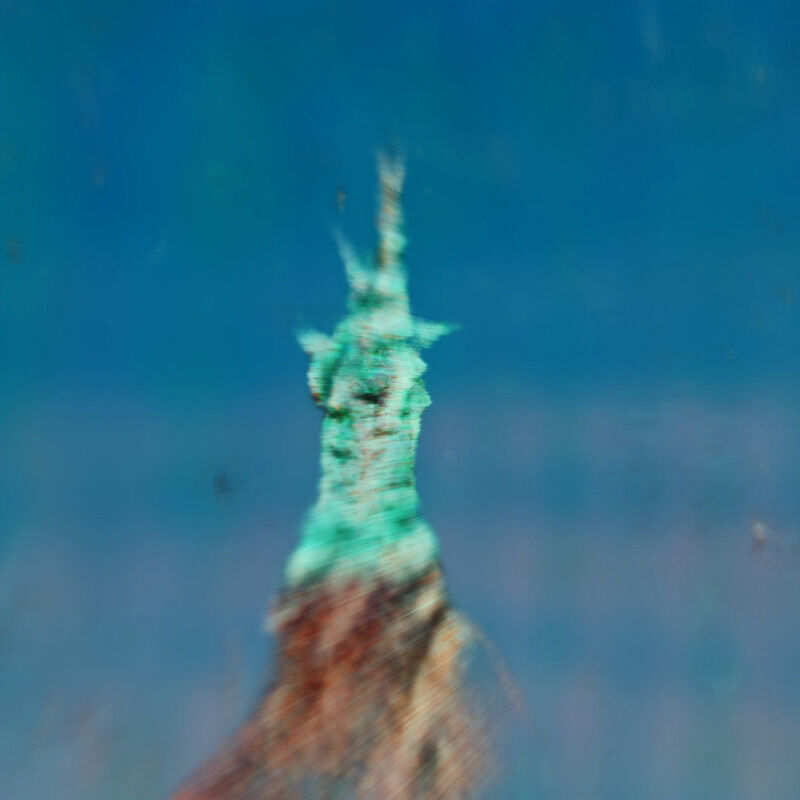}
\end{subfigure}
\begin{subfigure}{0.08 \textwidth}
\includegraphics[width=\textwidth]{images/ablation_clip/liberty/full/out66.jpg}
\end{subfigure}
\begin{subfigure}{0.08 \textwidth}
\includegraphics[width=\textwidth]{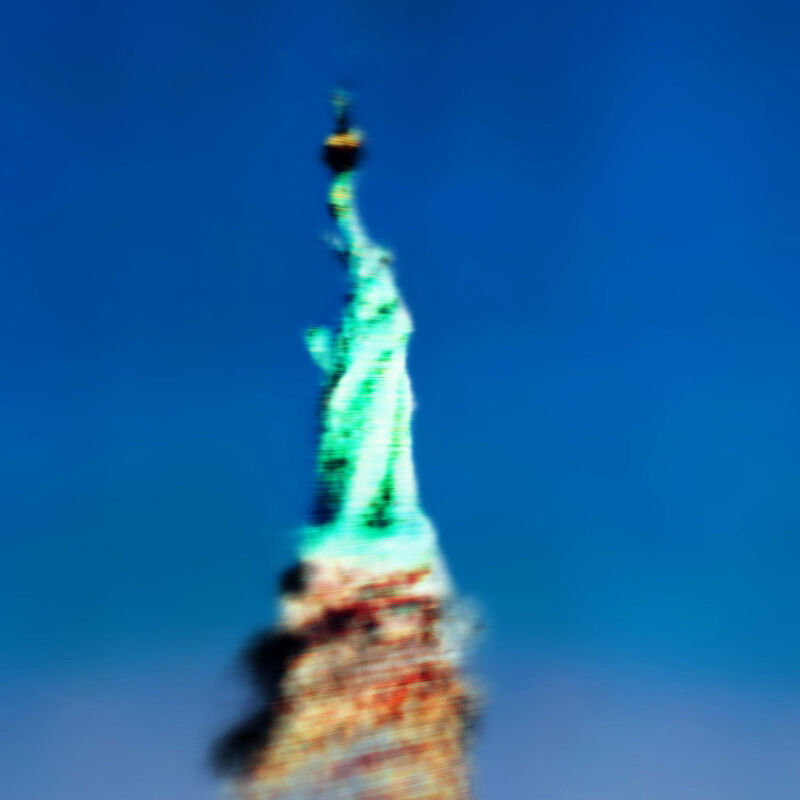}
\end{subfigure}
\caption{Ablation study on CLIP guidance for the diffusion prior.}
\vspace{-7mm}
\label{fig:ablation_clip}
\end{figure}

\subsection{Evaluation Workflow}

\looseness=-1
We conduct extensive experiments using both text-to-image synthetic images and real-world images. 
Synthetic images in our experiments are generated using PNDM Sampler~\cite{liu2022pseudo} with 50 sampling steps. Real images are either from CO3Dv2 dataset~\cite{reizenstein21co3d} or randomly crawled from the Google search engine. 
For each image, we use a pre-trained background removal framework~\cite{bgremoval} to segment the foreground objects. Then we obtain their depth using LeRes~\cite{leres} with the ``Boost Your Own depth''~\cite{Miangoleh2021Boosting} strategy. \cite{Miangoleh2021Boosting} inferences the monocular depth estimation network on multiple patches and then fuses the results for better details.
We compare with previous sparse view NeRFs, including Depth-supervised NeRF~\cite{deng2022depth} and SinNeRF~\cite{xu2022sinnerf}. We also compare with DietNeRF~\cite{jain2021putting} that uses CLIP as guidance.

\subsection{Comparisons}

We present quantitative results in eight scenes.
We report the average CLIP distance between rendered image and the reference image as a metric for how well the object is represented.
During testing, we use a CLIP model different from training time and evaluate on held out viewing directions with a larger distance-to-origin than training. We render 100 images on each scene for each method.
As shown in Tab.~\ref{tab:clip}, our method achieves the best CLIP distance among existing approaches. 

We also provide visual comparisons in Fig.~\ref{fig:results}. Since DS-NeRF and SinNeRF rely on the exact value of the available depth information, their performance using monocular depth estimation is far from satisfactory. It can also be seen that the DietNeRF suffers from unconstrained geometry. Our method, on the other hand, is able to synthesize visually pleasing novel views with 3D consistency.

\subsection{Ablation Study}

\paragraph{Variants of supervision.}
In this section, we perform detailed ablation studies on different components of our framework. ``w/o $\mathcal{L}_\text{ranking}$'' replaces the ranking loss with $L_1$ loss on depth. ``w/o $\mathcal{L}_\text{diff}$'' replaces the CLIP guided diffusion prior with CLIP. ``w/o fine-tuning'' refers to using the pre-trained diffusion model. As shown in Fig.~\ref{fig:ablation_sd}, our full model delivers the best performance, while other variants have different issues. ``w/o $\mathcal{L}_\text{ranking}$'' relies on the exact value from depth information and suffers from unreliable depth. ``w/o $\mathcal{L}_\text{diff}$'' uses CLIP as a prior, and the optimized result is blurry and collapsed. ``w/o fine-tuning'' uses a pre-trained diffusion model, since the text prompt is not very accurate, the resulting object has lettuce which doesn't exist in the reference image.

\vspace{-3mm}
\paragraph{Effectiveness of CLIP guided diffusion prior}
We validate the effectiveness of CLIP guidance in our diffusion prior design in Fig.~\ref{fig:ablation_clip}. It can be seen that without the CLIP guidance, although the model produces a 3D object, it doesn't look like the reference image and has a strange shape. In comparison, our full model generates a better shape and looks more similar to the reference image.

\section{Conclusions and Future Work}

We present NeuralLift-360, a novel framework to lift an in-the-wild 2D photo into a 3D object with 360-degree views. NeuralLift-360 learns probabilistic-driven 3D lifting with CLIP-guided diffusion priors and mitigates the depth errors by a scale-invariant depth ranking loss. Comprehensive experiments are conducted on real and synthetic images, where NeuralLift-360 outperforms the current state-of-the-art methods.   
While our current implementation is based on Stable Diffusion~\cite{ldm} version 1.4, the framework should also work well with other diffusion models such as Imagen~\cite{imagen} or DALL-E2~\cite{ramesh2022hierarchical}, once they are open-sourced. 

\noindent\textbf{Limitations.} Despite the encouraging visual results rendered by NeuralLift-360, the target resolution ($128 \times 128$) is still behind large generative models~\cite{imagen,ramesh2022hierarchical}. Moreover, challenging cases (e.g., multiple objects with occlusion) are not included in our assumptions, which need to be further explored. Going forward, we will investigate how to expand our method to more general scenarios.

{\small
\bibliographystyle{ieee_fullname}
\bibliography{egbib}
}

\clearpage

\appendix

\section{Introduction}

In this Supplementary Material, we first provide a detailed derivation of our probabilistic driven diffusion prior, then additional training recipe for NeuralLift-360. Furthermore, we present more implementation details and experimental comparisons.

\section{Detailed Derivation.}

In this section, we provide detailed mathematical derivation towards our objective loss. We repeat our notations and settings as below.

\paragraph{Notations.}
Given an image $\Mat{y} \in \real^N$ and its text description $\Mat{z} \in \real^D$, NeuralLift-360 intends to reconstruct a 3D scene $\Mat{V} \in \real^M$, where $\Mat{V}$ is a parameterization of 3D scene that can be be either a radiance volume or the implicit neural representation.
We regard $\Mat{y}$, $\Mat{z}$ and $\Mat{V}$ as random variables.
Further on, we define $h(\Mat{V}, \Mat{\Phi})$ as the rendering function that displays $\Mat{V}$ with respect to the camera pose $\Mat{\Phi}$.

\subsection{Pose Conditioned Latent Model}

To reconstruct $\Mat{V}$ from $\Mat{y}$ and $\Mat{z}$, we maximize a log-posterior and apply Bayesian rule:
\begin{align} \label{eqn:apx_max_posterior}
\log p(\Mat{V} | \Mat{y}, \Mat{z}) = \underbrace{\log p(\Mat{y} | \Mat{V}, \Mat{z})}_{\text{likelihood}} + \underbrace{\log p(\Mat{V} | \Mat{z})}_{\text{prior}} + const.,
\end{align}
where $const.$ denotes the evidence term, a constant.
Next step is to introduce camera pose $\Mat{\Phi} \in \mathbb{SO}(3) \times \real^3$ as a latent variable, then the likelihood term can be rewritten as:
\begin{align}
p(\Mat{y} | \Mat{V}, \Mat{z}) &= \int p(\Mat{y} | \Mat{V}, \Mat{\Phi}, \Mat{z}) p(\Mat{\Phi} | \Mat{V}, \Mat{z}) d\Mat{\Phi} \\
\label{eqn:cam_likelihood} & = \int p(\Mat{y} | \Mat{V}, \Mat{\Phi}, \Mat{z}) \frac{p(\Mat{V} | \Mat{\Phi}, \Mat{z}) p(\Mat{\Phi})}{p(\Mat{V} | \Mat{z})} d\Mat{\Phi} \\
&= \frac{1}{p(\Mat{V} | \Mat{z})} \mean_{\Mat{\Phi}} \left[ p(\Mat{y} | \Mat{V}, \Mat{\Phi}, \Mat{z}) p(\Mat{V} | \Mat{\Phi}, \Mat{z}) 
\right].
\end{align}
where we assume $\Mat{\Phi}$ is independent of $\Mat{z}$.
We proceed by substituting Eq. \ref{eqn:cam_likelihood} into Eq. \ref{eqn:apx_max_posterior}:
\begin{align}
\log p(\Mat{V} | \Mat{y}, \Mat{z}) &= \log \frac{\mean_{\Mat{\Phi}} \left[ p(\Mat{y} | \Mat{V}, \Mat{\Phi}, \Mat{z}) p(\Mat{V} | \Mat{\Phi}, \Mat{z})\right]}{p(\Mat{V} | \Mat{z})} \nonumber \\ & \quad + \log p(\Mat{V} | \Mat{z}) + const. \\
&= \log \mean_{\Mat{\Phi}} \left[ p(\Mat{y} | \Mat{V}, \Mat{\Phi}, \Mat{z}) p(\Mat{V} | \Mat{\Phi}, \Mat{z}) \right] + const.
\end{align}
After we apply Jensen inequality to obtain an evidence lower evidence of Eq. \ref{eqn:apx_max_posterior}:
\begin{align}
&\log \mean_{\Mat{\Phi}} \left[ p(\Mat{y} | \Mat{V}, \Mat{\Phi}, \Mat{z}) p(\Mat{V} | \Mat{\Phi}, \Mat{z}) \right] + const. \nonumber \\
&\ge \mean_{\Mat{\Phi}} \left[ \log p(\Mat{y} | \Mat{V}, \Mat{\Phi}, \Mat{z}) + \log p(\Mat{V} | \Mat{\Phi}, \Mat{z}) \right] + const.
\end{align}
Afterwards, we use rendering function $h(\Mat{V}, \Mat{\Phi})$ to bridge the 3D likelihood estimation to the image domain. Specifically, we let $p(\Mat{y} | \Mat{V}, \Mat{\Phi}, \Mat{z}) = p(\Mat{y} | h(\Mat{V}, \Mat{\Phi}), \Mat{z})$ and $p(\Mat{V} | \Mat{\Phi}, \Mat{z}) = p(h(\Mat{V}, \Mat{\Phi}) | \Mat{z})$. Then we derive a general training objective (omitting the constant):
\begin{align} \label{eqn:apx_prob_objective}
\mathcal{L} &= -\mean_{\Mat{\Phi}} \left[ \underbrace{\log p(\Mat{y} | h(\Mat{V}, \Mat{\Phi}), \Mat{z})}_{\text{referenced loss}} + \underbrace{\log p(h(\Mat{V}, \Mat{\Phi}) | \Mat{z})}_{\text{non-referenced loss}} \right] \\
& = -\mean_{\Mat{\Phi}} \left[ \log p(h(\Mat{V}, \Mat{\Phi}) | \Mat{y}, \Mat{z}) \right] + const.,
\end{align}
where the constant equals to $-\log p(\Mat{y} | \Mat{z})$.

\subsection{Diffusion Model as Evidence Lower Bound}

Our next step is to surrogate the probability densities with the score matching loss, which can leverage pre-trained generative prior for view regularization.
Our derivation mainly follows from \cite{ho2020denoising, luo2022understanding}.
Below we summarize key steps to attain the ELBO, and more details can be found in Chap. 3 of \cite{luo2022understanding}:
\begin{align}
\nonumber & \log p(\Mat{x} | \Mat{y}, \Mat{z}) \ge \mean \left[ \log \frac{p(\Mat{x}_1, \cdots, \Mat{x}_T | \Mat{y}, \Mat{z})}{q(\Mat{x}_1, \cdots, \Mat{x}_T | \Mat{x}_0)} \right]\\
&= \sum_{t=2}^{T} \mean_{q(\Mat{x}_t |  \Mat{x}_0)} \left[ \KL(q(\Mat{x}_{t-1} | \Mat{x}_t, \Mat{x}_0) \Vert p_{\theta}(\Mat{x}_{t-1} | \Mat{x}_t, \Mat{y}, \Mat{z}) ) \right] \nonumber \\ & \quad + \mean_{q(\Mat{x}_1 |  \Mat{x}_0)} \left[ \log p_{\theta} (\Mat{x}_0 | \Mat{x}_1) \right] + const.
\end{align}
Consider a diffusion process with marginal distribution: $q(\Mat{x}_{t} | \Mat{x}_0) = \mathcal{N}(\alpha_t \Mat{x}_0, \sigma_t^2 \Mat{I})$.
As shown by \cite{ho2020denoising, song2020denoising}, the KL divergence term can be simplified as:
\begin{align}
&\KL(q(\Mat{x}_{t-1} | \Mat{x}_t, \Mat{x}_0) \Vert p_{\theta}(\Mat{x}_{t-1} | \Mat{x}_t, \Mat{y}, \Mat{z}) ) \\ &=  w(t) \mean_{\Mat{\epsilon} \sim \mathcal{N}(\Mat{0}, \Mat{I})}  \left[ \left\lVert \Mat{\epsilon}_{\theta}(\alpha_t \Mat{x}_0 + \sigma_t \Mat{\epsilon} | \Mat{y}, \Mat{z}) - \Mat{\epsilon} \right\rVert_2^2 \right],
\end{align}
and similarly, we have:
\begin{align}
& \mean_{q(\Mat{x}_1 |  \Mat{x}_0)} \left[ \log p_{\theta} (\Mat{x}_0 | \Mat{x}_1) \right] \\ &=  w(1) \mean_{\Mat{\epsilon} \sim \mathcal{N}(\Mat{0}, \Mat{I})}  \left[ \left\lVert \Mat{\epsilon}_{\theta}(\alpha_1 \Mat{x}_0 + \sigma_1 \Mat{\epsilon} | \Mat{y}, \Mat{z}) - \Mat{\epsilon} \right\rVert_2^2 \right],
\end{align}
where $w(t)$ is a time dependent coefficient, $\Mat{\epsilon}_{\theta}(\Mat{x} | \Mat{y}, \Mat{z}) = \nabla \log p_{\theta}(\Mat{x} | \Mat{y}, \Mat{z})$ is known as the score function of the approximated image distribution.
Altogether, we can utilize the following objective in the place of Eq. \ref{eqn:apx_prob_objective} (omitting constants):
\begin{align} \label{eqn:apx_elbo_objective}
&\mathcal{L}_{diff} = \nonumber \\ & -\sum_{t=1}^{T} w(t) \mean_{\Mat{\Phi}, \Mat{\epsilon}}  \left[ \left\lVert \Mat{\epsilon}_{\theta}(\alpha_t h(\Mat{V}, \Mat{\Phi}) + \sigma_t \Mat{\epsilon} | \Mat{y}, \Mat{z}) - \Mat{\epsilon} \right\rVert_2^2 \right].
\end{align}
Finally, we introduce how to leverage classifier guidance idea \cite{lecun2006tutorial} to compute $\Mat{\epsilon}_{\theta}(\Mat{x} | \Mat{y}, \Mat{z})$. Define some off-the-shelf text-to-image diffusion model as $\Mat{\epsilon}_{\theta}(\Mat{x} | \Mat{z})$.
Note that it does not condition on the reference image $\Mat{y}$.
First, we use simple Bayesian rule to cast $p_{\theta}(\Mat{x} | \Mat{y}, \Mat{z})$ as:
\begin{align}
p_{\theta}(\Mat{x} | \Mat{y}, \Mat{z}) = \frac{p_{\theta}(\Mat{y} | \Mat{x}, \Mat{z}) p_{\theta}(\Mat{x} | \Mat{z})}{p_{\theta}(\Mat{y} | \Mat{z})}.
\end{align}
Then the score function can be written as:
\begin{align}
&\Mat{\epsilon}_{\theta}(\Mat{x} | \Mat{y}, \Mat{z}) = \nabla \log \left(\frac{p_{\theta}(\Mat{y} | \Mat{x}, \Mat{z}) p_{\theta}(\Mat{x} | \Mat{z})}{p_{\theta}(\Mat{y} | \Mat{z})} \right) \\
\label{eqn:class_guide_score} & = \nabla  \log p_{\theta}(\Mat{y} | \Mat{x}, \Mat{z}) + \Mat{\epsilon}_{\theta}(\Mat{x} | \Mat{z}).
\end{align}
One can further use classifier-free guidance to further improve Eq. \ref{eqn:class_guide_score} as:
\begin{align}
&\Mat{\epsilon}_{\theta}(\Mat{x} | \Mat{y}, \Mat{z}) = \nabla  \log p_{\theta}(\Mat{y} | \Mat{x}, \Mat{z}) + (1+\omega) \Mat{\epsilon}_{\theta}(\Mat{x} | \Mat{z}) - \omega \Mat{\epsilon}_{\theta}(\Mat{x}),
\end{align}
where $\omega$ is the guidance strength of the classifier.

\subsection{Computational Specifications}

After going through all the probabilistic derivations, we specify several implementations leading towards our final loss function:

\paragraph{CLIP Similarity Guidance.}
We choose to use CLIP as our reference image guidance. Specifically, we measure the discrimination with a distance metric on the feature space:
\begin{align}
p_{\theta}(\Mat{y} | \Mat{x}_t, \Mat{z}) \propto \exp\left[-\phi \left(\frac{\Mat{x}_t - \sigma_t \Mat{\epsilon}_{\theta}(\Mat{x}_t | \Mat{z})}{\alpha_t}, \Mat{y}\right)\right],
\end{align}
where we choose the inner product on the CLIP embedding space \cite{radford2021learning} as the similarity metric \footnote{Rigorously speaking, inner product is not a metric. However, it is widely used as a similarity score.}, i.e., $\phi(\Mat{x}, \Mat{y}) =  \langle \Mat{x}, \Mat{y} \rangle$. Then we can derive the guidance penalty term we use in our implementation:
\begin{align}
\log p_{\theta}(\Mat{y} | \Mat{x}_t, \Mat{z}) \propto -\left\langle F\left(\frac{\Mat{x}_t - \sigma_t \Mat{\epsilon}_{\theta}(\Mat{x}_t | \Mat{z})}{\alpha_t}\right), F(\Mat{y}) \right\rangle,
\end{align}
where $F(\cdot)$ is a CLIP image encoder.

\paragraph{Camera Sampling Strategy.}
Camera poses $\Mat{\Phi}$ can be in general uniformly sampled over $\mathbb{SO}(3) \times \real^3$ (with reasonably bounded sub-region).
Since our reconstruction tasks are object centric, we first sample the camera positions over a unit sphere and control the camera orientation to look at the origin point. 
What's more, we can leverage a more strategic scheme to fully utilize the reference view.
To be more concrete, we consider a Bernounii random variable $B \sim \operatorname{Bern}(\lambda)$ (as if a switch). When the variable is turned on (i.e., $B = 1$), we utilize the aforementioned sampling algorithm on the unit sphere, and apply $\mathcal{L}_{diff}$ as our loss function.
When it is turned off, we only select the fixed camera pose $\Mat{\Phi}_0$ associated with our reference image (i.e., a Dirac delta distribution centered at $\Mat{\Phi}_0$), and only maximize a likelihood between the synthesized image and the reference image:
\begin{align} \label{eqn:photometric_loss}
p_{\theta}(h(\Mat{V}, \Mat{\Phi}) | \Mat{y}, \Mat{z}, B=0) = \mathcal{N}(h(\Mat{V}, \Mat{\Phi}) | \Mat{y}, \sigma^2 \Mat{I}).
\end{align}
This leads to the definition of our final objective:
\begin{align}
&\mathcal{L}_{total} = -\mean_{\Mat{\Phi}} \left[ \log p(h(\Mat{V}, \Mat{\Phi}) | \Mat{y}, \Mat{z}) \right] \\
&= -p(B=1) \mean_{\Mat{\Phi} | B=1} \left[ \log p(h(\Mat{V}, \Mat{\Phi}) | \Mat{y}, \Mat{z}, B=1) \right] \nonumber \\ & \quad - p(B=0) \log p(h(\Mat{V}, \Mat{\Phi}_0) | \Mat{y}, \Mat{z}, B=0) \\
&\le \lambda \mathcal{L}_{diff} + (1-\lambda)/\sigma^2 \left\lVert h(\Mat{V}, \Mat{\Phi}_0) - \Mat{y} \right\rVert_2^2.
\end{align}

\paragraph{Fine-tuning Diffusion Model.}
We also interpret our domain adaption finetuning as a step to condition the diffusion model $\Mat{\epsilon}(\Mat{x} | \Mat{z})$ on the reference information  $\Mat{y}$:
\begin{align}
\Mat{\epsilon}_{\theta}(\Mat{x} | \Mat{z}) \xrightarrow[]{\mathcal{L}_{\text{finetune}} \text{ w/ } \Mat{y}} \Mat{\epsilon}_{\theta^*}(\Mat{x} | \Mat{y}, \Mat{z}).
\end{align}

\paragraph{Pseudo-Depth Supervision.}
The pseudo-depth supervision on the reference view can be viewed as adding a regularization term onto Eq. \ref{eqn:photometric_loss}:
\begin{align}
& p_{\theta}(h(\Mat{V}, \Mat{\Phi}) | \Mat{y}, \Mat{z}, B=0) = \\ & \quad \mathcal{N}(h_{rgb}(\Mat{V}, \Mat{\Phi}) | \Mat{y}, \sigma^2 \Mat{I}) \nonumber \cdot p(h_{depth}(\Mat{V}, \Mat{\Phi}) | \Mat{y}),
\end{align}
where we extend the renderer $h(\Mat{V}, \Mat{\Phi})$ to independently render RGB image and depth map, and we further define $p(\Mat{d} | \Mat{y}) \propto \exp(-\mathcal{L}_{\text{ranking}}(\Mat{d}, G(\Mat{y})))$, $G(\cdot)$ is a monocular depth estimator.

\section{Additional Training Recipe}
\label{sec:regularizer}

\paragraph{Lighting Augmentation with Shading}

NeRF is prone to poor geometry, so we incorporate the supervision of surface normal from RefNeRF~
\cite{verbin2022ref} to improve the geometry quality. Similar to DreamFusion~\cite{dreamfusion} and RefNeRF~\cite{verbin2022ref}, we replace NeRF’s parameterization of
view-dependent outgoing radiance with the surface itself. Compared with traditional NeRF that emits radiance conditioning on the viewing direction, our implementation expresses the geometry itself and allows additional shading.

The surface normal vector is defined as the negative gradient of the volume density with respect to the 3D location~\cite{verbin2022ref,boss2021nerd,srinivasan2021nerv},
\begin{equation}
\hat{\mathbf{n}}(\mathbf{x})=-\frac{\nabla \sigma(\mathbf{x})}{\|\nabla \sigma(\mathbf{x})\|}
\end{equation}
As a result, we render the color along the ray during diffuse reflectance~\cite{ramamoorthi2001signal} as follows:
\begin{equation}
\mathbf{c}=\boldsymbol{\rho} \circ\left(\boldsymbol{\ell}_\rho \circ \max (0, \boldsymbol{n} \cdot(\boldsymbol{\ell}-\boldsymbol{\mu}) /\|\boldsymbol{\ell}-\boldsymbol{\mu}\|)+\boldsymbol{\ell}_a\right),
\end{equation}
where $l$ is the point light location, $l_p$ is the color of the point light, and $l_a$ is the color of the ambient light. We generate the point light location by randomly sampling an offset from the camera location $l \sim \mathcal{N}(l_\text{cam}, \mathbf{I})$.

\begin{figure*}
    \centering
   \begin{tabular}{P{0.12\textwidth}P{0.14\textwidth}P{0.2\textwidth}P{0.2\textwidth}P{0.18\textwidth}}
    \scriptsize Reference & \scriptsize DSNeRF~\cite{deng2022depth} & \scriptsize DietNeRF~\cite{jain2021putting} & \scriptsize SinNeRF~\cite{xu2022sinnerf} & \scriptsize \textbf{NeuralLift-360}\\
    \end{tabular}
\begin{subfigure}{0.1 \textwidth}
\includegraphics[width=\textwidth]{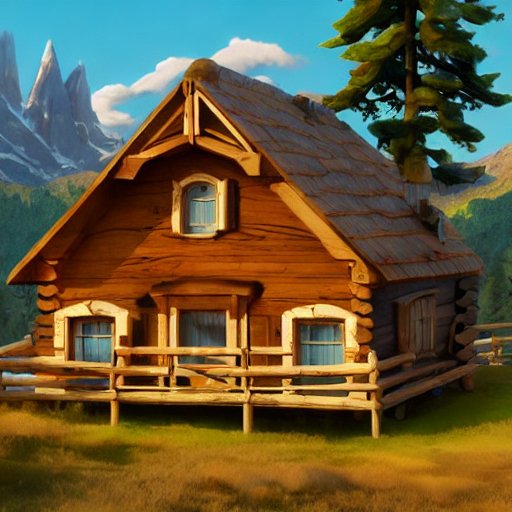}
\end{subfigure}
\begin{subfigure}{0.1 \textwidth}
\includegraphics[width=\textwidth]{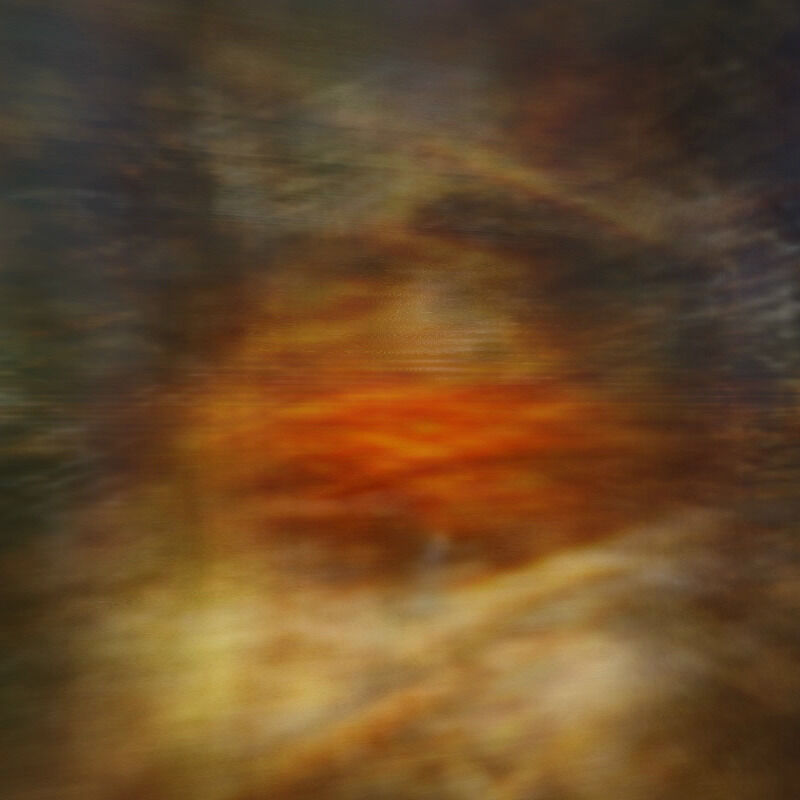}
\end{subfigure}
\begin{subfigure}{0.1 \textwidth}
\includegraphics[width=\textwidth]{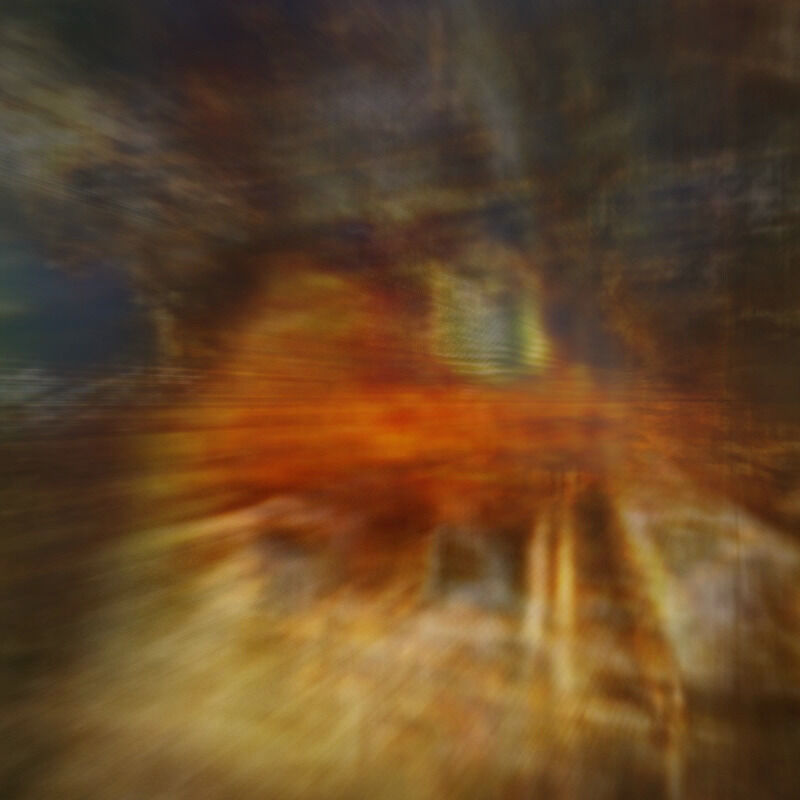}
\end{subfigure}
\begin{subfigure}{0.1 \textwidth}
\includegraphics[width=\textwidth]{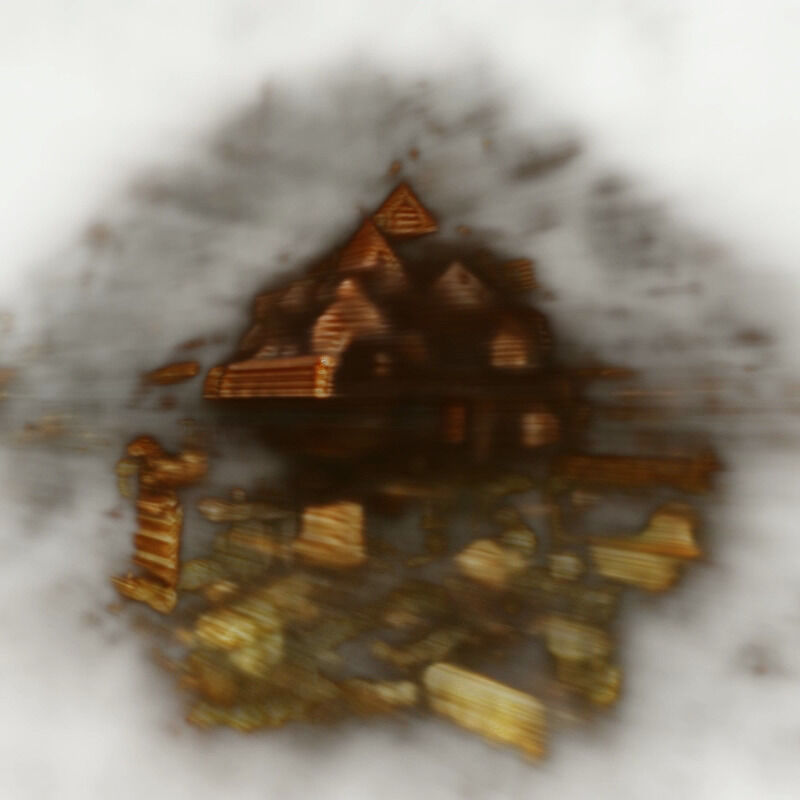}
\end{subfigure}
\begin{subfigure}{0.1 \textwidth}
\includegraphics[width=\textwidth]{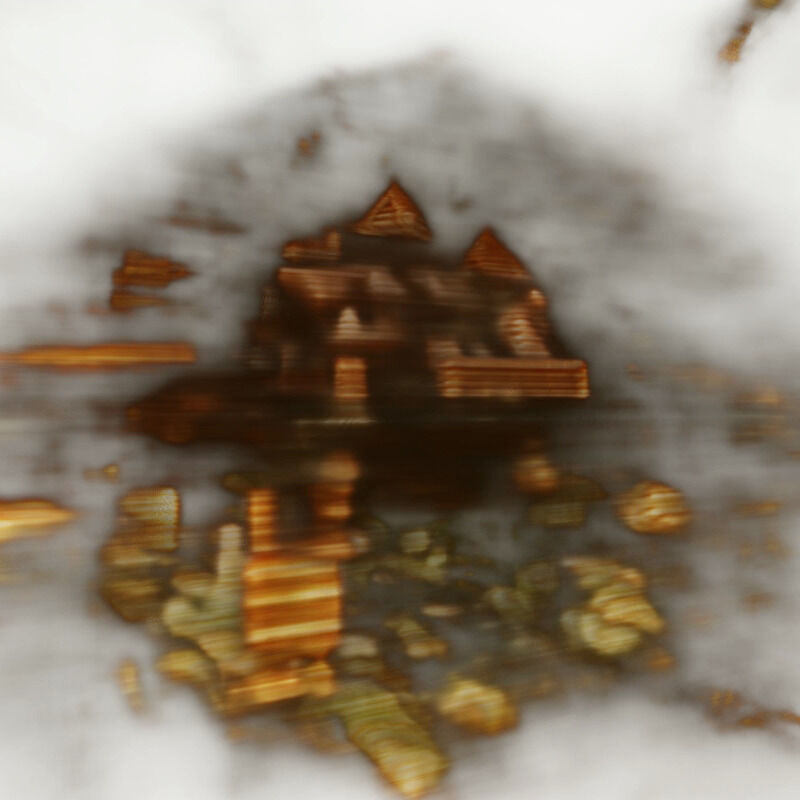}
\end{subfigure}
\begin{subfigure}{0.1 \textwidth}
\includegraphics[width=\textwidth]{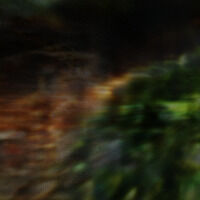}
\end{subfigure}
\begin{subfigure}{0.1 \textwidth}
\includegraphics[width=\textwidth]{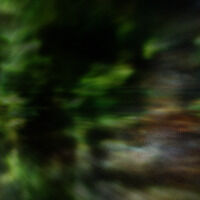}
\end{subfigure}
\begin{subfigure}{0.1 \textwidth}
\includegraphics[width=\textwidth]{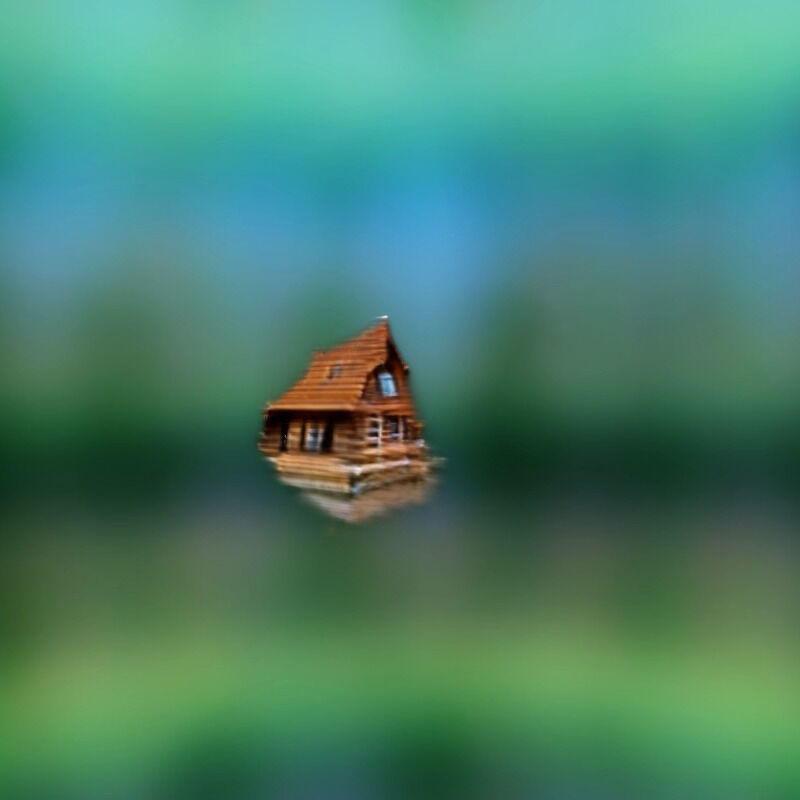}
\end{subfigure}
\begin{subfigure}{0.1 \textwidth}
\includegraphics[width=\textwidth]{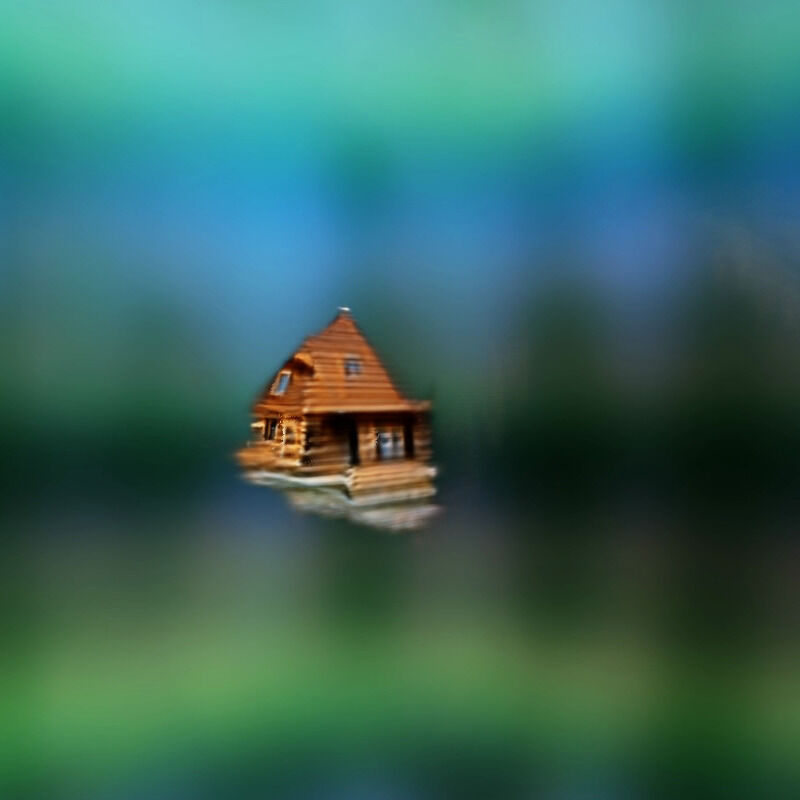}
\end{subfigure}
\\
\hspace*{.1\textwidth}
\begin{subfigure}{0.1 \textwidth}
\includegraphics[width=\textwidth]{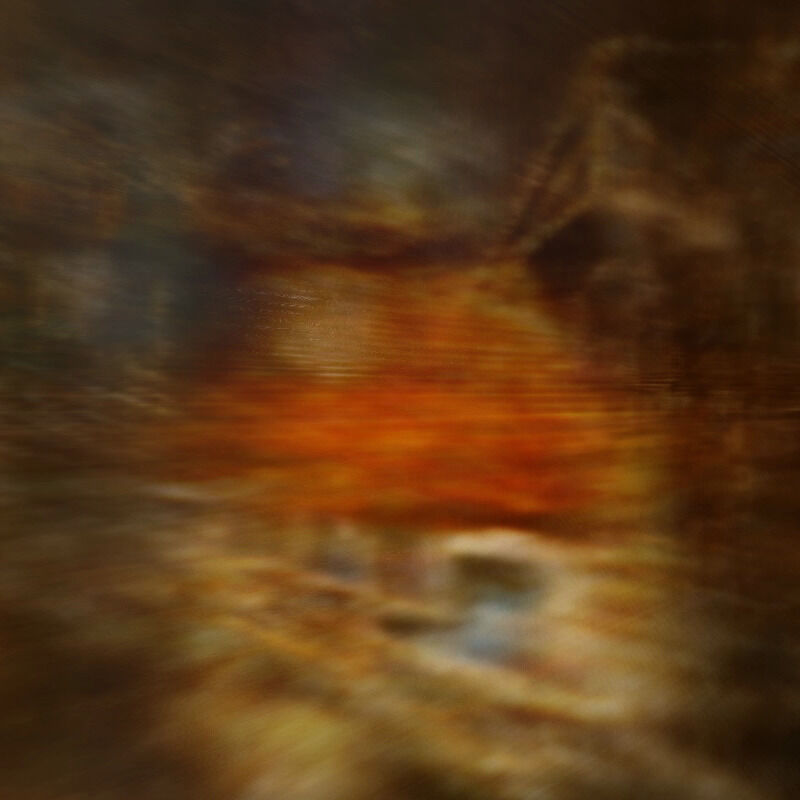}
\end{subfigure}
\begin{subfigure}{0.1 \textwidth}
\includegraphics[width=\textwidth]{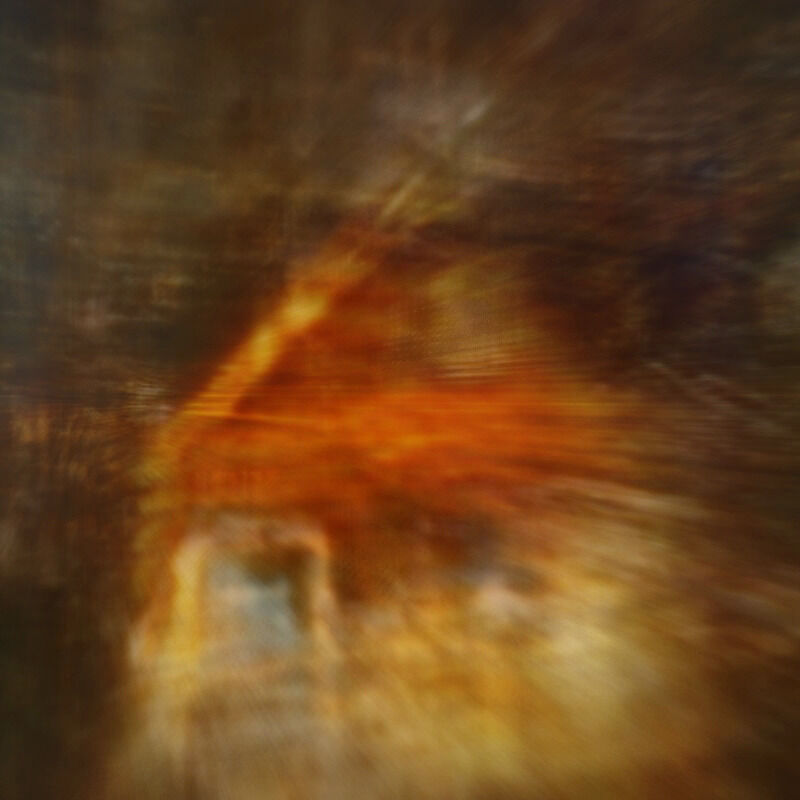}
\end{subfigure}
\begin{subfigure}{0.1 \textwidth}
\includegraphics[width=\textwidth]{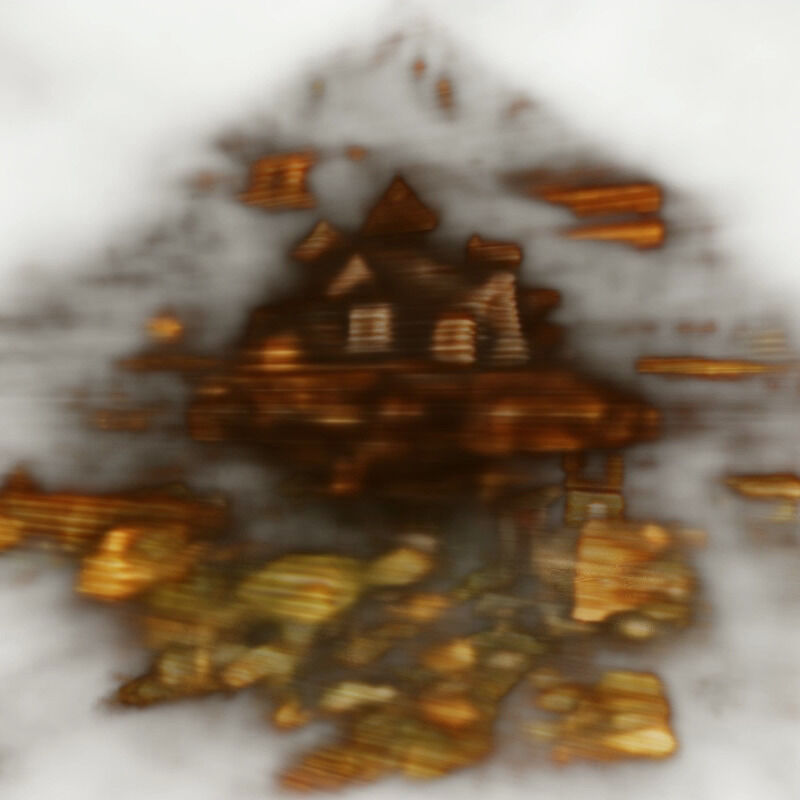}
\end{subfigure}
\begin{subfigure}{0.1 \textwidth}
\includegraphics[width=\textwidth]{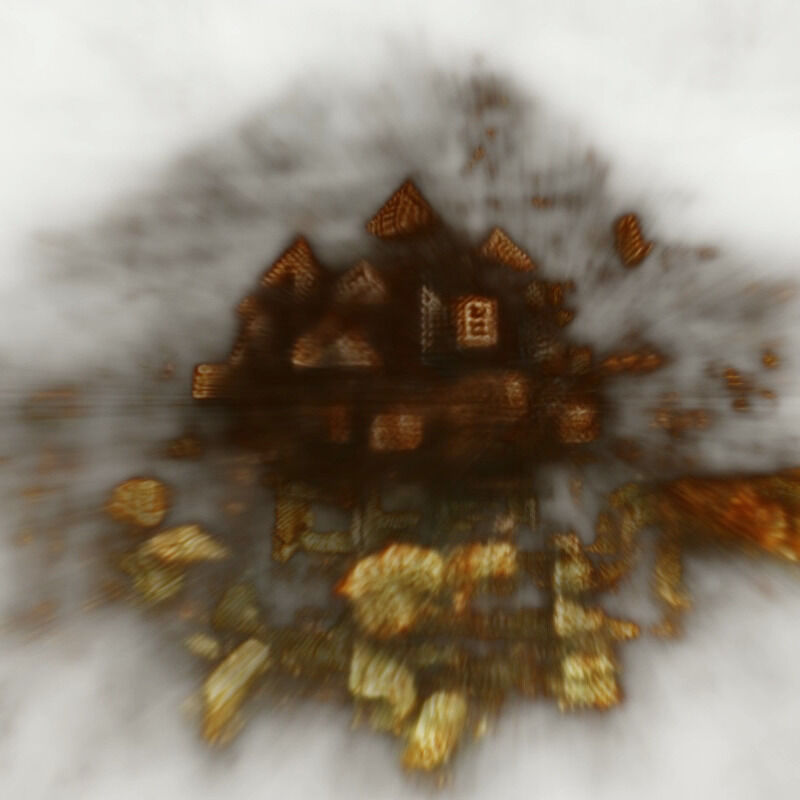}
\end{subfigure}
\begin{subfigure}{0.1 \textwidth}
\includegraphics[width=\textwidth]{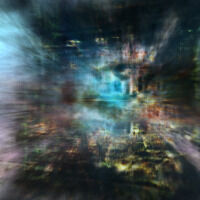}
\end{subfigure}
\begin{subfigure}{0.1 \textwidth}
\includegraphics[width=\textwidth]{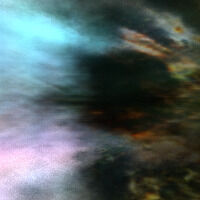}
\end{subfigure}
\begin{subfigure}{0.1 \textwidth}
\includegraphics[width=\textwidth]{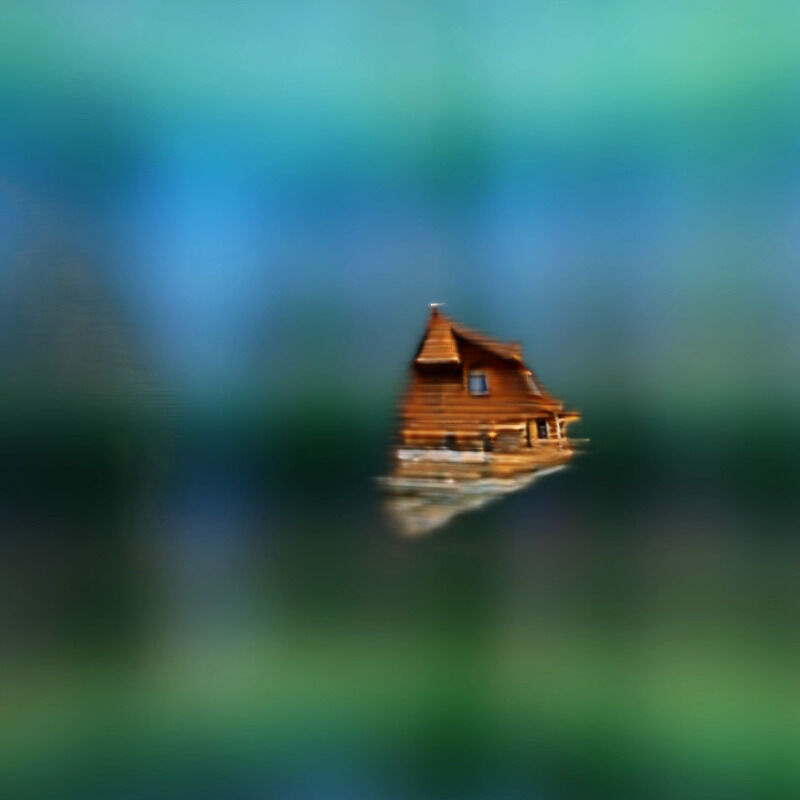}
\end{subfigure}
\begin{subfigure}{0.1 \textwidth}
\includegraphics[width=\textwidth]{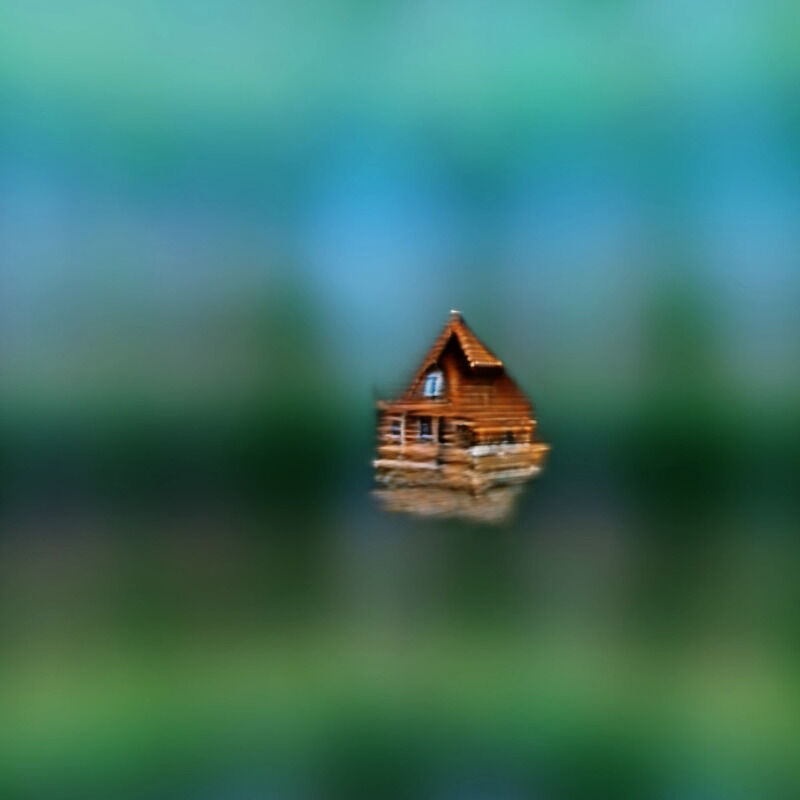}
\end{subfigure}
\begin{subfigure}{0.1 \textwidth}
\includegraphics[width=\textwidth]{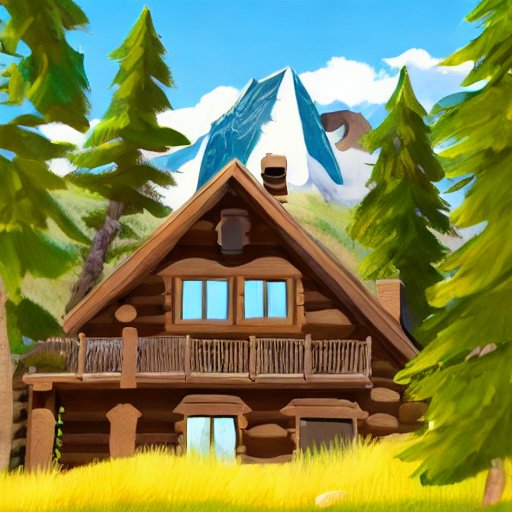}
\end{subfigure}
\begin{subfigure}{0.1 \textwidth}
\includegraphics[width=\textwidth]{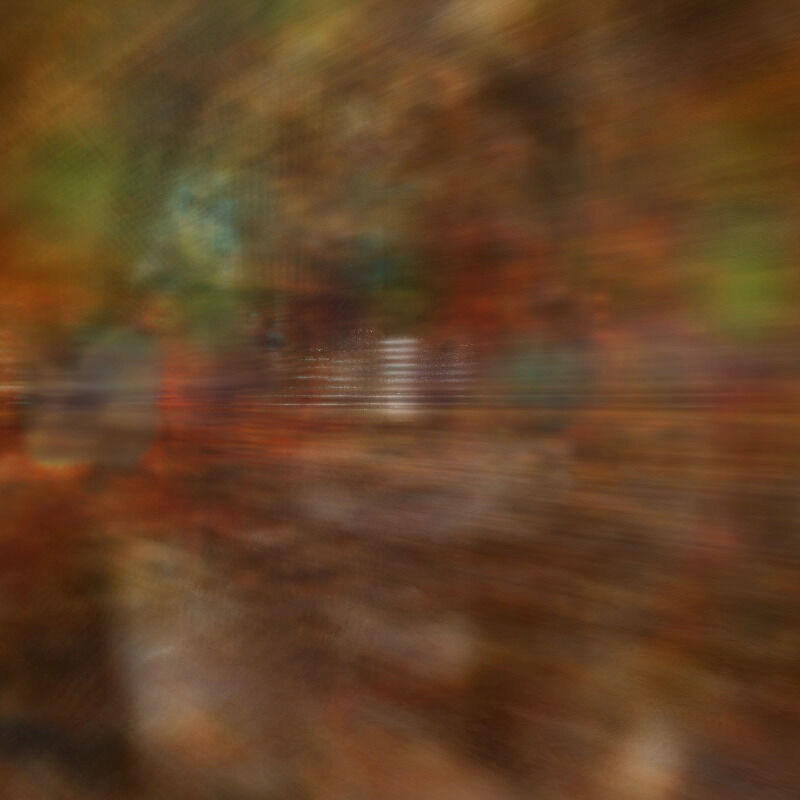}
\end{subfigure}
\begin{subfigure}{0.1 \textwidth}
\includegraphics[width=\textwidth]{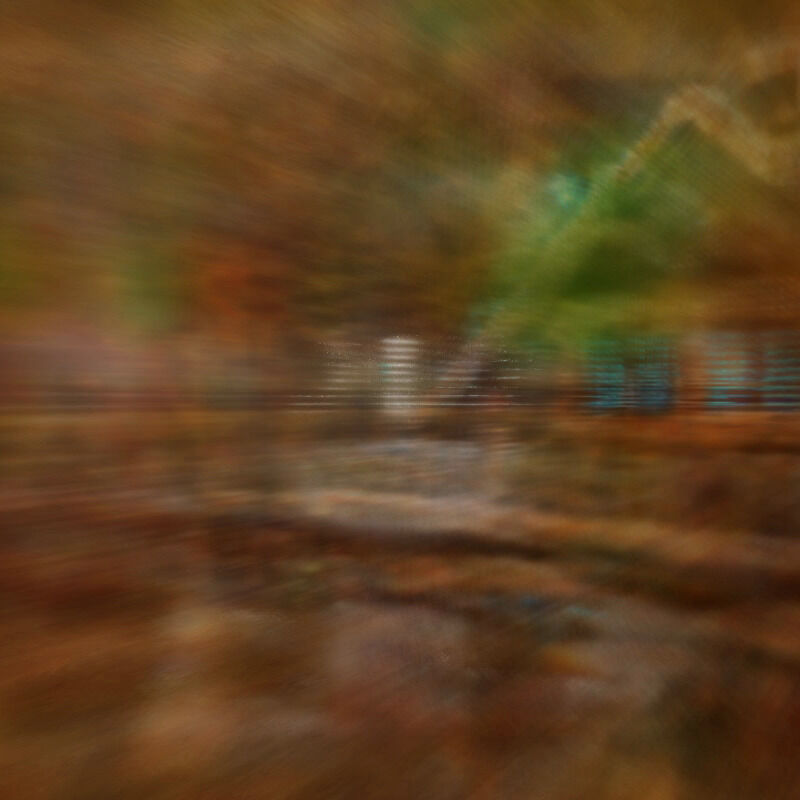}
\end{subfigure}
\begin{subfigure}{0.1 \textwidth}
\includegraphics[width=\textwidth]{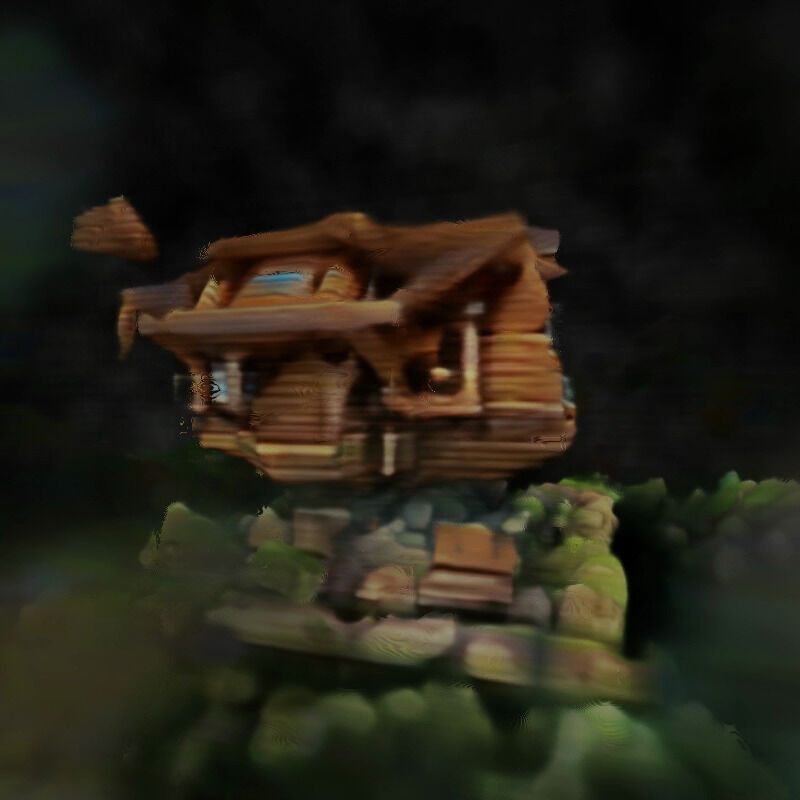}
\end{subfigure}
\begin{subfigure}{0.1 \textwidth}
\includegraphics[width=\textwidth]{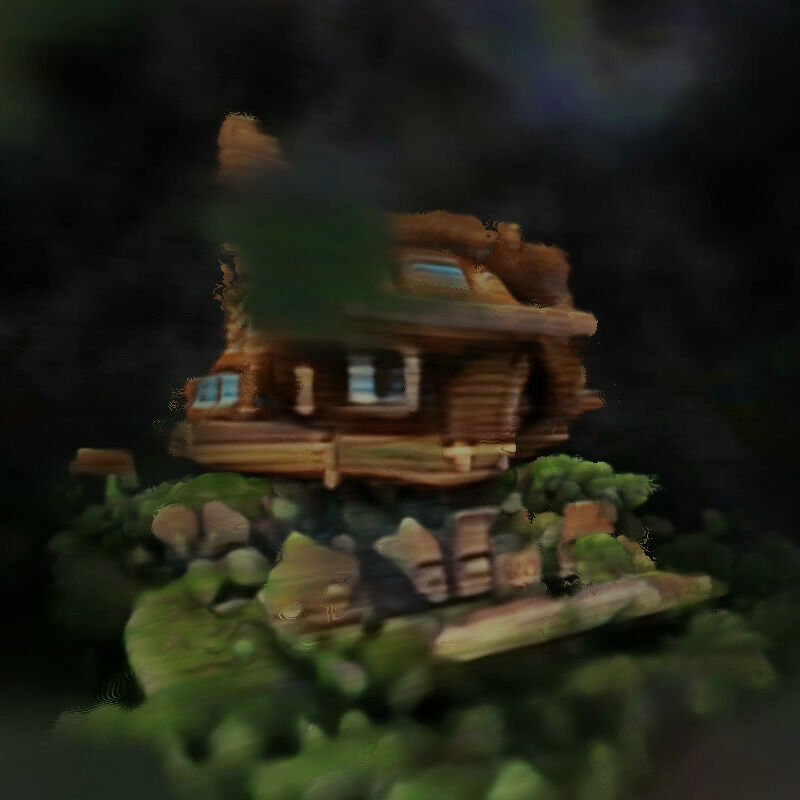}
\end{subfigure}
\begin{subfigure}{0.1 \textwidth}
\includegraphics[width=\textwidth]{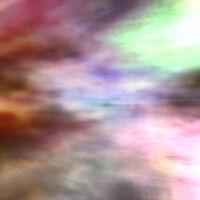}
\end{subfigure}
\begin{subfigure}{0.1 \textwidth}
\includegraphics[width=\textwidth]{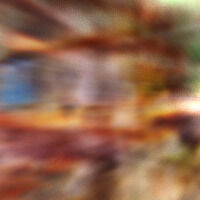}
\end{subfigure}
\begin{subfigure}{0.1 \textwidth}
\includegraphics[width=\textwidth]{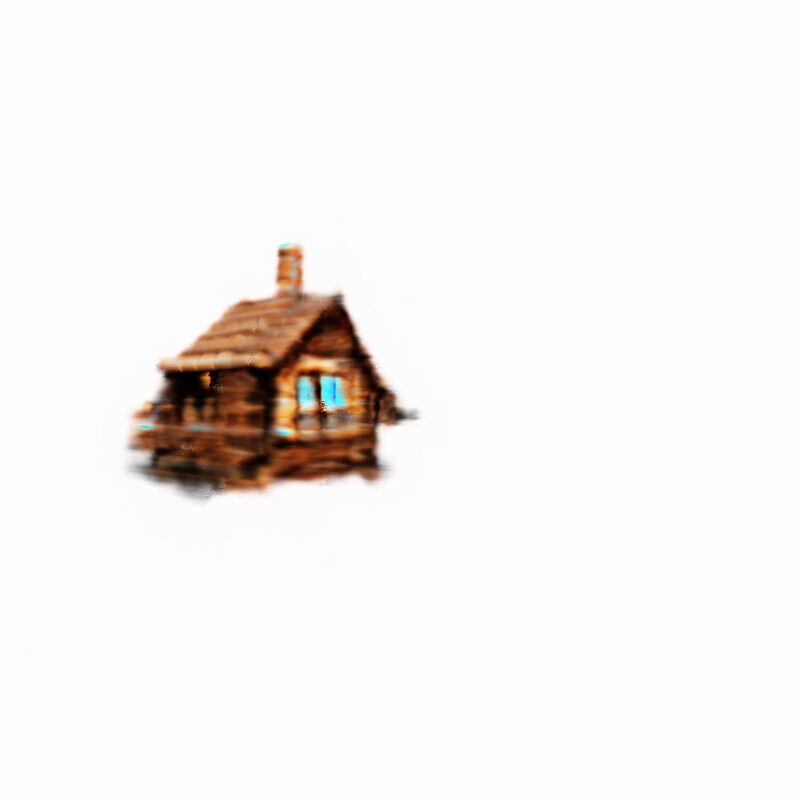}
\end{subfigure}
\begin{subfigure}{0.1 \textwidth}
\includegraphics[width=\textwidth]{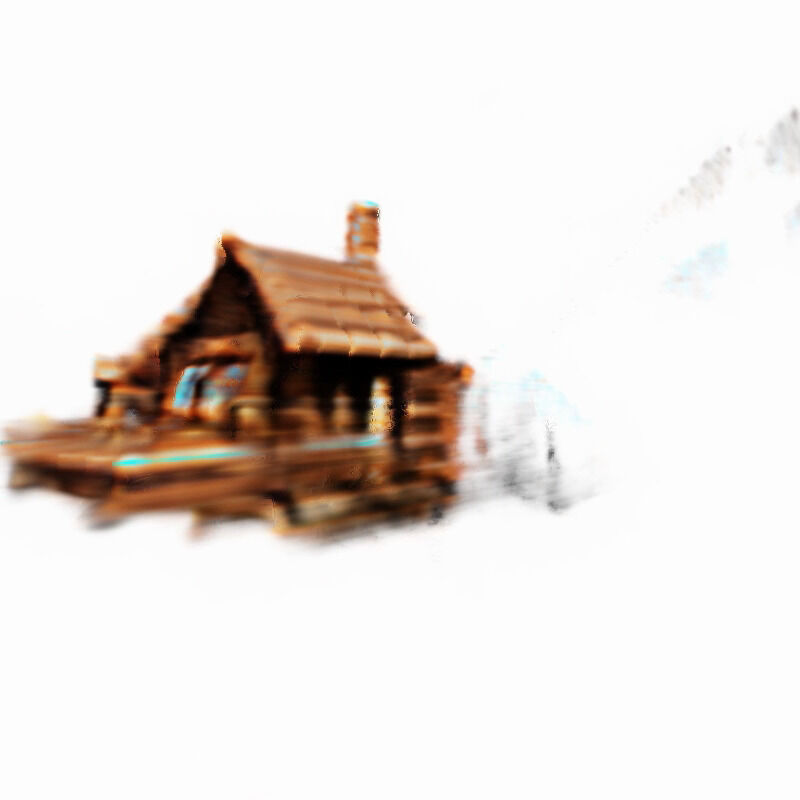}
\end{subfigure}
\\
\hspace*{.1\textwidth}
\begin{subfigure}{0.1 \textwidth}
\includegraphics[width=\textwidth]{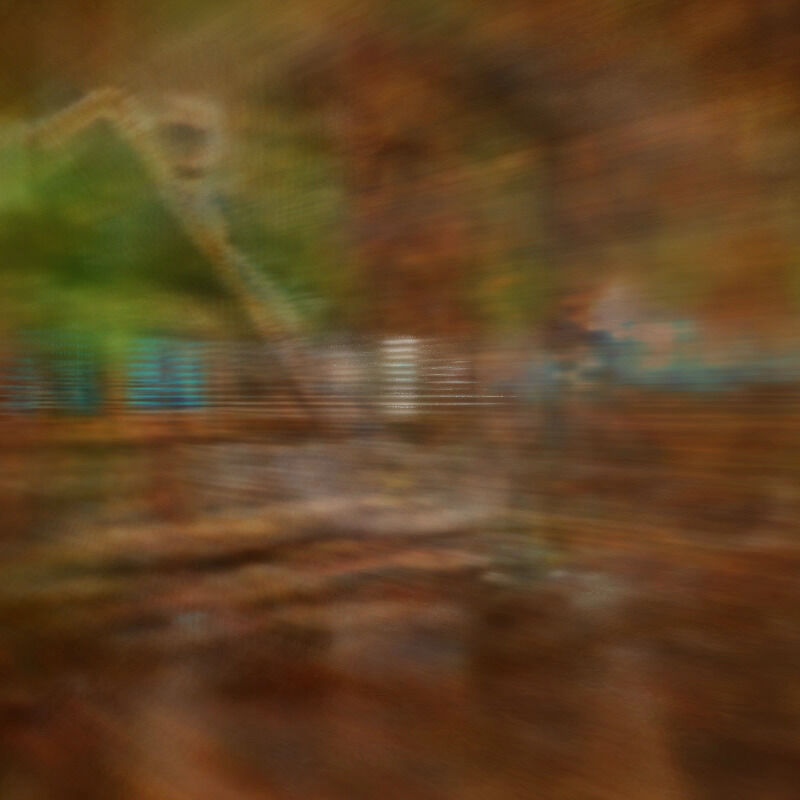}
\end{subfigure}
\begin{subfigure}{0.1 \textwidth}
\includegraphics[width=\textwidth]{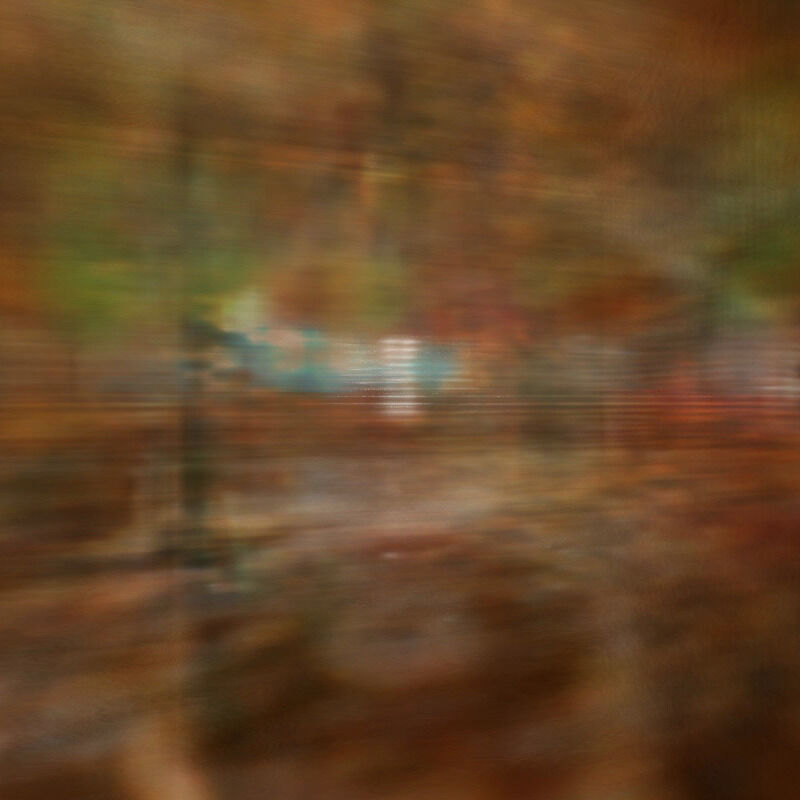}
\end{subfigure}
\begin{subfigure}{0.1 \textwidth}
\includegraphics[width=\textwidth]{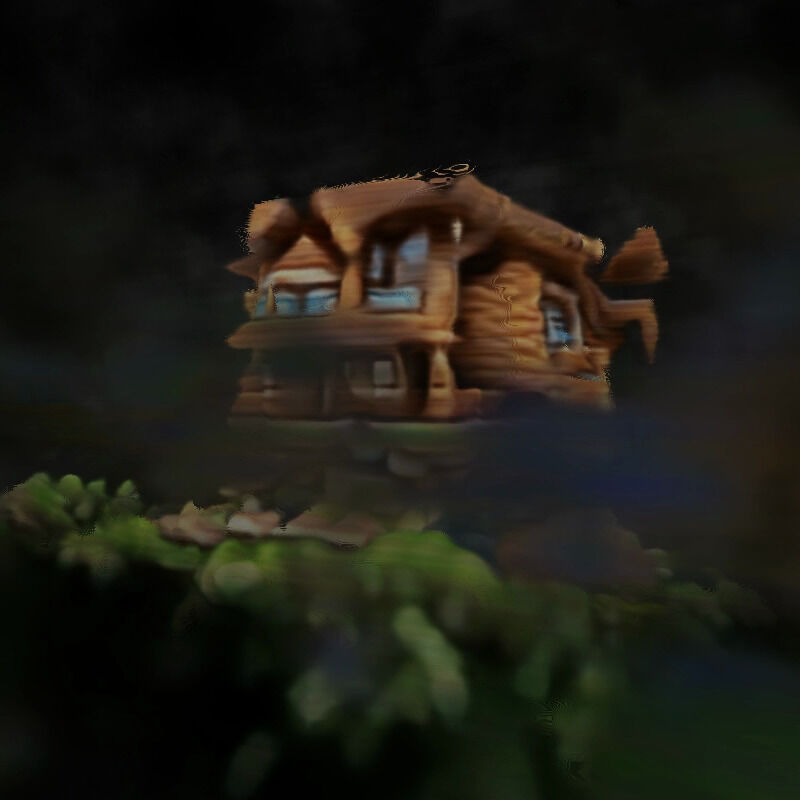}
\end{subfigure}
\begin{subfigure}{0.1 \textwidth}
\includegraphics[width=\textwidth]{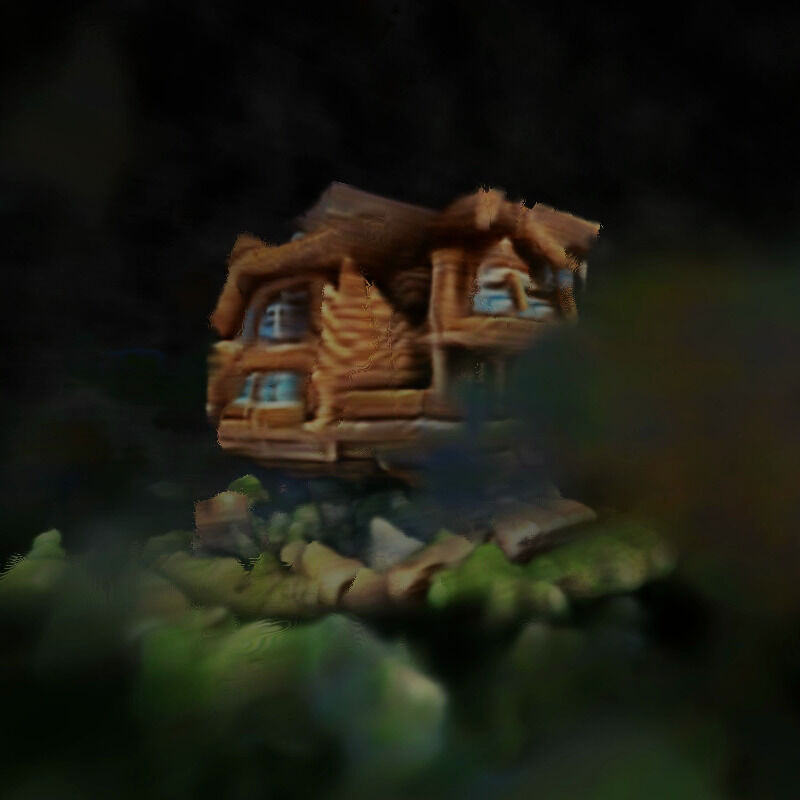}
\end{subfigure}
\begin{subfigure}{0.1 \textwidth}
\includegraphics[width=\textwidth]{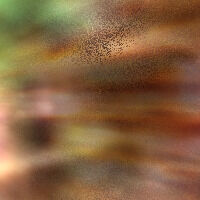}
\end{subfigure}
\begin{subfigure}{0.1 \textwidth}
\includegraphics[width=\textwidth]{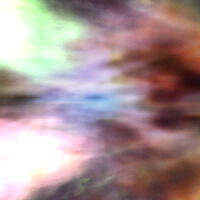}
\end{subfigure}
\begin{subfigure}{0.1 \textwidth}
\includegraphics[width=\textwidth]{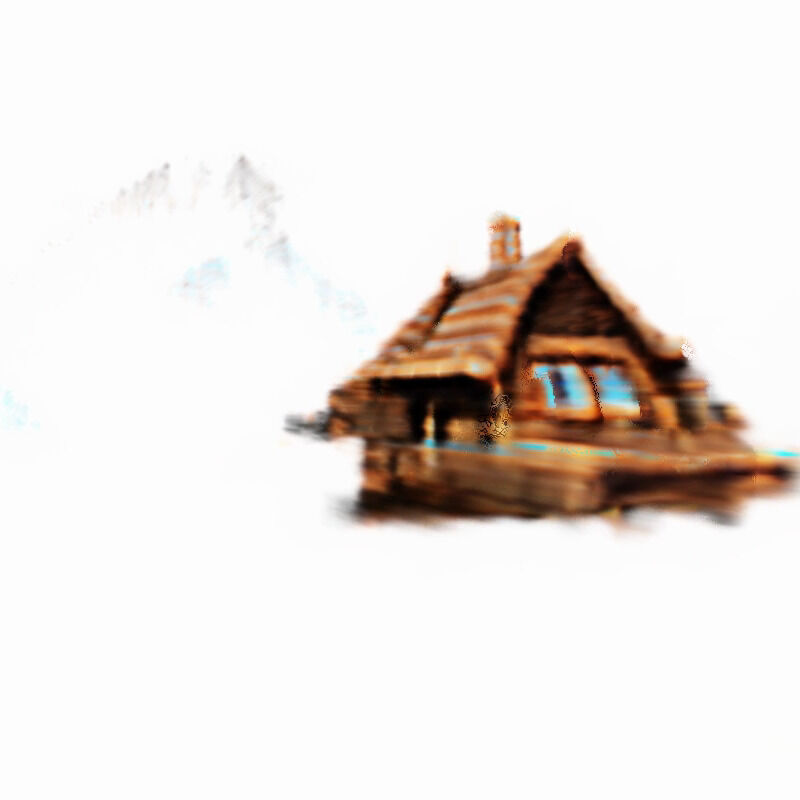}
\end{subfigure}
\begin{subfigure}{0.1 \textwidth}
\includegraphics[width=\textwidth]{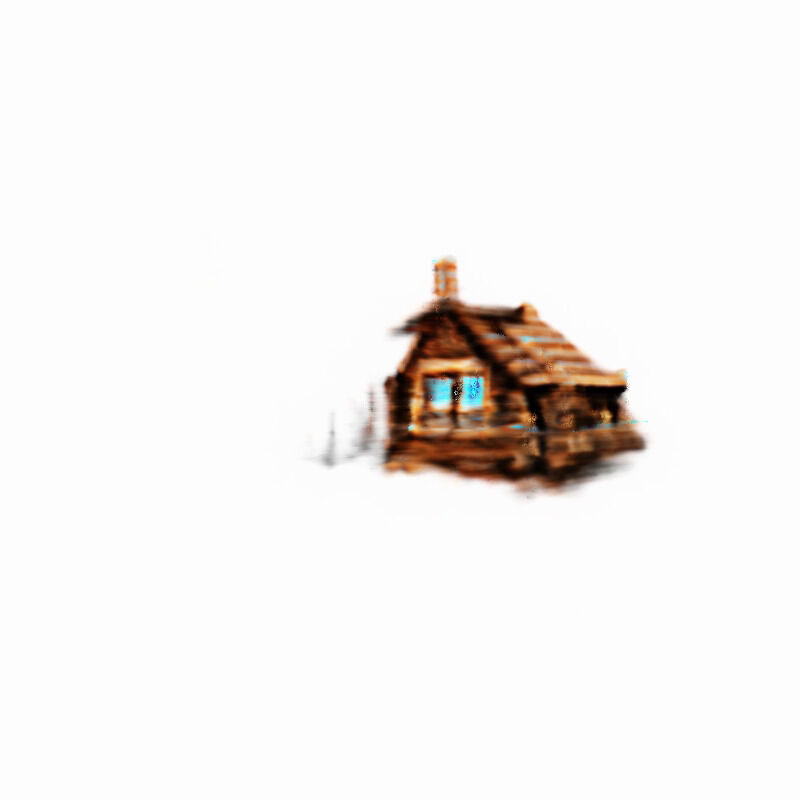}
\end{subfigure}
\begin{subfigure}{0.1 \textwidth}
\includegraphics[width=\textwidth]{images/reference/liberty5.jpg}
\end{subfigure}
\begin{subfigure}{0.1 \textwidth}
\includegraphics[width=\textwidth]{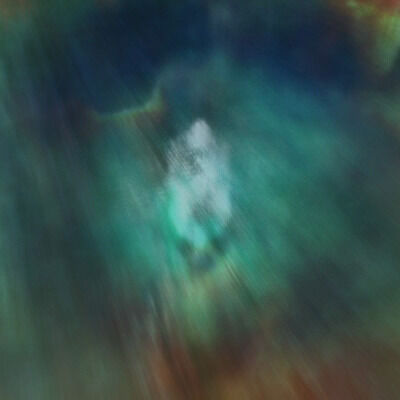}
\end{subfigure}
\begin{subfigure}{0.1 \textwidth}
\includegraphics[width=\textwidth]{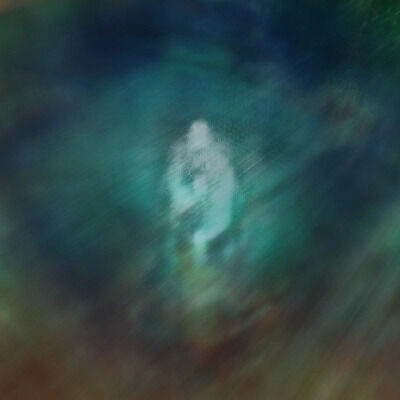}
\end{subfigure}
\begin{subfigure}{0.1 \textwidth}
\includegraphics[width=\textwidth]{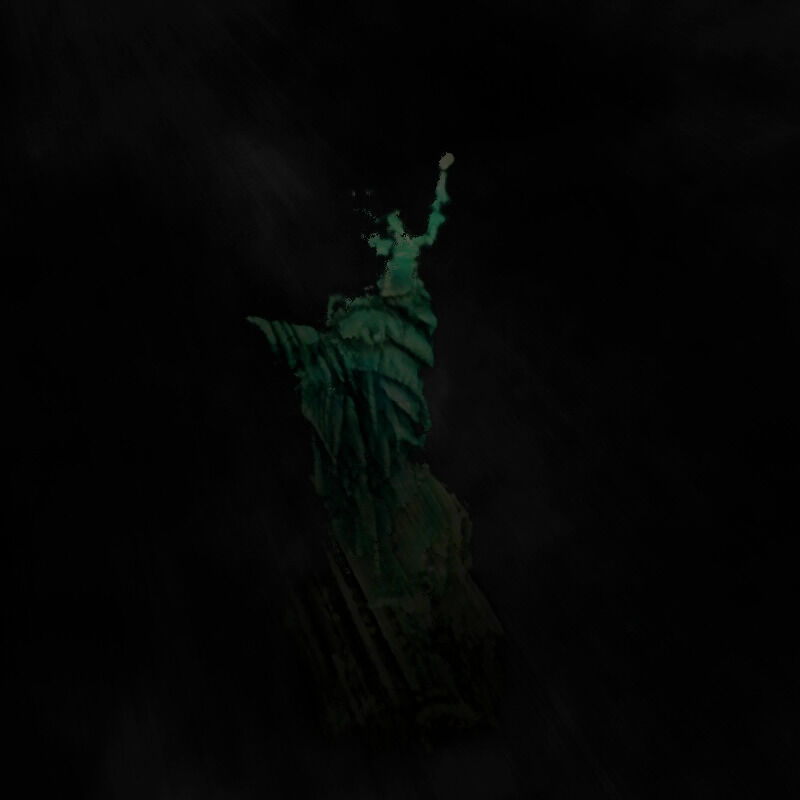}
\end{subfigure}
\begin{subfigure}{0.1 \textwidth}
\includegraphics[width=\textwidth]{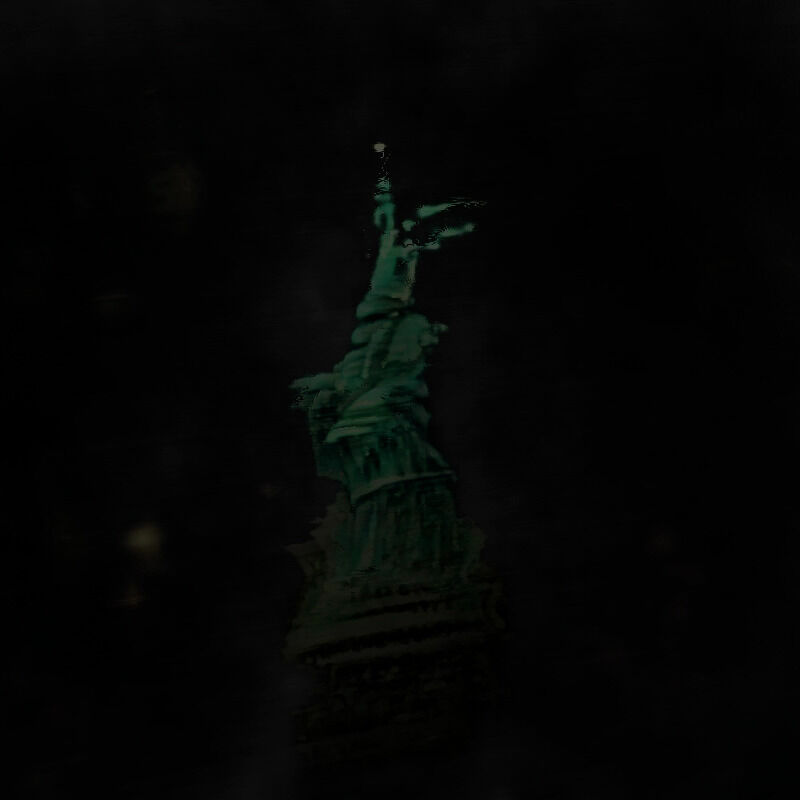}
\end{subfigure}
\begin{subfigure}{0.1 \textwidth}
\includegraphics[width=\textwidth]{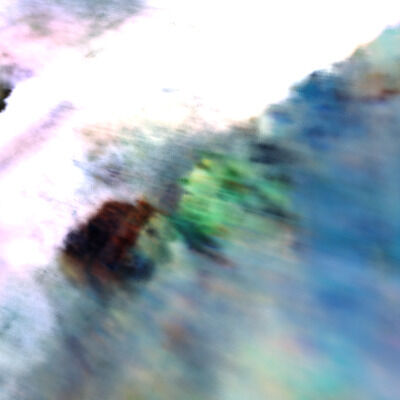}
\end{subfigure}
\begin{subfigure}{0.1 \textwidth}
\includegraphics[width=\textwidth]{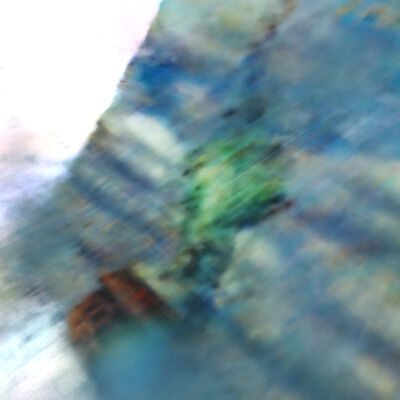}
\end{subfigure}
\begin{subfigure}{0.1 \textwidth}
\includegraphics[width=\textwidth]{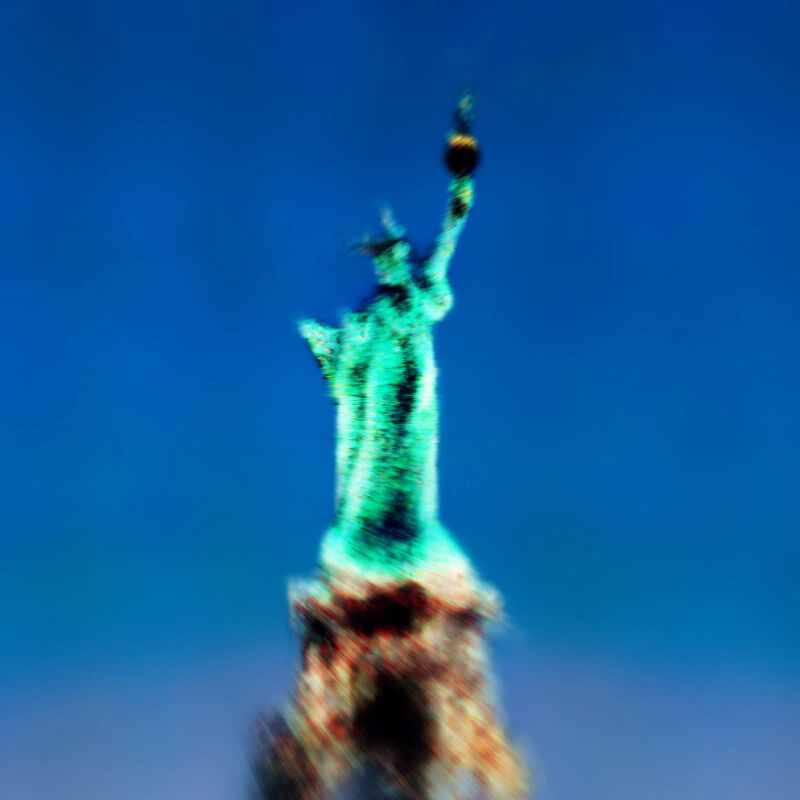}
\end{subfigure}
\begin{subfigure}{0.1 \textwidth}
\includegraphics[width=\textwidth]{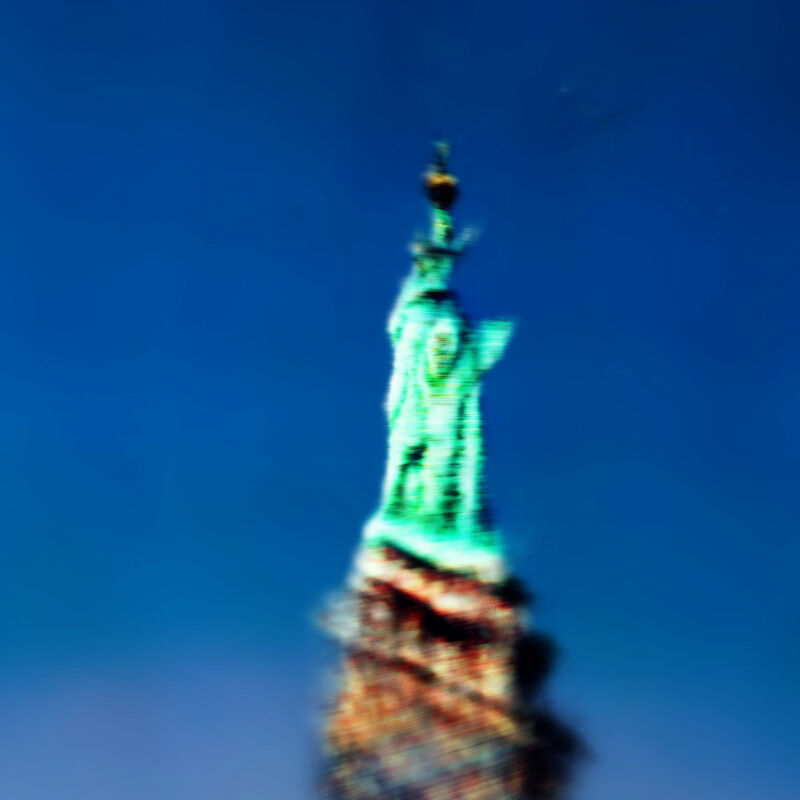}
\end{subfigure}
\\
\hspace*{.1\textwidth}
\begin{subfigure}{0.1 \textwidth}
\includegraphics[width=\textwidth]{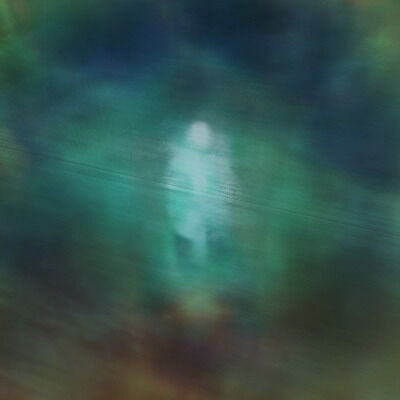}
\end{subfigure}
\begin{subfigure}{0.1 \textwidth}
\includegraphics[width=\textwidth]{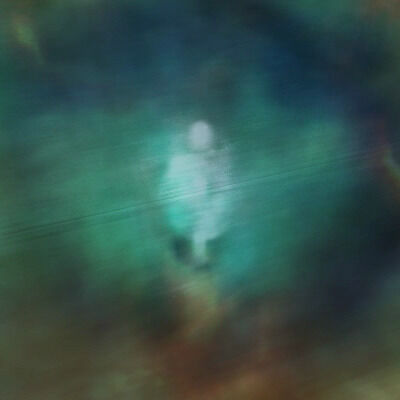}
\end{subfigure}
\begin{subfigure}{0.1 \textwidth}
\includegraphics[width=\textwidth]{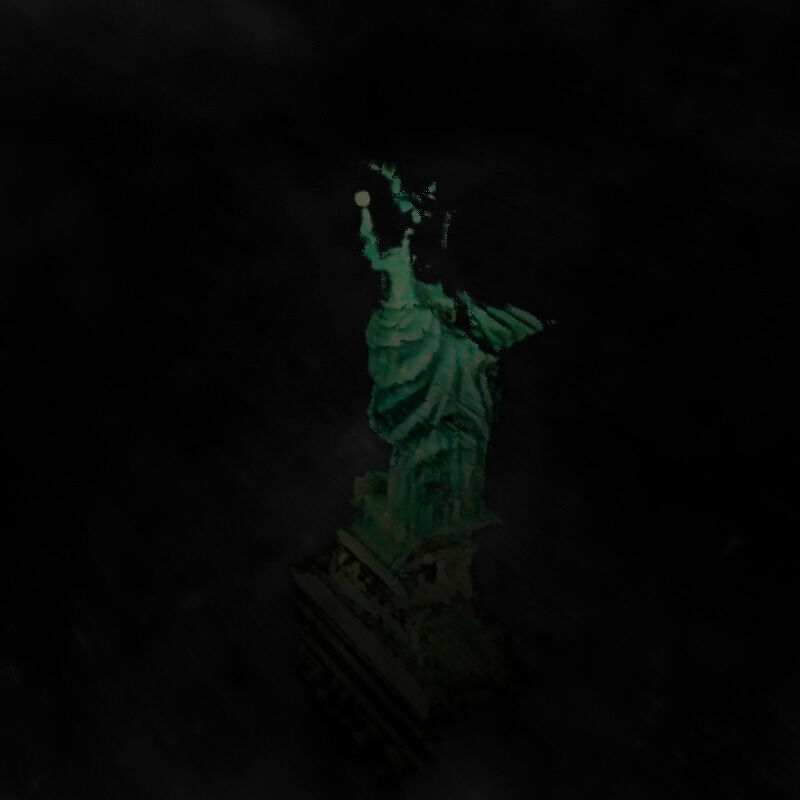}
\end{subfigure}
\begin{subfigure}{0.1 \textwidth}
\includegraphics[width=\textwidth]{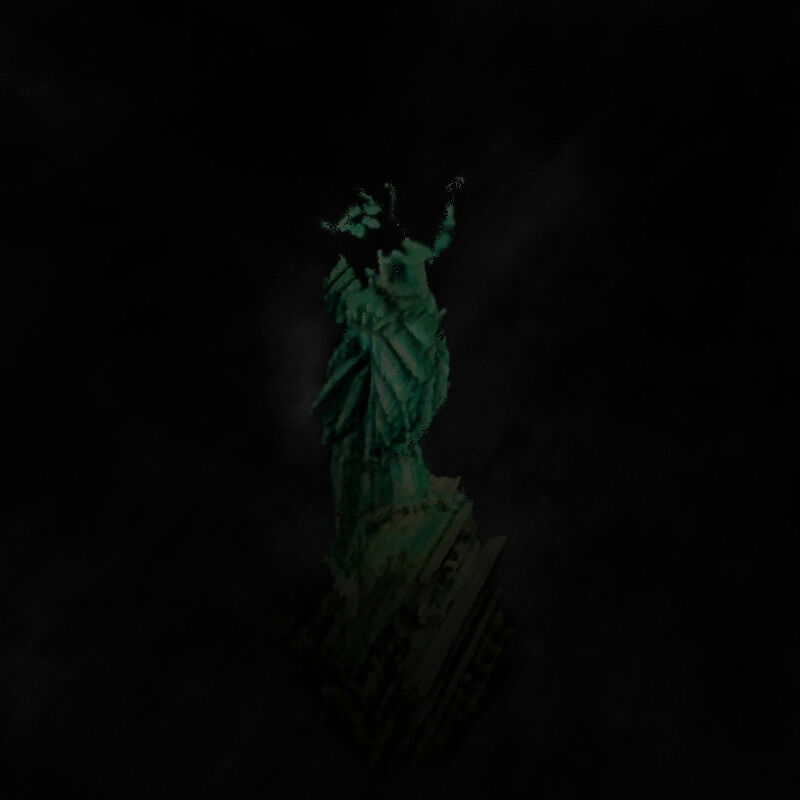}
\end{subfigure}
\begin{subfigure}{0.1 \textwidth}
\includegraphics[width=\textwidth]{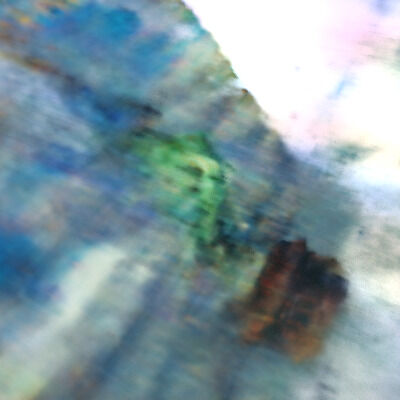}
\end{subfigure}
\begin{subfigure}{0.1 \textwidth}
\includegraphics[width=\textwidth]{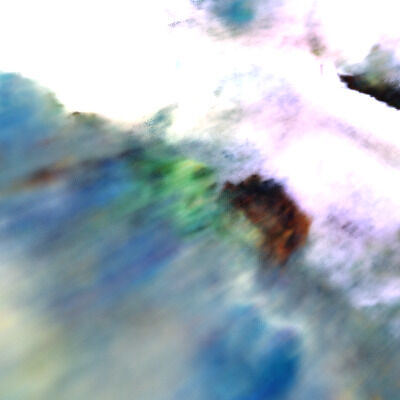}
\end{subfigure}
\begin{subfigure}{0.1 \textwidth}
\includegraphics[width=\textwidth]{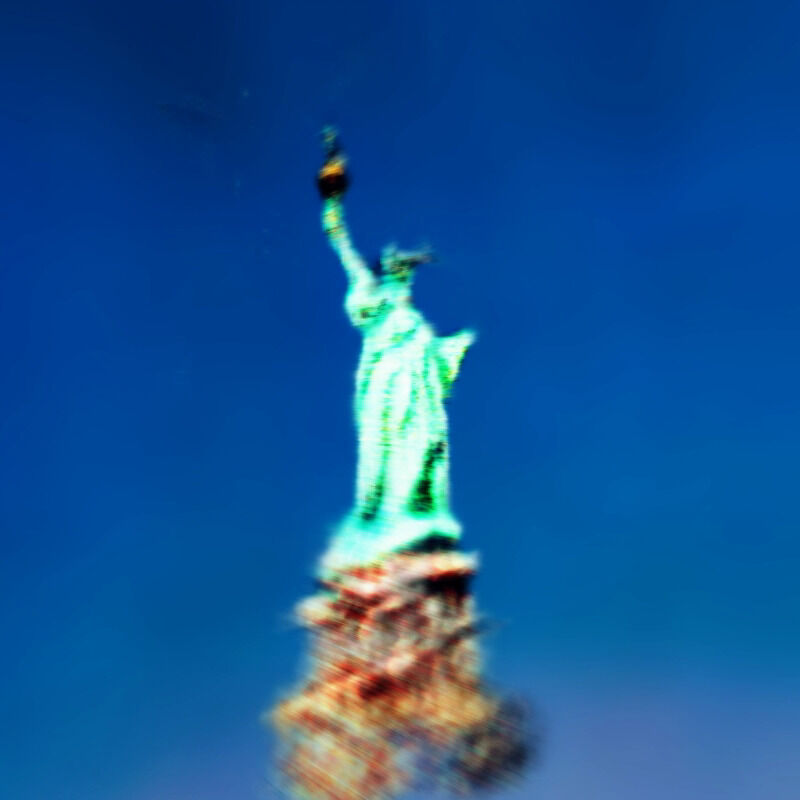}
\end{subfigure}
\begin{subfigure}{0.1 \textwidth}
\includegraphics[width=\textwidth]{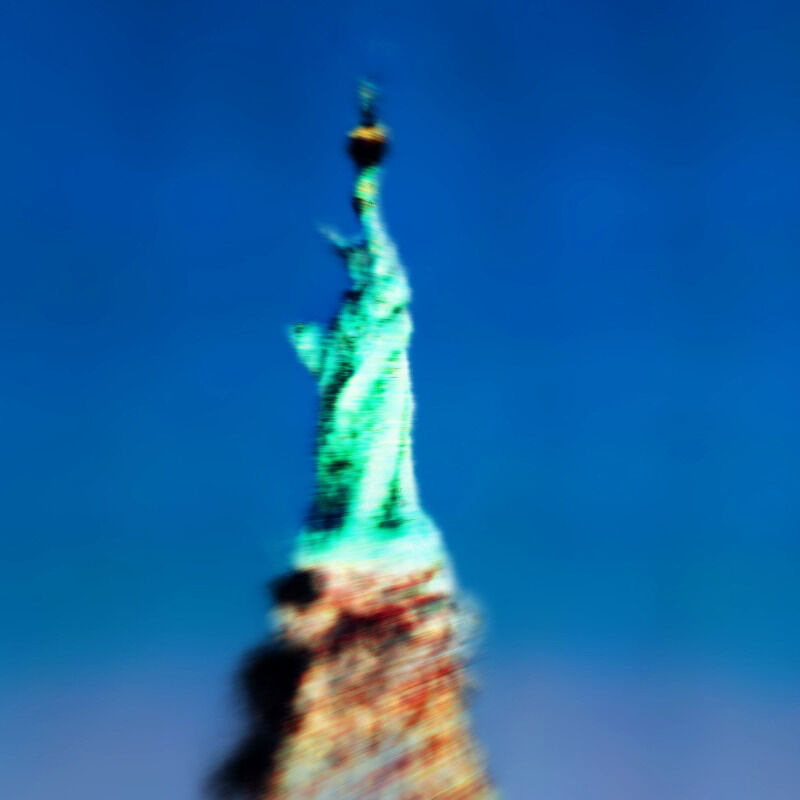}
\end{subfigure}

\caption{More visual comparisons with state-of-the-art methods.}
\label{fig:results}
\end{figure*}

\paragraph{Geometry Regularizations}

Without any regularization, the NeRF model is free to generate arbitrary geometry in those unobserved regions. Traditionally, this doesn't affect the image quality, but when we perform additional shading, the image quality becomes dependent on the geometry quality, which requires additional priors.

We observe that the generated 3D object tends to produce a flat object with foggy floaters in the back. This is due to the object generating a semi-transparent surface in the back and emitting the front view directly. We avoid this issue by penalizing the backward-facing surface normal~\cite{verbin2022ref}, 
\begin{equation}
\mathcal{L}_{\text{orient}}=\sum_i \text{detach}(w_i) \max \left(0, \hat{\mathbf{n}}_i^{\prime} \cdot {\mathbf{d}}\right)^2,
\end{equation}
where $d$ is the ray direction.
This orient loss prevents the surface normal from being in the same direction as the ray direction, so that the shading will not result in a totally black surface.

Since we don't have depth regularization for exact values, the generated contents might be floating in front of the camera. While this can deliver good RGB images in some viewing directions, the exploding geometry usually suffers from severe artifacts. To this end, we utilize distortion loss~\cite{SunSC22_2,barron2022mip} and sparsity loss to provide unsupervised regularizations. The distortion loss encourages each ray to be as compact as possible:
\begin{equation}
\begin{aligned}
\mathcal{L}_{\mathrm{dist}}(s, w) &=\sum_{i=0}^{N-1} \sum_{j=0}^{N-1} w_i w_j\left|\frac{s_i+s_{i+1}}{2}-\frac{s_j+s_{j+1}}{2}\right| \\
&+\frac{1}{3} \sum_{i=0}^{N-1} w_i^2\left(s_{i+1}-s_i\right),
\end{aligned}
\end{equation}
where ($s_{i+1}-s_i$) is the length and ($\frac{s_i+s_{i+1}}{2}$) is the midpoint of the $i$-th query interval. The sparsity loss further improves the sparsity of the occupancy by imposing an L1 regularization on the generated alpha map.
\begin{equation}
    \mathcal{L}_{\text{sparsity}} = \left\|\int_{t_{n}}^{t_{f}} T(t) \sigma({r}(t)) dt\right\|.
\end{equation}

To avoid the depth we generate being spiky, we further incorporate the inverse depth smoothing loss~\cite{wang2018learning} as a self-supervision:
\begin{equation}
\mathcal{L}_{\text {smooth}}\left(d_{i}\right)=e^{-\nabla^{2} \mathcal{I}\left(\mathbf{x}_{i}\right)}\left(\left|\partial_{x x} d_{i}\right|+\left|\partial_{x y} d_{i}\right|+\left|\partial_{y y} d_{i}\right|\right),
\end{equation}
where $d_i$ is the depth map, $\nabla^2 \mathcal{I}(\mathbf{x}_{i})$ refers to the Laplacian of pixel value at location $x_i$. This helps our rendered depth to be consistent with the rendered RGB, instead of only following the internal ranking relationship.

\paragraph{Multi-resolution rendering}

Since we use Instant-NGP~\cite{muller2022instant} based implementation to support rendering patches of $128 \times 128$ during training, obtaining the gradient using autograd is tedious to implement. We instead utilize the finite difference method to obtain the density gradient. To do so, we have to query our model eight more times in the 3D neighborhood of each point. The GPU consumption is huge if we try to obtain a normal map in high resolution, so we only render a low resolution $100 \times 100$ normal map to lower the computation burden.

\paragraph{Background Scene Representation}

The diffusion models are usually trained on images with diverse backgrounds, but we only focus on the foreground object. To solve this gap, we utilize another background NeRF to generate coherent background. The background NeRF takes viewing direction as input and generates a harmonious background. We empirically find that incorporating this background module alleviates the burden for the foreground NeRF, and it helps to remove floating artifacts.

\paragraph{Guidance weight choice}
Large guidance weight hurts the performance of general text-to-image models such as  Imagen~\cite{imagen} and Stable Diffusion~\cite{ldm} because they are performing the iterative sampling. Given a large guidance weight, the result $x_0$ prediction might exceed the range of $[-1, 1]$. This leads to a train-test mismatch for them since, during training, the input for the diffusion models is generally inside the range of $[-1, 1]$. However, such a range issue doesn't exist in our case since we only do one iteration of sampling for each rendered image. We empirically choose the guidance weight as 100.

\section{Implementation Details}

\paragraph{NeRF Architecture}

We implement our NeRF architecture based on the open-source implementation torch-ngp~\cite{torchngp}. The NeRF consists of a hash grid encoder with 16 levels and an MLP to query the features. The coarse resolution is 16, and the finest resolution is 2048. The number of feature dimensions per entry is 2 for the hash grid. Our MLP consists of a sequence of four fully-connected layers with residual connections. The activations layers in between are SiLU~\cite{elfwing2018sigmoid}. The final output layer of the NeRF MLP is four channels, consisting of color and density.  We further employ a Sigmoid activation on the density before volumetric rendering.
We also adopt a background NeRF to module the background scene. The background module conditions on the ray direction and has three fully-connected layers.

\paragraph{Diffusion Prior Settings}
We utilize Stable Diffusion version 1.4 as our diffusion prior. During training, we uniformly sample the timesteps in the range of $[50, 950]$. This design stabilizes training by avoiding very high and very low noise levels. We use PNDM~\cite{liu2022pseudo} Sampler with a training timestep of 1,000 to perform the noise perturbation and calculate the $\hat{x_0}$ estimate. The $\beta$ range is [0.00085, 0.012], and the $\beta$ schedule is scaled linear.

\paragraph{Training Settings}

We implement our framework using PyTorch. To speed up the training, we implement gradient propagation 
for depth using CUDA with the help of PyTorch extensions. The whole framework is trained on a single A6000 GPU for 10k steps. The overall training takes about 1.5 hours. For the first 1k steps, we disable the shading, and for the remaining steps, we use diffuse shading for 50\% of the time. During inference, we disable the shading.

The initial learning rate is $1e-3$, and decay the learning rate using LambdaLR Scheduler in PyTorch. The optimizer is Adam, and the betas are set to $(0.9, 0.99)$.
We set the weight for $\mathcal{L}_\text{orient}$ to be 10. For $\mathcal{L}_\text{dist}$, we find reasonable weights to lie in [$1e-2$, $1e-1$]. For $\mathcal{L}_\text{sparsity}$, we set the weight to be $1e-3$. For $\mathcal{L}_\text{smooth}$, we set the weight to be $0.1$.

\end{document}